# Incremental Clustering and Expansion for Faster Optimal Planning in Decentralized POMDPs

**Frans A. Oliehoek** FRANS.OLIEHOEK@MAASTRICHTUNIVERSITY.NL
*Maastricht University*
*Maastricht, The Netherlands*

**Matthijs T.J. Spaan** M.T.J.SPAAN@TUDELFT.NL
*Delft University of Technology*
*Delft, The Netherlands*

**Christopher Amato** CAMATO@CSAIL.MIT.EDU
*Massachusetts Institute of Technology*
*Cambridge, MA, USA*

**Shimon Whiteson** S.A.WHITESON@UVA.NL
*University of Amsterdam*
*Amsterdam, The Netherlands*

## Abstract

This article presents the state-of-the-art in optimal solution methods for *decentralized partially observable Markov decision processes (Dec-POMDPs)*, which are general models for collaborative multiagent planning under uncertainty. Building off the *generalized multiagent A* (*GMAA**)* algorithm, which reduces the problem to a tree of one-shot *collaborative Bayesian games (CBGs)*, we describe several advances that greatly expand the range of Dec-POMDPs that can be solved optimally. First, we introduce lossless *incremental clustering* of the CBGs solved by GMAA*, which achieves exponential speedups without sacrificing optimality. Second, we introduce *incremental expansion* of nodes in the GMAA* search tree, which avoids the need to expand all children, the number of which is in the worst case doubly exponential in the node's depth. This is particularly beneficial when little clustering is possible. In addition, we introduce new hybrid heuristic representations that are more compact and thereby enable the solution of larger Dec-POMDPs. We provide theoretical guarantees that, when a suitable heuristic is used, both incremental clustering and incremental expansion yield algorithms that are both complete and *search equivalent*. Finally, we present extensive empirical results demonstrating that GMAA*-ICE, an algorithm that synthesizes these advances, can optimally solve Dec-POMDPs of unprecedented size.

## 1. Introduction

A key goal of artificial intelligence is the development of intelligent agents that interact with their environment in order to solve problems, achieve goals, and maximize utility. While such agents sometimes act alone, researchers are increasingly interested in collaborative *multiagent systems*, in which teams of agents work together to perform all manner of tasks. Multiagent systems are appealing, not only because they can tackle inherently distributed problems, but because they facilitate the decomposition of problems too complex to be tackled by a single agent (Huhns, 1987; Sycara, 1998; Panait & Luke, 2005; Vlassis, 2007; Buşoniu, Babuška, & De Schutter, 2008).

One of the primary challenges of multiagent systems is the presence of uncertainty. Even in single-agent systems, the outcome of an action may be uncertain, e.g., the action may fail with some probability. Furthermore, in many problems the state of the environment may be uncertain due to limited or noisy sensors. However, in multiagent settings these problems are often greatly exacerbated. Since agents have access only to their own sensors, typically a small fraction of those of the complete system, their ability to predict how other agents will act is limited, complicating cooperation. If such uncertainties are not properly addressed, arbitrarily bad performance may result.

In principle, agents can use communication to synchronize their beliefs and coordinate their actions. However, due to bandwidth constraints, it is typically infeasible for all agents to broadcast the necessary information to all other agents. In addition, in many realistic scenarios, communication may be unreliable, precluding the possibility of eliminating all uncertainty about other agents' actions.

Especially in recent years, much research has focused on approaches to (collaborative) multiagent systems that deal with uncertainty in a principled way, yielding a wide variety of models and solution methods (Pynadath & Tambe, 2002; Goldman & Zilberstein, 2004; Seuken & Zilberstein, 2008). This article focuses on the *decentralized partially observable Markov decision process (Dec-POMDP)*, a general model for collaborative multiagent planning under uncertainty. Unfortunately, solving a Dec-POMDP, i.e., computing an optimal plan, is generally intractable (NEXP-complete) (Bernstein, Givan, Immerman, & Zilberstein, 2002) and in fact even computing solutions with absolutely bounded error (i.e., $\epsilon$-approximate solutions) is also NEXP-complete (Rabinovich, Goldman, & Rosenschein, 2003). In particular, the number of joint policies grows exponentially with the number of agents and observations and doubly exponentially with respect to the horizon of the problem.[1] Though these complexity results preclude methods that are efficient on all problems, developing better optimal solution methods for Dec-POMDPs is nonetheless an important goal, for several reasons.

First, since the complexity results describe only the worst case, there is still great potential to improve the performance of optimal methods in practice. In fact, there is evidence that many problems can be solved much faster than the worst-case complexity bound indicates (Allen & Zilberstein, 2007). In this article, we present experiments that clearly demonstrate this point: on many problems, the methods we propose scale vastly beyond what would be expected for a doubly-exponential dependence on the horizon.

Second, as computer speed and memory capacity increase, a growing set of small and medium-sized problems can be solved optimally. Some of these problems arise naturally while others result from the decomposition of larger problems. For instance, it may be possible to extrapolate optimal solutions of problems with shorter planning horizons, using them as the starting point of policy search for longer-horizon problems as in the work of Eker and Akın (2013), or to use such shorter-horizon, no-communication solutions inside problems with communication (Nair, Roth, & Yohoo, 2004; Goldman & Zilberstein, 2008). More generally, optimal policies of smaller problems can potentially be used to find good solutions for larger problems. For instance, *transfer planning* (Oliehoek, 2010; Oliehoek, Whiteson, & Spaan,

1. Surprisingly, the number of states in a Dec-POMDP is less important, e.g., brute-force search depends on the number of states only via its policy evaluation routine, which scales linearly in the number of states.

2013) employs optimal solutions to problems with few agents to better solve problems with many agents. By performing (approximate) influence-based abstraction and influence search (Witwicki, 2011; Oliehoek, Witwicki, & Kaelbling, 2012), optimal solutions of component problems can potentially be used to find (near-)optimal solutions of larger problems.

Third, optimal methods offer important insights into the nature of specific Dec-POMDP problems and their solutions. For instance, the methods introduced in this article enabled the discovery of certain properties of the BroadcastChannel benchmark problem that make it much easier to solve.

Fourth, optimal methods provide critical inspiration for principled approximation methods. In fact, almost all successful approximate Dec-POMDP methods are based on optimal ones (see, e.g., Seuken & Zilberstein, 2007b, 2007a; Dibangoye, Mouaddib, & Chai-draa, 2009; Amato, Dibangoye, & Zilberstein, 2009; Wu, Zilberstein, & Chen, 2010a; Oliehoek, 2010) or locally optimal ones (Velagapudi, Varakantham, Scerri, & Sycara, 2011)[2], and the clustering technique presented in this article forms the basis of a recently introduced approximate clustering technique (Wu, Zilberstein, & Chen, 2011).

Finally, optimal methods are essential for benchmarking approximate methods. In recent years, there have been huge advances in the approximate solution of Dec-POMDPs, leading to the development of solution methods that can deal with large horizons, hundreds of agents and many states (e.g., Seuken & Zilberstein, 2007b; Amato et al., 2009; Wu et al., 2010a; Oliehoek, 2010; Velagapudi et al., 2011). However, since computing even $\epsilon$-approximate solutions is NEXP-complete, any method whose complexity is not doubly exponential cannot have any guarantees on the absolute error of the solution (assuming EXP$\neq$NEXP). As such, existing effective approximate methods have no quality guarantees.[3]

Consequently, it is difficult to meaningfully interpret their empirical performance without the upper bounds optimal methods supply. While approximate methods can also be benchmarked against lower bounds (e.g., other approximate methods), such comparisons cannot detect when a method fails to find good solutions. Doing so requires benchmarking against upper bounds and, unfortunately, upper bounds that are easier to compute, such as QMDP and QPOMDP, are too loose to be helpful (Oliehoek, Spaan, & Vlassis, 2008). As such, benchmarking with respect to optimal solutions is an important part of the verification of any approximate algorithm. Since existing optimal methods can only tackle very small problems, scaling optimal solutions to larger problems is a critical goal.

## 1.1 Contributions

This article presents the state-of-the-art in optimal solution methods for Dec-POMDPs. In particular, it describes several advances that greatly expand the horizon to which many Dec-POMDPs can be solved optimally. In addition, it proposes and evaluates a complete algorithm that synthesizes these advances. Our approach is based on the *generalized multiagent* A* (GMAA*) algorithm (Oliehoek, Spaan, & Vlassis, 2008), which makes it possible to reduce the problem to a tree of one-shot *collaborative Bayesian games (CBGs).* The appeal of this

2. The method by Velagapudi et al. (2011) repeatedly computes best responses in a way similar to DP-JESP (Nair, Tambe, Yokoo, Pynadath, & Marsella, 2003). The best response computation, however, exploits sparsity of interactions.
3. Note that we refer to methods without quality guarantees as *approximate* rather than *heuristic* to avoid confusion with *heuristic search*, which is used throughout this article and is exact.

approach is the abstraction layer it introduces, which has led to various insights into Dec-POMDPs and, in turn, to the improved solution methods we describe.

The specific contributions of this article are:[4]

1. We introduce lossless clustering of CBGs, a technique to reduce the size of the CBGs for which GMAA* enumerates all possible solutions, while preserving optimality. This can exponentially reduce the number of child nodes in the GMAA* search tree, leading to huge increases in efficiency. In addition, by applying *incremental clustering (IC)* to GMAA*, our GMAA*-IC method can avoid clustering exponentially sized CBGs.

2. We introduce *incremental expansion (IE)* of nodes in the GMAA* search tree. Although clustering may reduce the number of children of a search node, this number is in the worst case still doubly exponential in the node's depth. GMAA*-ICE, which applies IE to GMAA*-IC, addresses this problem by creating a next child node only when it is a candidate for further expansion.

3. We provide theoretical guarantees for both GMAA*-IC and GMAA*-ICE. In particular, we show that, when using a suitable heuristic, both algorithms are both complete and *search equivalent*.

4. We introduce an improved heuristic representation. Tight heuristics like those based on the underlying POMDP solution ($Q_{\text{POMDP}}$) or the value function resulting from assuming 1-step-delayed communication ($Q_{\text{BG}}$) are essential for heuristic search methods like GMAA* (Oliehoek, Spaan, & Vlassis, 2008). However, the space needed to store these heuristics grows exponentially with the problem horizon. We introduce hybrid representations that are more compact and thereby enable the solution of larger problems.

5. We present extensive empirical results that show substantial improvements over the current state-of-the-art. Whereas Seuken and Zilberstein (2008) argued that GMAA* can at best optimally solve Dec-POMDPs only one horizon further than brute-force search, our results demonstrate that GMAA*-ICE can do much better. In addition, we provide a comparative overview of the results of competitive optimal solution methods from the literature.

The primary aim of the techniques introduced in this article is to improve scalability with respect to the horizon. Our empirical results confirm that these techniques are highly successful in this regard. As an added bonus, our experiments also demonstrate improvement in scalability with respect to the number of agents. In particular, we present the first optimal results on general (non-special case) Dec-POMDPs with more than three agents. Extensions of our techniques to achieve further improvements with respect to the number of agents, as well as promising ways to combine the ideas behind our methods with state-of-the-art approximate approaches, are discussed under future work in Section 7.

4. This article synthesizes and extends research that was already reported in two conference papers (Oliehoek, Whiteson, & Spaan, 2009; Spaan, Oliehoek, & Amato, 2011).

### 1.2 Organization

The article is organized as follows. Section 2 provides background on the Dec-POMDP model, the GMAA* heuristic search solution method, as well as suitable heuristics. In Section 3, we introduce lossless clustering of the CBGs and its integration into GMAA*. Section 4 introduces the incremental expansion of search nodes. The empirical evaluation of the proposed techniques is reported in Section 5. We give a treatment of related work in Section 6. Future work is discussed in Section 7 and conclusions are drawn in Section 8.

## 2. Background

In a Dec-POMDP, multiple agents must collaborate to maximize the sum of the common rewards they receive over multiple timesteps. Their actions affect not only their immediate rewards but also the state to which they transition. While the current state is not known to the agents, at each timestep each agent receives a private observation correlated with that state.

**Definition 1.** A Dec-POMDP is a tuple $\left\langle \mathcal{D}, \mathcal{S}, \boldsymbol{\mathcal{A}}, T, \boldsymbol{\mathcal{O}}, O, R, \boldsymbol{b}^0, h \right\rangle$, where

- $\mathcal{D} = \{1, \ldots, n\}$ is the finite set of agents.
- $\mathcal{S} = \left\{s_1, \ldots, s_{|\mathcal{S}|}\right\}$ is the finite set of states.
- $\boldsymbol{\mathcal{A}} = \times_i \mathcal{A}_i$ is the set of *joint actions* $\boldsymbol{a} = \langle a_1, \ldots, a_n \rangle$, where $\mathcal{A}_i$ is the finite set of actions available to agent $i$.
- $T$ is a transition function specifying the state transition probabilities $\Pr(s'|s,\boldsymbol{a})$.
- $\boldsymbol{\mathcal{O}} = \times_i \mathcal{O}_i$ is the finite set of joint observations. At every stage one joint observation $\boldsymbol{o} = \langle o_1,...,o_n \rangle$ is received. Each agent $i$ observes only its own component $o_i$.
- $O$ is the observation function, which specifies observation probabilities $\Pr(\boldsymbol{o}|\boldsymbol{a},s')$.
- $R(s,\boldsymbol{a})$ is the immediate reward function mapping $(s,\boldsymbol{a})$-pairs to real numbers.
- $\boldsymbol{b}^0 \in \Delta(\mathcal{S})$ is the initial state distribution at time $t = 0$, where $\Delta(\mathcal{S})$ denotes the infinite set of probability distributions over the finite set $\mathcal{S}$.
- $h$ is the horizon, i.e., the number of stages. We consider the case where $h$ is finite.

At each stage $t = 0 \ldots h - 1$, each agent takes an individual action and receives an individual observation.

*Example* 1 (Recycling Robots). To illustrate the Dec-POMDP model, consider a team of robots tasked with removing trash from an office building, depicted in Fig. 1. The robots have sensors to find marked trash cans, motors to move around in order to look for cans, as well as gripper arms to grasp and carry a can. Small trash cans are light and compact enough for a single robot to carry, but large trash cans require multiple robots to carry them out together. Because more people use them, the larger trash cans fill up more quickly. Each robot must also ensure that its battery remains charged by moving to a charging station before it expires. The battery level for a robot degrades due to the distance the robot travels and the weight of the item being carried. Each robot knows its own battery level but not that of the other robots and only the location of other robots within sensor range. The goal of this problem is to remove as much trash as possible in a given time period.

This problem can be represented as a Dec-POMDP in a natural way. The states, $\mathcal{S}$, consist of the different locations of each robot, their battery levels and the different amounts of trash in the cans. The actions, $\mathcal{A}_i$, for each robot consist of movements in different directions as well as decisions to pick up a trash can or recharge the battery (when in range of a can or a charging station). The observations, $\mathcal{O}_i$, of each robot consist of its own battery level, its own location, the locations of other robots in sensor range and the amount of trash in cans within range. The rewards, $R$, could consist of a large positive value for a pair of robots emptying a large (full) trash can, a small positive value for a single robot emptying a small trash can and negative values for a robot depleting its battery or a trash can overflowing. An optimal solution is a joint policy that leads to the expected behavior (given that the rewards are properly specified). That is, it ensures that the robots cooperate to empty the large trash cans when appropriate and the small ones individually while considering battery usage.

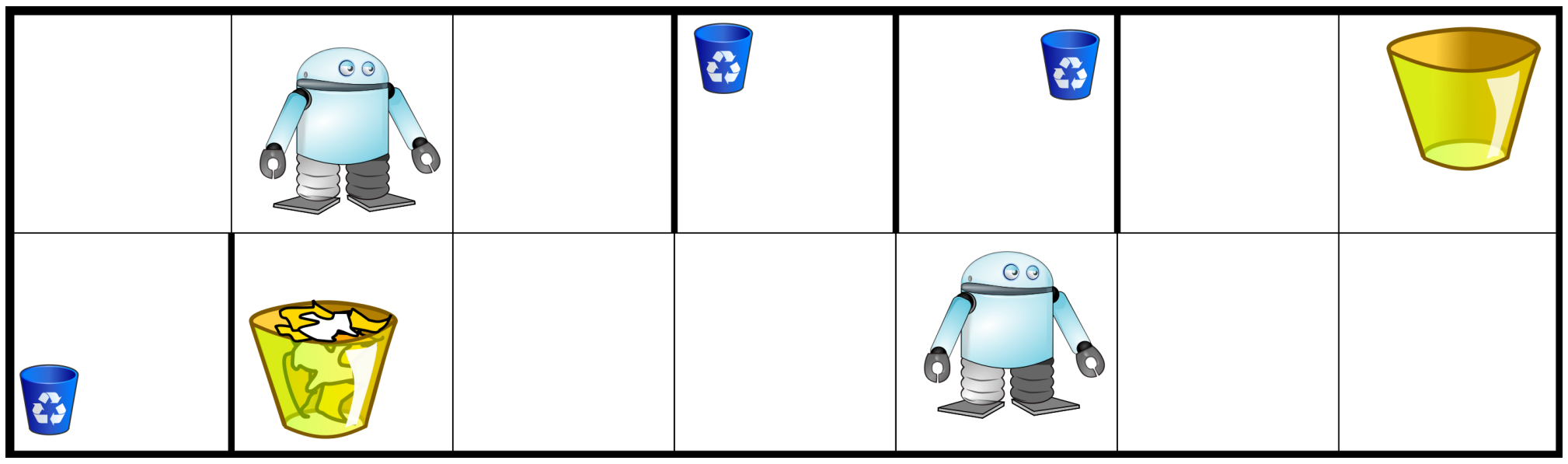

Figure 1: Illustration of the Recycling Robots example, in which two robots have to remove trash in an office environment with three small (blue) trash cans and two large (yellow) ones. In this situation, the left robot might observe that the large trash can next to it is full, and the other robot that the small trash can is empty. However, none of them is sure of the trash cans' state due to limited sensing capabilities, nor do they see the state of trash cans further away. In particular, one robot has no knowledge regarding the observations of the other robot.

For explanatory purposes, we also consider a much simpler problem, the so-called decentralized tiger problem (Nair et al., 2003).

*Example* 2 (Dec-Tiger). The Dec-Tiger problem concerns two agents that find themselves in a hallway with two doors. Behind one door, there is a treasure and behind the other is a tiger. The state describes which door the tiger is behind—left ($s_l$) or right ($s_r$)—each occurring with 0.5 probability (i.e., the initial state distribution $\boldsymbol{b}^0$ is uniform). Each agent can perform three actions: open the left door ($a_{\text{OL}}$), open the right door ($a_{\text{OR}}$) or listen ($a_{\text{Li}}$). Clearly, opening the door to the treasure will yield a reward, but opening the door to the tiger will result in a severe penalty. A greater reward is given for both agents opening the correct door at the same time. As such, a good strategy will probably involve listening first. The listen actions, however, also have a minor cost (negative reward). At every stage the agents get an observation. The agents can either hear the tiger behind the left ($o_{\text{HL}}$) or right ($o_{\text{HR}}$) door, but each agent has a 15% chance of hearing it incorrectly (getting the wrong observation). Moreover, the observation is informative only if both agents listen; if either agent opens a door, both agents receive an uninformative (uniformly drawn) observation and the problem resets to $s_l$ or $s_r$ with equal probability. At this point the problem just continues, such that the agents may be able to open the door to the treasure multiple times. Also note that, since the only two observations the agents can get are $o_{\text{HL}}, o_{\text{HR}}$, the agents have no way of detecting that the problem has been reset: if one agent opens the door while the other listens, the other agent will not be able to tell that the door was opened. For a complete specification, see the discussion by Nair et al. (2003).

Given a Dec-POMDP, the agents' common goal is to maximize the expected *cumulative reward* or *return*. The planning task entails finding a *joint* policy $\boldsymbol{\pi} = \langle \pi_1, \ldots, \pi_n \rangle$ from the space of joint policies $\boldsymbol{\Pi}$, that specifies an individual policy $\pi_i$ for each agent $i$. Such an individual policy in general specifies an individual action for each *action-observation history* (AOH) $\vec{\theta}_i^t = (a_i^0, o_i^1, \ldots, a_i^{t-1}, o_i^t)$, e.g., $\pi_i(\vec{\theta}_i^t) = a_i^t$. However, it is possible to restrict our attention to *deterministic* or *pure* policies, in which case $\pi_i$ maps each *observation history* (OH) $(o_i^1, \ldots, o_i^t) = \vec{o}_i^t \in \vec{\mathcal{O}}_i^t$ to an action, e.g., $\pi_i(\vec{o}_i^t) = a_i^t$. The number of such policies is $|\mathcal{A}_i|^{(|\mathcal{O}_i|^h - 1)/(|\mathcal{O}_i| - 1)}$ and the number of joint policies is therefore

$$O\Big(|\mathcal{A}_*|^{\frac{n|\mathcal{O}_*|^h - 1}{|\mathcal{O}_*| - 1}}\Big), \tag{2.1}$$

where $\mathcal{A}_*$ and $\mathcal{O}_*$ denote the largest individual action and observation sets. The quality of a particular joint policy is expressed by the expected cumulative reward it induces, also referred to as its value.

**Definition 2.** The *value* $V(\boldsymbol{\pi})$ of a joint policy $\boldsymbol{\pi}$ is

$$V(\boldsymbol{\pi}) \triangleq \mathbf{E}\Big[\sum_{t=0}^{h-1} R(s^t, \boldsymbol{a}^t) \Big| \boldsymbol{\pi}, \boldsymbol{b}^0\Big], \tag{2.2}$$

where the expectation is over sequences of states, actions and observations.

The planning problem for a Dec-POMDP is to find an optimal joint policy $\boldsymbol{\pi}^*$, i.e., a joint policy that maximizes the value: $\boldsymbol{\pi}^* = \arg\max_{\boldsymbol{\pi}} V(\boldsymbol{\pi})$.

Because an individual policy $\pi_i$ depends only on the local information $\vec{o}_i$ available to an agent, the on-line execution phase is truly decentralized: no communication takes place other than that modeled via actions and observations. The planning itself however, may take place in an off-line phase and be centralized. This is the scenario that we consider in this article. For a more detailed introduction to Dec-POMDPs see, e.g., the work of Seuken and Zilberstein (2008) and Oliehoek (2012).

### 2.1 Heuristic Search Methods

In recent years, numerous Dec-POMDP solution methods have been proposed. Most of these methods fall into one of two categories: *dynamic programming* and *heuristic search* methods. Dynamic programming methods take a backwards or 'bottom-up' perspective by first considering policies for the last time step $t = h - 1$ and using them to construct policies for stage $t = h - 2$, etc. In contrast, heuristic search methods take a forward or 'top-down' perspective by first constructing plans for $t = 0$ and extending them to later stages.

In this article, we focus on the heuristic search approach that has shown state-of-the-art results. As we make clear in this section, this method can be interpreted as searching over a tree of *collaborative Bayesian games (CBGs).* These CBGs provide a convenient abstraction layer that facilitates the explanation of the techniques introduced in this article.

This section provides some concise background on heuristic search methods. For a more detailed description, see the work of Oliehoek, Spaan, and Vlassis (2008). For a further description of dynamic programming methods and their relationship to heuristic search methods, see the work of Oliehoek (2012).

#### 2.1.1 Multiagent A*

Szer, Charpillet, and Zilberstein (2005) introduced a heuristically guided policy search method called *multiagent A* (*MAA**).* It performs an A* search over partially specified joint policies, pruning joint policies that are guaranteed to be worse than the best (fully specified) joint policy found so far. Oliehoek, Spaan, and Vlassis (2008) generalized the algorithm by making explicit the expand and selection operators performed in the heuristic search. The resulting algorithm, *generalized MAA* (GMAA*)* offers a unified perspective of MAA* and the *forward sweep policy computation* method (Emery-Montemerlo, 2005), which differ in how they implement GMAA*'s expand operator: forward sweep policy computation solves (i.e., finds the best policy for) collaborative Bayesian games, while MAA* finds *all* policies for those collaborative Bayesian games, as we describe in Section 2.1.2.

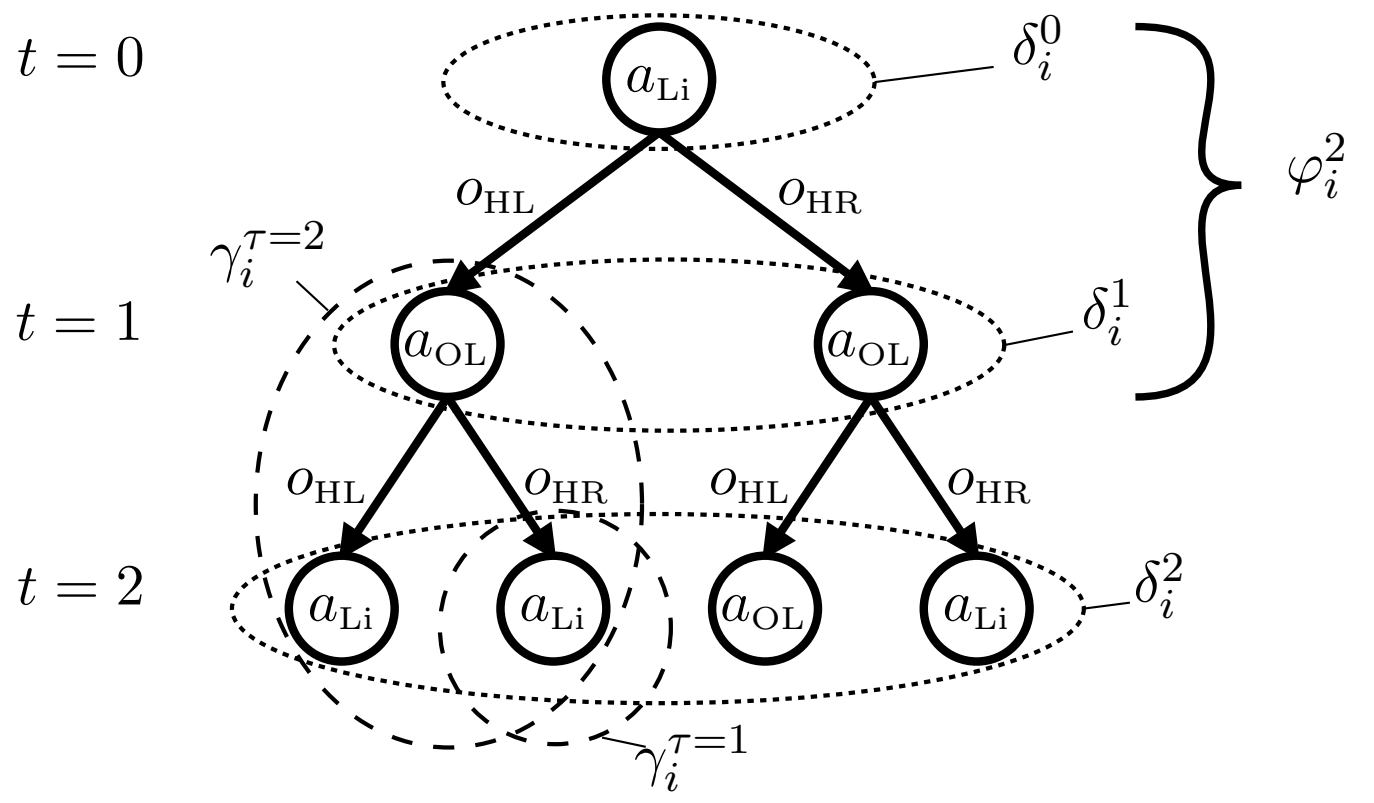

Figure 2: An arbitrary policy for the Dec-Tiger problem. The figure illustrates the different types of partial policies used in this paper. The shown past policy $\varphi_i^2$ consists of two decision rules $\delta_i^0$, $\delta_i^1$. Also shown are two sub-tree policies $\gamma_i^{\tau=1}$, $\gamma_i^{\tau=2}$ (introduced in Section 3.1.2).

The GMAA* algorithm considers joint policies that are partially specified with respect to time. These partially specified policies can be formalized as follows.

**Definition 3.** A *decision rule* $\delta_i^t$ for agent $i$'s decision for stage $t$ is a mapping from action-observation histories for stage $t$ to actions $\delta_i^t : \vec{\Theta}_i^t \to \mathcal{A}_i$.

In this article, we consider only deterministic policies. Since such policies need to condition their actions only on observation histories, they are made up of decision rules that map length-$t$ observation histories to actions: $\delta_i^t : \vec{\mathcal{O}}_i^t \to \mathcal{A}_i$. A joint decision rule $\boldsymbol{\delta}^t = \langle \delta_1^t, \ldots, \delta_n^t \rangle$ specifies a decision rule for each agent. Fig. 2 illustrates this concept, as well as that of a past policy, which we introduce shortly. As discussed below, decision rules allow partial policies to be defined and play a crucial role in GMAA* and the algorithms developed in this article.

**Definition 4.** A partial or *past policy for stage* $t$, $\varphi_i^t$, specifies the part of agent $i$'s policy that relates to stages $t' < t$. That is, it specifies the decision rules for the first $t$ stages: $\varphi_i^t = (\delta_i^0, \delta_i^1, \ldots, \delta_i^{t-1})$. A past policy for stage $h$ is just a regular, or *fully specified*, policy $\varphi_i^h = \pi_i$. A past *joint* policy $\boldsymbol{\varphi}^t = (\boldsymbol{\delta}^0, \boldsymbol{\delta}^1, \ldots, \boldsymbol{\delta}^{t-1})$ specifies joint decision rules for the first $t$ stages.

GMAA* performs a heuristic search over such partial joint policies $\boldsymbol{\varphi}^t$ by constructing a search tree as illustrated in Fig. 3a. Each node $q = \langle \boldsymbol{\varphi}^t, \hat{v} \rangle$ in the search tree specifies a past joint policy $\boldsymbol{\varphi}^t$ and heuristic value $\hat{v}$. This heuristic value $\hat{v}$ of the node represents an optimistic estimate of the past joint policy $\widehat{V}(\boldsymbol{\varphi}^t)$, which can be computed via

$$\widehat{V}(\boldsymbol{\varphi}^t) = V^{0 \ldots t-1}(\boldsymbol{\varphi}^t) + H^{t \ldots h-1}(\boldsymbol{\varphi}^t), \tag{2.3}$$

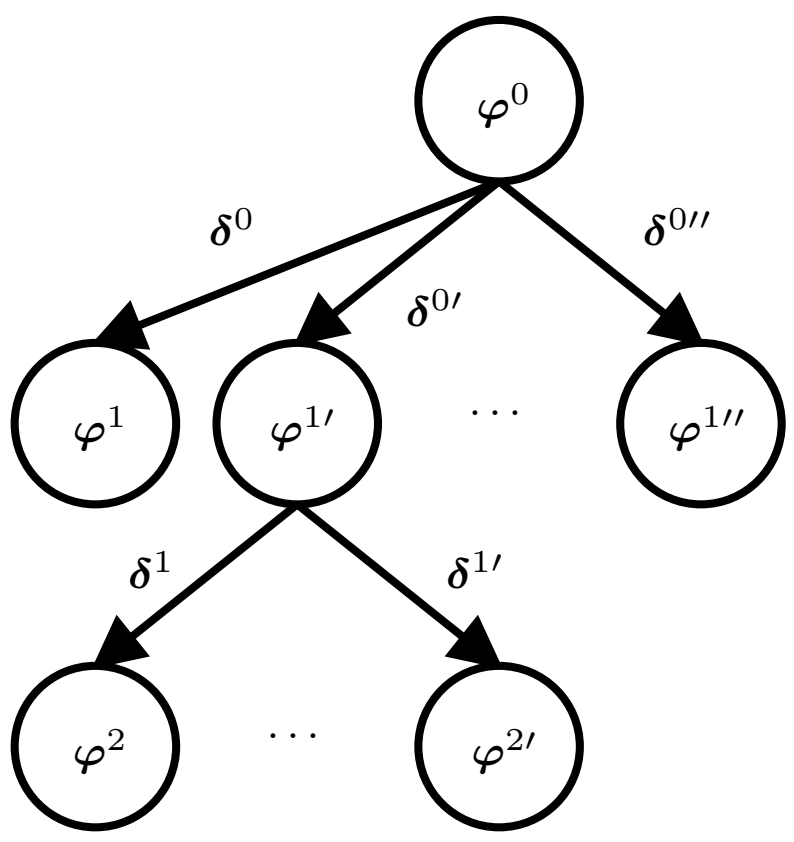

(a) The MAA* perspective.

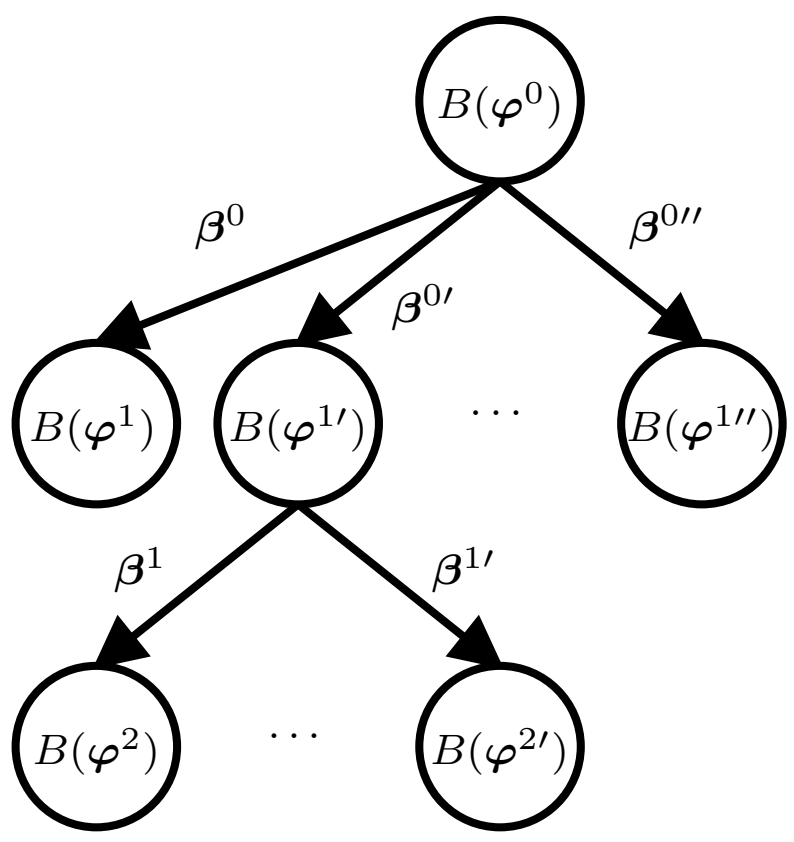

(b) The CBG perspective.

Figure 3: Generalized MAA*. Associated with every node is a heuristic value. The search trees for the two perspectives shown are equivalent under certain assumptions on the heuristic, as explained in Section 2.2.

where $H^{t\ldots h-1}$ is a heuristic value for the remaining $h-t$ stages and $V^{0\ldots t-1}(\boldsymbol{\varphi}^t)$ is the actual expected reward $\boldsymbol{\varphi}^t$ achieves over the first $t$ stages (for its definition, see Appendix A.3). Clearly, when $H^{t\ldots h-1}$ is an *admissible heuristic*—a guaranteed overestimation—so is $\widehat{V}(\boldsymbol{\varphi}^t)$.[5]

Algorithm 1 illustrates GMAA*. It starts by creating a node $q^0$ for a completely unspecified joint policy $\boldsymbol{\varphi}^0$ and placing it in an open list L. Then, it selects nodes (Algorithm 2) and expands them (Algorithm 3), repeating this process until it is certain that it has found the optimal joint policy.

The `Select` operator returns the highest ranked node, as defined by the following comparison operator.

**Definition 5.** The *node comparison operator* $<$ is defined for two nodes $q = \langle \boldsymbol{\varphi}^t, \hat{v} \rangle, q' = \langle \boldsymbol{\varphi}^{t\prime}, \hat{v}' \rangle$ as follows:

$$q < q' = \begin{cases} \hat{v} < \hat{v}' & \text{, if } \hat{v} \neq \hat{v}' \\ \text{depth}(q) < \text{depth}(q') & \text{, otherwise if depth}(q) \neq \text{depth}(q') \\ \boldsymbol{\varphi}^t < \boldsymbol{\varphi}^{t\prime} & \text{, otherwise.} \end{cases} \quad (2.4)$$

That is, the comparison operator first compares the heuristic values. If those are equal, it compares the depth of the nodes. Finally, if nodes have equal value and equal depth, it lexically compares the past joint policies. This ranking leads to A* behavior (i.e., selecting the node from the open list with the highest heuristic value) of GMAA*, as well as guaranteeing the same selection order in our incremental expansion technique (introduced in Section 4). Ranking nodes with greater depth higher in case of equal heuristic value helps find tight lower bounds early by first expanding deeper nodes (Szer et al., 2005) and is also useful in incremental expansion.

5. More formally, $H$ should not underestimate the value. Note that, unlike classical A* applications such as path planning–in which an admissible heuristic should not overestimate–in our setting we maximize reward, rather than minimize cost.

**Algorithm 1** Generalized multiagent A*.

**Input:** a Dec-POMDP, an admissible heuristic $H$, an empty open list L
**Output:** an optimal joint policy $\pi^*$
1: $\underline{\mathrm{v}}^{GMAA} \leftarrow -\infty$
2: $q^0 \leftarrow \langle \boldsymbol{\varphi}^0 = (), \hat{v} = +\infty \rangle$
3: L.insert($q^0$)
4: **repeat**
5:  $q \leftarrow$ `Select`(L)
6:  $\mathcal{Q}_{\texttt{Expand}} \leftarrow$ `Expand`$(q, H)$
7:  **if** depth$(q) = h - 1$ **then**
8:   { $\mathcal{Q}_{\texttt{Expand}}$ contains fully specified joint policies, we only are interested in the best one }
9:   $\langle \boldsymbol{\pi}, v \rangle \leftarrow$ `BestJointPolicyAndValue`$(\mathcal{Q}_{\texttt{Expand}})$
10:   **if** $v > \underline{\mathrm{v}}^{GMAA}$ **then**
11:    $\boldsymbol{\pi}^* \leftarrow \boldsymbol{\pi}$ {found a new best joint policy}
12:    $\underline{\mathrm{v}}^{GMAA} \leftarrow v$
13:    L.Prune($\underline{\mathrm{v}}^{GMAA}$) {(optionally) prune the open list}
14:   **end if**
15:  **else**
16:   L.insert($\{q' \in \mathcal{Q}_{\texttt{Expand}} \mid q'.\hat{v} > \underline{\mathrm{v}}^{GMAA}\}$) {add expanded children to open list}
17:  **end if**
18:  `PostProcessNode`$(q, \mathrm{L})$
19: **until** L is empty
20: **return** $\boldsymbol{\pi}^*$

**Algorithm 2** `Select`(L): Return the highest ranked node from the open list.

**Input:** open list L, total order on nodes $<$
**Output:** the highest ranked node $q^*$
$q^* \leftarrow q \in \mathrm{L}$ s.t. $\forall_{q' \in \mathrm{L}} \ (q' \neq q \implies q' < q)$
**return** $q^*$

The `Expand` operator constructs $\mathcal{Q}_{\texttt{Expand}}$, the set of all child nodes. That is, given a node that contains partial joint policy $\boldsymbol{\varphi}^t = (\boldsymbol{\delta}^0, \boldsymbol{\delta}^1, \ldots, \boldsymbol{\delta}^{t-1})$, it constructs $\Phi^{t+1}$, the set of all $\boldsymbol{\varphi}^{t+1} = (\boldsymbol{\delta}^0, \boldsymbol{\delta}^1, \ldots, \boldsymbol{\delta}^{t-1}, \boldsymbol{\delta}^t)$, by appending all possible joint decision rules $\boldsymbol{\delta}^t$ for the next time step $t$. For all these $\boldsymbol{\varphi}^{t+1}$, a heuristic value is computed and a node is constructed.

After expansion, the algorithm checks (line 7) if the expansion resulted in fully specified joint policies. If not, all children with sufficient heuristic value are placed in the open list

**Algorithm 3** `Expand`$(q, H)$. The expand operator of plain MAA*.

**Input:** $q = \langle \boldsymbol{\varphi}^t, \hat{v} \rangle$ the search node to expand, $H$ the admissible heuristic.
**Output:** $\mathcal{Q}_{\texttt{Expand}}$ the set containing all expanded child nodes.
$\mathcal{Q}_{\texttt{Expand}} \leftarrow \{\}$
$\Phi^{t+1} \leftarrow \{\boldsymbol{\varphi}^{t+1} \mid \boldsymbol{\varphi}^{t+1} = (\boldsymbol{\varphi}^t, \boldsymbol{\delta}^t)\}$
**for** $\boldsymbol{\varphi}^{t+1} \in \Phi^{t+1}$ **do**
 $\widehat{V}(\boldsymbol{\varphi}^{t+1}) \leftarrow V^{0\ldots t}(\boldsymbol{\varphi}^{t+1}) + H(\boldsymbol{\varphi}^{t+1})$
 $q' \leftarrow \langle \boldsymbol{\varphi}^{t+1}, \widehat{V}(\boldsymbol{\varphi}^{t+1}) \rangle$ {create child node}
 $\mathcal{Q}_{\texttt{Expand}}$.Insert($q'$)
**end for**
**return** $\mathcal{Q}_{\texttt{Expand}}$

---

**Algorithm 4** `PostProcessNode`($q$,L)

---

**Input:** $q$ the expanded parent node, L the open list.
**Output:** the expanded node is removed.
1: L.Pop($q$)

---

(line 16). If the children are fully specified, `BestJointPolicyAndValue` returns only the best joint policy (and its value) from $\mathcal{Q}_{\texttt{Expand}}$ (see Algorithm 12 in Appendix A.1 for details of `BestJointPolicyAndValue`). GMAA* also maintains a lower bound $\underline{v}^{GMAA}$ which corresponds to the actual value of the best fully-specified joint policy found so far. If the newly found joint policy has a higher value this lower bound is updated (lines 11 and 12). Also, any nodes for partial joint policies $\boldsymbol{\varphi}^{t+1}$ with an upper bound that is lower than the best solution so far, $\widehat{V}(\boldsymbol{\varphi}^{t+1}) < \underline{v}^{GMAA}$, can be pruned (line 13). This pruning takes additional time, but can save memory. Finally, `PostProcessNode` simply removes the parent node from the open list (this procedure is augmented for incremental expansion in Section 4). The search ends when the list becomes empty, at which point an optimal joint policy has been found.

GMAA* is *complete*, i.e., it will search until it finds a solution. Therefore, in theory, GMAA* is guaranteed to eventually produce an optimal joint policy (Szer et al., 2005). However, in practice, this is often infeasible for larger problems. A major source of complexity is the full expansion of a search node. The number of joint decision rules for stage $t$ that can form the children of a node at depth $t$ in the search tree[6] is

$$O\Big(|\mathcal{A}_*|^{n(|\mathcal{O}_*|^t)}\Big), \tag{2.5}$$

which is doubly exponential in $t$. Comparing (2.1) with (2.5), we see that the worst case complexity of expanding a node for the deepest level in the tree $t = h - 1$ is comparable to that of brute force search for the entire Dec-POMDP. Consequently, Seuken and Zilberstein (2008) conclude that MAA* "can at best solve problems whose horizon is only 1 greater than those that can already be solved by naïve brute force search."

#### 2.1.2 The Bayesian Game Perspective

GMAA* makes it possible to interpret MAA* as the solution of a collection of *collaborative Bayesian games (CBGs)*. We employ this approach throughout this article, as it facilitates the improvements to GMAA* that we introduce, each of which results in significant advances in the state-of-the-art in Dec-POMDP solutions.

A *Bayesian game (BG)* models a one-shot interaction between a number of agents. It is an extension of the well-known strategic game (also known as a normal form game) in which each agent holds some private information (Osborne & Rubinstein, 1994). A CBG is a BG in which the agents receive identical payoffs. In the Bayesian game perspective, each node $q$ in the GMAA* search tree, along with its corresponding partial joint policy $\boldsymbol{\varphi}^t$, defines a CBG (Oliehoek, Spaan, & Vlassis, 2008). That is, given state distribution $\boldsymbol{b}^0$, for each $\boldsymbol{\varphi}^t$, it is possible to construct a CBG $B(\boldsymbol{b}^0,\boldsymbol{\varphi}^t)$ that represents the decision-making problem for stage $t$ given that $\boldsymbol{\varphi}^t$ was followed for the first $t$ stages starting from $\boldsymbol{b}^0$. When it is clear what $\boldsymbol{b}^0$ is, we simply write $B(\boldsymbol{\varphi}^t)$.

6. We follow the convention that the root has depth 0.

**Definition 6.** A *collaborative Bayesian game (CBG)* $B(\boldsymbol{b}^0,\boldsymbol{\varphi}^t) = \langle \mathcal{D}, \boldsymbol{\mathcal{A}}, \boldsymbol{\Theta}, \Pr(\cdot), u\rangle$ modeling stage $t$ of a Dec-POMDP, given initial state distribution $\boldsymbol{b}^0$ and past joint policy $\boldsymbol{\varphi}^t$, consists of:

- $\mathcal{D}$, the set of agents $\{1 \dots n\}$,
- $\boldsymbol{\mathcal{A}}$, the set of joint actions,
- $\boldsymbol{\Theta}$, the set of their *joint types*, each of which specifies a *type* for each agent $\boldsymbol{\theta} = \langle \theta_1, \dots ,\theta_n\rangle$,
- $\Pr(\cdot)$, a probability distribution over joint types,
- $u$, a (heuristic) payoff function mapping joint type and action to a real number: $u(\boldsymbol{\theta},\boldsymbol{a})$.

In any Bayesian game, the type $\theta_i$ of an agent $i$ represents the private information it holds. For instance, in a Bayesian game modeling of a job recruitment scenario, the type of an agent may indicate whether that agent is a hard worker. In a CBG for a Dec-POMDP, an agent's private information is its individual AOH. Therefore, the type $\theta_i$ of an agent $i$ corresponds to $\vec{\theta}_i^t$, its history of actions and observations: $\theta_i \leftrightarrow \vec{\theta}_i^t$. Similarly, a joint type corresponds to a joint AOH: $\boldsymbol{\theta} \leftrightarrow \vec{\boldsymbol{\theta}}^t$.

Consequently, $u$ should provide a (heuristic) estimate for the long-term payoff of each $(\vec{\boldsymbol{\theta}}^t,\boldsymbol{a})$-pair. In other words, the payoff function corresponds to a heuristic Q-value: $u(\boldsymbol{\theta},\boldsymbol{a}) \leftrightarrow \widehat{Q}(\vec{\boldsymbol{\theta}}^t,\boldsymbol{a})$. We discuss how to compute such heuristics in Section 2.2. Given $\boldsymbol{\varphi}^t$, $\boldsymbol{b}^0$, and the correspondence of joint types and AOHs, the probability distribution over joint types is:

$$\Pr(\boldsymbol{\theta}) \triangleq \Pr(\vec{\boldsymbol{\theta}}^t|\boldsymbol{b}^0,\boldsymbol{\varphi}^t), \tag{2.6}$$

where the latter probability is the marginal of $\Pr(s,\vec{\boldsymbol{\theta}}^t|\boldsymbol{b}^0,\boldsymbol{\varphi}^t)$ as defined by (A.2) used in the computation of the value of a partial joint policy $V^{0\dots t-1}(\boldsymbol{\varphi}^t)$ in Appendix A.3. Note that due to the correspondence between types and AOHs, the size of a CBG $B(\boldsymbol{b}^0,\boldsymbol{\varphi}^t)$ for a stage $t$ is exponential in $t$.

In a CBG, each agent uses a Bayesian game policy $\beta_i$ that maps individual types to actions: $\beta_i(\theta_i) = a_i$. Because of the correspondence between types and AOHs, a (joint) policy for the CBG $\boldsymbol{\beta}$ corresponds to a (joint) decision rule: $\boldsymbol{\beta} \leftrightarrow \boldsymbol{\delta}^t$. In the remainder of this article, we assume deterministic past joint policies $\boldsymbol{\varphi}^t$, which implies that only one $\vec{\boldsymbol{\theta}}^t$ will have non-zero probability given the observation history $\vec{\mathbf{o}}^t$. Thus, $\boldsymbol{\beta}$ effectively maps observation histories to actions. The number of such $\boldsymbol{\beta}$ for $B(\boldsymbol{b}^0,\boldsymbol{\varphi}^t)$ is given by (2.5). The value of a joint CBG policy $\boldsymbol{\beta}$ for a CBG $B(\boldsymbol{b}^0,\boldsymbol{\varphi}^t)$ is:

$$\widehat{V}(\boldsymbol{\beta}) = \sum_{\vec{\boldsymbol{\theta}}^t} \Pr(\vec{\boldsymbol{\theta}}^t|\boldsymbol{b}^0,\boldsymbol{\varphi}^t)\widehat{Q}(\vec{\boldsymbol{\theta}}^t,\boldsymbol{\beta}(\vec{\boldsymbol{\theta}}^t)), \tag{2.7}$$

where $\boldsymbol{\beta}^t(\vec{\boldsymbol{\theta}}^t) = \langle \beta_i(\vec{\theta}_i^t)\rangle_{i=1\dots n}$ denotes the joint action that results from application of the individual CBG-policies to the individual AOH $\vec{\theta}_i^t$ specified by $\vec{\boldsymbol{\theta}}^t$.

*Example* 3. Consider a CBG for Dec-Tiger given the past joint policy $\boldsymbol{\varphi}^2$ that specifies to listen at the first two stages. At stage $t = 2$, each agent has four possible observation histories: $\vec{\mathcal{O}}_i^2 = \{(o_{\text{HL}},o_{\text{HL}}), (o_{\text{HL}},o_{\text{HR}}), (o_{\text{HR}},o_{\text{HL}}), (o_{\text{HR}},o_{\text{HR}})\}$ that correspond directly to its possible types. The probabilities of these joint types given $\boldsymbol{\varphi}^2$ are listed in Fig. 4a. Since the joint OHs together with $\boldsymbol{\varphi}^2$ determine the joint AOHs, they also correspond to so-called *joint beliefs*: probability distributions over states (introduced formally in Section 2.2). Fig. 4b shows these joint beliefs, which can serve as the basis for the heuristic payoff function (as further discussed in Section 2.2).

| $\vec{o}_1^2$ | $\vec{o}_2^2$: $(o_{\mathrm{HL}},o_{\mathrm{HL}})$ | $(o_{\mathrm{HL}},o_{\mathrm{HR}})$ | $(o_{\mathrm{HR}},o_{\mathrm{HL}})$ | $(o_{\mathrm{HR}},o_{\mathrm{HR}})$ |
|---|---|---|---|---|
| $(o_{\mathrm{HL}},o_{\mathrm{HL}})$ | 0.261 | 0.047 | 0.047 | 0.016 |
| $(o_{\mathrm{HL}},o_{\mathrm{HR}})$ | 0.047 | 0.016 | 0.016 | 0.047 |
| $(o_{\mathrm{HR}},o_{\mathrm{HL}})$ | 0.047 | 0.016 | 0.016 | 0.047 |
| $(o_{\mathrm{HR}},o_{\mathrm{HR}})$ | 0.016 | 0.047 | 0.047 | 0.261 |

(a) The joint type probabilities.

| $\vec{o}_1^2$ | $\vec{o}_2^2$: $(o_{\mathrm{HL}},o_{\mathrm{HL}})$ | $(o_{\mathrm{HL}},o_{\mathrm{HR}})$ | $(o_{\mathrm{HR}},o_{\mathrm{HL}})$ | $(o_{\mathrm{HR}},o_{\mathrm{HR}})$ |
|---|---|---|---|---|
| $(o_{\mathrm{HL}},o_{\mathrm{HL}})$ | 0.999 | 0.970 | 0.970 | 0.5 |
| $(o_{\mathrm{HL}},o_{\mathrm{HR}})$ | 0.970 | 0.5 | 0.5 | 0.030 |
| $(o_{\mathrm{HR}},o_{\mathrm{HL}})$ | 0.970 | 0.5 | 0.5 | 0.030 |
| $(o_{\mathrm{HR}},o_{\mathrm{HR}})$ | 0.5 | 0.030 | 0.030 | 0.001 |

(b) The induced joint beliefs. Listed is the probability $\Pr(s_l|\vec{\boldsymbol{\theta}}^2,\boldsymbol{b}^0)$ of the tiger being behind the left door.

Figure 4: Illustration for the DEC-TIGER problem with a past joint policy $\boldsymbol{\varphi}^2$ that specifies only listen actions for the first two stages.

**Algorithm 5** `Expand`-CBG$(q, H)$. The expand operator of GMAA* that makes use of CBGs.

**Input:** $q = \langle \boldsymbol{\varphi}^t, \hat{v} \rangle$ the search node to expand.
**Input:** $H$ the admissible heuristic that is of the form $\widehat{Q}(\vec{\boldsymbol{\theta}},\boldsymbol{a})$.
**Output:** $\mathcal{Q}_{\texttt{Expand}}$ the set containing all expanded child nodes.
1: $B(\boldsymbol{b}^0,\boldsymbol{\varphi}^t) \leftarrow$ ConstructBG$(\boldsymbol{b}^0,\boldsymbol{\varphi}^t,\widehat{Q})$ {as explained in Section 2.1.2}
2: $\mathcal{Q}_{\texttt{Expand}} \leftarrow$ `GenerateAllChildrenForCBG`$(B(\boldsymbol{b}^0,\boldsymbol{\varphi}^t))$
3: **return** $\mathcal{Q}_{\texttt{Expand}}$

A solution to the CBG is a $\boldsymbol{\beta}$ that maximizes (2.7). A CBG is equivalent to a team decision process and finding a solution is NP-complete (Tsitsiklis & Athans, 1985). However, in the Bayesian game perspective of GMAA*, illustrated in Fig. 3b, the issue of *solving* a CBG (i.e., finding the highest payoff $\boldsymbol{\beta}$) is not so relevant because we need to expand *all* $\boldsymbol{\beta}$. That is, the `Expand` operator enumerates all $\boldsymbol{\beta}$ and appends them to $\boldsymbol{\varphi}^t$ to form the set of extended joint policies

$$\Phi^{t+1} = \big\{(\boldsymbol{\varphi}^t,\boldsymbol{\beta}) \mid \boldsymbol{\beta} \text{ is a joint CBG policy of } B(\boldsymbol{b}^0,\boldsymbol{\varphi}^t)\big\}$$

and uses this set to construct $\mathcal{Q}_{\texttt{Expand}}$, the set of child nodes. The heuristic value of such a child node $q \in \mathcal{Q}_{\texttt{Expand}}$ that specifies $\boldsymbol{\varphi}^{t+1} = (\boldsymbol{\varphi}^t,\boldsymbol{\beta})$ is given by

$$\widehat{V}(\boldsymbol{\varphi}^{t+1}) = V^{0\dots t-1}(\boldsymbol{\varphi}^t) + \widehat{V}(\boldsymbol{\beta}). \tag{2.8}$$

The `Expand` operator that makes use of CBGs is summarized in Algorithm 5, which uses the `GenerateAllChildrenForCBG` subroutine (Algorithm 13 in Appendix A.1). Fig. 3b illustrates the Bayesian game perspective of GMAA*.

### 2.2 Heuristics

To perform heuristic search, GMAA* defines the heuristic value $\widehat{V}(\boldsymbol{\varphi}^t)$ using (2.3). In contrast, the Bayesian game perspective uses (2.8). These two formulations are equivalent when the heuristic $\widehat{Q}$ faithfully represents the expected immediate reward (Oliehoek, Spaan, & Vlassis, 2008). The consequence is that GMAA* via CBGs is complete (and thus finds optimal solutions) as stated by the following theorem.

**Theorem 1.** *When using a heuristic of the form*

$$\widehat{Q}(\vec{\boldsymbol{\theta}}^t,\boldsymbol{a}) = \mathbf{E}_{s^t}[R(s^t,\boldsymbol{a}) \mid \vec{\boldsymbol{\theta}}^t] + \mathbf{E}_{\vec{\boldsymbol{\theta}}^{t+1}}[\widehat{V}(\vec{\boldsymbol{\theta}}^{t+1}) \mid \vec{\boldsymbol{\theta}}^t, \boldsymbol{a}], \tag{2.9}$$

*where* $\widehat{V}(\vec{\boldsymbol{\theta}}^{t+1}) \geq Q_{\boldsymbol{\pi}^*}(\vec{\boldsymbol{\theta}}^{t+1}, \boldsymbol{\pi}^*(\vec{\boldsymbol{\theta}}^{t+1}))$ *is an overestimation of the value of an optimal joint policy* $\boldsymbol{\pi}^*$*,* GMAA* *via CBGs is complete.*

*Proof.* See appendix. □

In this theorem, $Q_{\boldsymbol{\pi}}(\vec{\boldsymbol{\theta}}^t,\boldsymbol{a})$ is the Q-value, i.e., the expected future cumulative reward of performing $\boldsymbol{a}$ from $\vec{\boldsymbol{\theta}}^t$ under joint policy $\boldsymbol{\pi}$ (Oliehoek, Spaan, & Vlassis, 2008). The expectation of the immediate reward will also be written as $R(\vec{\boldsymbol{\theta}}^t,\boldsymbol{a}) = \sum_{s\in\mathcal{S}} R(s,\boldsymbol{a}) \Pr(s|\vec{\boldsymbol{\theta}}^t,\boldsymbol{b}^0)$. It can be computed using $\Pr(s|\vec{\boldsymbol{\theta}}^t,\boldsymbol{b}^0)$, a quantity we refer to as the *joint belief* resulting from $\vec{\boldsymbol{\theta}}^t$ and that we also denote as $\boldsymbol{b}$. The joint belief itself can be computed via repeated application of Bayes' rule (Kaelbling, Littman, & Cassandra, 1998), or as the conditional of (A.2).

The rest of this subsection reviews several heuristics that have been used for GMAA*.

#### 2.2.1 $\mathrm{Q_{MDP}}$

One way to obtain an admissible heuristic $\widehat{Q}(\vec{\boldsymbol{\theta}},\boldsymbol{a})$ is to solve the underlying MDP, i.e., to assume the joint action is chosen by a single 'puppeteer' agent that can observe the true state. This approach, known as $\mathrm{Q_{MDP}}$ (Littman, Cassandra, & Kaelbling, 1995), uses the MDP value function $Q^{t,*}_{\mathrm{M}}(s^t,\boldsymbol{a})$, which can be computed using standard dynamic programming techniques (Puterman, 1994). In order to transform the $Q^{t,*}_{\mathrm{M}}(s^t,\boldsymbol{a})$-values to $\widehat{Q}_{\mathrm{M}}(\vec{\boldsymbol{\theta}}^t,\boldsymbol{a})$-values, we compute:

$$\widehat{Q}_{\mathrm{M}}(\vec{\boldsymbol{\theta}}^t,\boldsymbol{a}) = \sum_{s\in\mathcal{S}} Q^{t,*}_{\mathrm{M}}(s,\boldsymbol{a}) \Pr(s|\vec{\boldsymbol{\theta}}^t,\boldsymbol{b}^0). \tag{2.10}$$

Solving the underlying MDP has time complexity that is linear in $h$, which makes it, especially compared to the Dec-POMDP, easy to compute. In addition, it is only necessary to store a value for each $(s,\boldsymbol{a})$-pair, for each stage $t$. However, the bound it provides on the optimal Dec-POMDP $Q^*$-value function is loose (Oliehoek & Vlassis, 2007).

#### 2.2.2 $\mathrm{Q_{POMDP}}$

Similar to the underlying MDP, one can define the underlying POMDP of a Dec-POMDP, i.e., assuming the joint action is chosen by a single agent with access to the joint observation.[7]

7. Alternatively one can view this POMDP as a *multiagent POMDP* in which the agents can instantaneously broadcast their private observations.

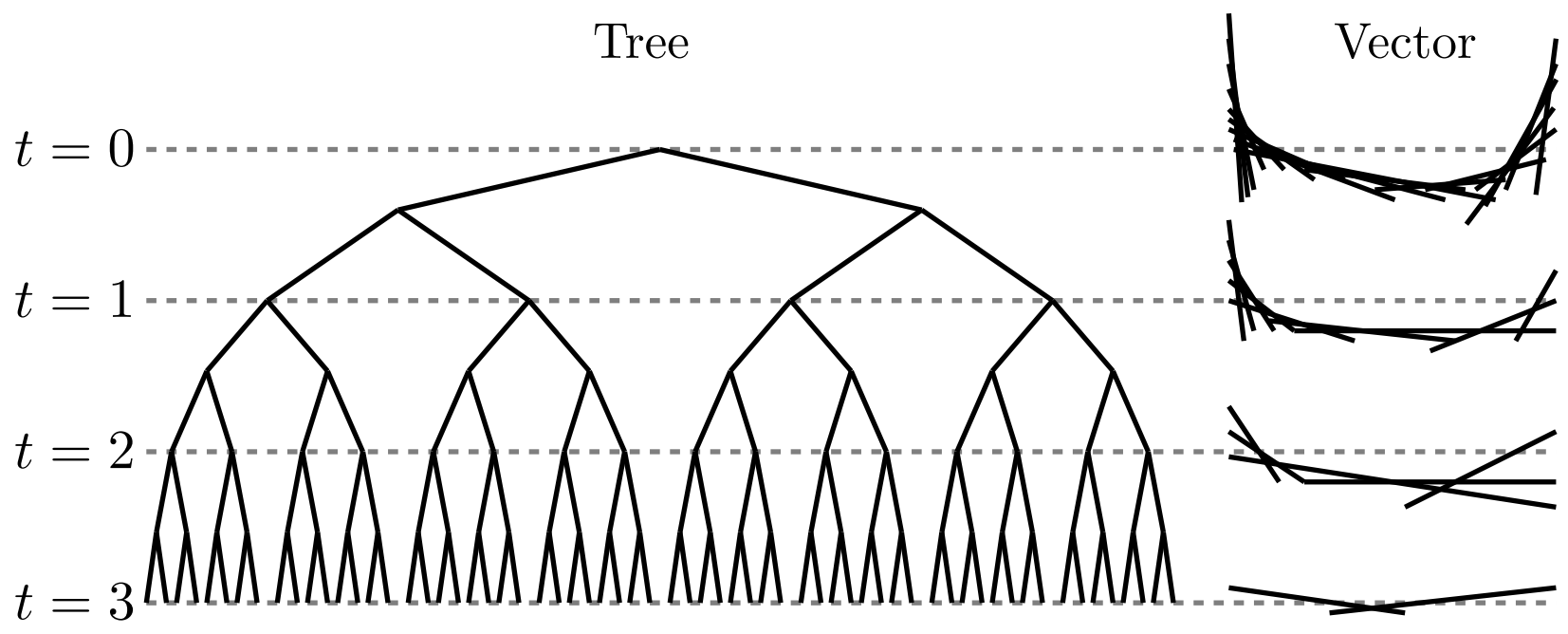

Figure 5: Visual comparison of tree and vector-based $Q$ representations.

The resulting solution can be used as a heuristic, called $\mathrm{Q_{POMDP}}$ (Szer et al., 2005; Roth, Simmons, & Veloso, 2005). The optimal $\mathrm{Q_{POMDP}}$ value function satisfies:

$$Q^*_{\mathrm{P}}(\boldsymbol{b}^t, \boldsymbol{a}) = R(\boldsymbol{b}^t,\boldsymbol{a}) + \sum_{\boldsymbol{o}^{t+1} \in \boldsymbol{\mathcal{O}}} P(\boldsymbol{o}^{t+1}|\boldsymbol{b}^t,\boldsymbol{a}) \max_{\boldsymbol{a}^{t+1}} Q^*_{\mathrm{P}}(\boldsymbol{b}^{t+1},\boldsymbol{a}^{t+1}), \tag{2.11}$$

where $\boldsymbol{b}^t$ is the joint belief, $R(\boldsymbol{b}^t,\boldsymbol{a}) = \sum_{s\in\mathcal{S}} R(s,\boldsymbol{a})\boldsymbol{b}^t(s)$ is the immediate reward, and $\boldsymbol{b}^{t+1}$ is the joint belief resulting from $\boldsymbol{b}^t$ by action $\boldsymbol{a}$ and joint observation $\boldsymbol{o}^{t+1}$. To use $\mathrm{Q_{POMDP}}$, for each $\vec{\boldsymbol{\theta}}^t$, we can directly use the value for the induced joint belief: $\widehat{Q}_{\mathrm{P}}(\vec{\boldsymbol{\theta}}^t,\boldsymbol{a}) \triangleq Q^t_{\mathrm{P}}(\boldsymbol{b}^{\vec{\boldsymbol{\theta}}^t},\boldsymbol{a})$.

There are two approaches to computing $\mathrm{Q_{POMDP}}$. One is to construct the 'belief MDP tree' of all joint beliefs, illustrated in Fig. 5 (left). Starting with $\boldsymbol{b}^0$ (corresponding to the empty joint AOH $\vec{\boldsymbol{\theta}}^0$), for each $\boldsymbol{a}$ and $\boldsymbol{o}$ we compute the resulting $\vec{\boldsymbol{\theta}}^1$ and corresponding belief $\boldsymbol{b}^{\vec{\boldsymbol{\theta}}^1}$ and continue recursively. Given this tree, it is possible to compute values for all the nodes by standard dynamic programming.

Another possibility is to apply vector-based POMDP techniques (see Fig. 5 (right)). The Q-value function for a stage $Q^t_{\mathrm{P}}(\boldsymbol{b},\boldsymbol{a})$ can be represented using a set of vectors for each joint action $\mathcal{V}^t = \{\mathcal{V}^t_1, \ldots, \mathcal{V}^t_{|\boldsymbol{\mathcal{A}}|}\}$ (Kaelbling et al., 1998). $Q^t_{\mathrm{P}}(\boldsymbol{b},\boldsymbol{a})$ is then defined as the maximum inner product:

$$Q^t_{\mathrm{P}}(\boldsymbol{b},\boldsymbol{a}) \triangleq \max_{v^t_{\boldsymbol{a}} \in \mathcal{V}^t_{\boldsymbol{a}}} \boldsymbol{b} \cdot v^t_{\boldsymbol{a}}.$$

Given $\mathcal{V}^{h-1}$, the vector representation of the last stage, we can compute $\mathcal{V}^{h-2}$, etc. In order to limit the growth of the number of vectors, dominated vectors can be pruned.

Since $\mathrm{Q_{MDP}}$ is an upper bound on the POMDP value function (Hauskrecht, 2000), $\mathrm{Q_{POMDP}}$ provides a tighter upper bound to $Q^*$ than $\mathrm{Q_{MDP}}$. However, it is also more costly to compute and store: both the tree-based and the vector-based approach may need to store a number of values exponential in $h$.

### 2.2.3 $Q_{BG}$

A third heuristic, called $Q_{BG}$, assumes that each agent in the team has access only to its individual observation but it can communicate with a 1-step delay.[8] We define $Q_{BG}$ as

$$Q^*_B(\vec{\boldsymbol{\theta}}^t,\boldsymbol{a}) = R(\vec{\boldsymbol{\theta}}^t,\boldsymbol{a}) + \max_{\boldsymbol{\beta}} \sum_{\boldsymbol{o}^{t+1}\in\mathcal{O}} \Pr(\boldsymbol{o}^{t+1}|\vec{\boldsymbol{\theta}}^t,\boldsymbol{a}) Q^*_B(\vec{\boldsymbol{\theta}}^{t+1},\boldsymbol{\beta}(\boldsymbol{o}^{t+1})), \tag{2.12}$$

where $\boldsymbol{\beta} = \langle\beta_1(o_1^{t+1}),...,\beta_n(o_n^{t+1})\rangle$ is a tuple of individual policies $\beta_i : \mathcal{O}_i \rightarrow \mathcal{A}_i$ for the CBG constructed for $\vec{\boldsymbol{\theta}}^t,\boldsymbol{a}$. Like $Q_{POMDP}$, $Q_{BG}$ can also be represented using vectors (Varaiya & Walrand, 1978; Hsu & Marcus, 1982; Oliehoek, Spaan, & Vlassis, 2008) and the same two manners of computation (tree and vector based) apply. It yields a tighter heuristic than $Q_{POMDP}$, but its computation has an additional exponential dependence on the maximum number of individual observations (Oliehoek, Spaan, & Vlassis, 2008), which is particularly troubling for the vector-based computation, since it precludes effective application of incremental pruning (A. Cassandra, Littman, & Zhang, 1997). To overcome this problem, Oliehoek and Spaan (2012) introduce novel tree-based pruning methods.

## 3. Clustering

GMAA* solves Dec-POMDPs by repeatedly constructing CBGs and expanding all the joint BG policies $\boldsymbol{\beta}$ for them. However, the number of such $\boldsymbol{\beta}$ is equal to the number of regular MAA* child nodes given by (2.5) and thus grows doubly exponentially with the horizon $h$. In this section, we propose a new approach for improving scalability with respect to $h$ by clustering individual AOHs. This reduces the number of $\boldsymbol{\beta}$ and therefore the number of constructed child nodes in the GMAA* search tree.[9]

Previous research has also investigated such clustering: Emery-Montemerlo, Gordon, Schneider, and Thrun (2005) propose clustering types based on the profiles of the payoff functions of the CBGs. However, the resulting method is ad hoc. Even given bounds on the error of clustering two types in a CBG, no guarantees can be made about the quality of the Dec-POMDP solution, as the bound is with respect to a heuristic payoff function.

In contrast, we propose to cluster histories based on the probability these histories induce over histories of the other agents and over states. The critical advantage of this criterion, which we call *probabilistic equivalence (PE)*, is that the resulting clustering is *lossless*: the solution for the clustered CBG can be used to construct the solution for the original CBG and the values of the two CBGs are identical. Thus, the criterion allows for clustering of AOHs in CBGs that represent Dec-POMDPs while preserving optimality.[10]

In Section 3.1, we describe how histories in Dec-POMDPs can be clustered using the notions of probabilistic and best-response equivalence. This allows histories to be clustered

8. The name $Q_{BG}$ stems from the fact that such a 1-step delayed communication scenario can be modeled as a CBG. Note, however, that the CBGs used to compute $Q_{BG}$ are of a different form than the $B(\boldsymbol{b}^0,\boldsymbol{\varphi}^t)$ discussed in Section 2.1.2: in the latter, types correspond to length-$t$ (action-) observation histories; in the former, types correspond to length-1 observation histories.
9. While CBGs are not essential for clustering, they provide a convenient level of abstraction that simplifies exposition of our techniques. Moreover, this level of abstraction makes it possible to employ our results concerning CBGs outside the context of Dec-POMDPs.
10. The probabilistic equivalence criterion and lossless clustering were introduced by Oliehoek et al. (2009). This article presents a new, simpler proof of the optimality of clustering based on PE.

when it is rational to always choose the same action. In Section 3.2, we describe the application of these results to GMAA*. Section 3.3 introduces improved heuristic representations that allow for the computation over longer horizons.

### 3.1 Lossless Clustering in Dec-POMDPs

In this section, we discuss lossless clustering based on the notion of probabilistic equivalence. We show that this clustering is lossless by demonstrating that probabilistic equivalence implies *best response equivalence*, which describes the conditions that a rational agent will select the same action for two of its types. To prove this implication, we show that the best response depends only on the *multiagent belief* (i.e., the probability distribution over states and policies of the other agents), which is the same for two probabilistically equivalent histories. Relations to other equivalence notions are discussed in Section 6.

#### 3.1.1 Probabilistic Equivalence Criterion

We first introduce the probabilistic equivalence criterion, which can be used to decide whether two individual histories $\vec{\theta}_i^a$,$\vec{\theta}_i^b$ can be clustered without loss in value.

**Criterion 1** (Probabilistic Equivalence)**.** *Two AOHs* $\vec{\theta}_i^a$,$\vec{\theta}_i^b$ *for agent i are* probabilistically equivalent (PE)*, written* $PE(\vec{\theta}_i^a,\vec{\theta}_i^b)$*, when the following holds:*

$$\forall_{\vec{\boldsymbol{\theta}}_{\neq i}} \forall_s \quad \Pr(s,\vec{\boldsymbol{\theta}}_{\neq i}|\vec{\theta}_i^a) = \Pr(s,\vec{\boldsymbol{\theta}}_{\neq i}|\vec{\theta}_i^b). \tag{3.1}$$

These probabilities can be computed as the conditional of $\Pr(s,\vec{\boldsymbol{\theta}}^t|\boldsymbol{b}^0,\boldsymbol{\varphi}^t)$, defined by (A.2).

In subsections 3.1.2–3.1.4, we formally prove that PE is a sufficient criterion to guarantee that clustering is lossless. In the remainder of Section 3.1.1 we discuss some key properties of the PE criterion in order to build intuition.

Note that the criterion can be decomposed into the following two criteria:

$$\forall_{\vec{\boldsymbol{\theta}}_{\neq i}} \quad \Pr(\vec{\boldsymbol{\theta}}_{\neq i}|\vec{\theta}_i^a) = \Pr(\vec{\boldsymbol{\theta}}_{\neq i}|\vec{\theta}_i^b), \tag{3.2}$$

$$\forall_{\vec{\boldsymbol{\theta}}_{\neq i}} \forall_s \quad \Pr(s|\vec{\boldsymbol{\theta}}_{\neq i},\vec{\theta}_i^a) = \Pr(s|\vec{\boldsymbol{\theta}}_{\neq i},\vec{\theta}_i^b). \tag{3.3}$$

These criteria give a natural interpretation: the first says that the probability distribution over the other agents' AOHs must be identical for both $\vec{\theta}_i^a$ and $\vec{\theta}_i^b$. The second demands that the resulting joint beliefs are identical.

The above probabilities are not well defined without the initial state distribution $\boldsymbol{b}^0$ and past joint policy $\boldsymbol{\varphi}^t$. However, since we consider clustering of histories within a particular CBG (for some stage $t$) constructed for a particular $\boldsymbol{b}^0$,$\boldsymbol{\varphi}^t$, they are implicitly specified. Therefore we drop these arguments, clarifying the notation.

*Example* 4. In Example 3, the types ($o_{\text{HL}}$,$o_{\text{HR}}$) and ($o_{\text{HR}}$,$o_{\text{HL}}$) of each agent are PE. To see this, note that the rows (columns for the second agent) for these histories are identical in both Fig. 4a and Fig. 4b. Thus, they specify the same distribution over histories of the other agents (cf. equation (3.2)) and the induced joint beliefs are the same (cf. equation (3.3)).

Probabilistic equivalence has a convenient property that our algorithms exploit: if it holds for a particular pair of histories, then it will also hold for all *identical extensions* of those histories, i.e., it propagates forwards regardless of the policies of the other agents.

**Definition 7** (Identical extensions)**.** Given two AOHs $\vec{\theta}_i^{a,t},\vec{\theta}_i^{b,t}$, their respective extensions $\vec{\theta}_i^{a,t+1} = (\vec{\theta}_i^{a,t},a_i,o_i)$ and $\vec{\theta}_i^{b,t+1} = (\vec{\theta}_i^{b,t},a_i',o_i')$ are called *identical extensions* if and only if $a_i = a_i'$ and $o_i = o_i'$.

**Lemma 1** (Propagation of PE)**.** *Given $\vec{\theta}_i^{a,t},\vec{\theta}_i^{b,t}$ that are PE, regardless of the decision rule the other agents use ($\boldsymbol{\delta}_{\neq i}^t$), identical extensions are also PE:*

$$\forall_{a_i^t}\forall_{o_i^{t+1}}\forall_{\boldsymbol{\delta}_{\neq i}^t}\forall_{s^{t+1}}\forall_{\vec{\boldsymbol{\theta}}_{\neq i}^{t+1}} \quad \Pr(s^{t+1},\vec{\boldsymbol{\theta}}_{\neq i}^{t+1}|\vec{\theta}_i^{a,t},a_i^t,o_i^{t+1},\boldsymbol{\delta}_{\neq i}^t) = \Pr(s^{t+1},\vec{\boldsymbol{\theta}}_{\neq i}^{t+1}|\vec{\theta}_i^{b,t},a_i^t,o_i^{t+1},\boldsymbol{\delta}_{\neq i}^t) \quad (3.4)$$

*Proof.* The proof is listed in the appendix, but holds intuitively because if the probabilities described above were the same before, they will also be the same after taking the same action and seeing the same observation. □

Note that, while the probabilities defined in (3.1) superficially resemble beliefs used in POMDPs, they are substantially different. In a POMDP, the single agent can compute its individual belief using only its AOH. It can then use this belief to determine the value of any future policy, as it is a sufficient statistic of the history to predict the future rewards (Kaelbling et al., 1998; Bertsekas, 2005). Thus, it is trivial to show equivalence of AOHs that induce the same individual belief in a POMDP. Unfortunately, Dec-POMDPs are more problematic. The next section elaborates on this issue by discussing the relation to *multiagent beliefs.*

### 3.1.2 Sub-Tree Policies, Multiagent Beliefs and Expected Future Value

To describe the relationship between multiagent beliefs and probabilistic equivalence, we must first discuss the policies an agent may follow and their resulting values. We begin by introducing the concept of *sub-tree policies.* As illustrated in Fig. 2 (on page 456), a (deterministic) policy $\pi_i$ can be represented as a tree with nodes labeled using actions and edges labeled using observations: the root node corresponds to the first action taken, other nodes specify the action for the observation history encoded by the path from the root node. As such, it is possible to define sub-tree policies, $\gamma_i$, which correspond to sub-trees of agent $i$'s policy $\pi_i$ (also illustrated in Fig. 2). In particular, we write

$$\pi_i\Downarrow_{\vec{o}_i^t} = \gamma_i^{\tau=h-t} \quad (3.5)$$

for the sub-tree policy of $\pi_i$ corresponding to observation history $\vec{o}_i^t$ that specifies the actions for the last $\tau = h - t$ stages. We refer to $\Downarrow$ as the *policy consumption operator*, since it 'consumes' the part of the policy corresponding to $\vec{o}_i^t$. Similarly we write $\gamma_i^{\tau=k}\Downarrow_{\vec{o}_i^l} = \gamma_i^{\tau=k-l}$ (note that in (3.5), $\pi_i$ is just a $\tau = h$-steps-to-go sub-tree policy) and use similar notation, $\boldsymbol{\gamma}^{\tau=k}$, for joint sub-tree policies. For a more extensive treatment of these different forms of policy, we refer to the discussion by Oliehoek (2012).

Given these concepts, we can define the value of a $\tau = k$-stages-to-go joint policy starting from state $s$:

$$V(s,\boldsymbol{\gamma}^{\tau=k}) = R(s,\boldsymbol{a}) + \sum_{s'}\sum_{\boldsymbol{o}} \Pr(s',\boldsymbol{o}|s,\boldsymbol{a})V(s',\boldsymbol{\gamma}^{\tau=k}\Downarrow_{\boldsymbol{o}}). \quad (3.6)$$

Here, $\boldsymbol{a}$ is the joint action specified by the roots of the individual sub-tree policies specified by $\boldsymbol{\gamma}^{\tau=k}$ for stage $t = h - k$.

From this definition, it follows directly that the probability distribution over states $s$ and sub-tree policies over other agents $\boldsymbol{\gamma}_{\neq i}$ is sufficient to predict the value of a sub-tree policy $\gamma_i$. In fact, such a distribution is known as a *multiagent belief* $b_i(s,\boldsymbol{\gamma}_{\neq i})$ (Hansen, Bernstein, & Zilberstein, 2004). Its value is given by

$$V(b_i) = \max_{\gamma_i} \sum_s \sum_{\boldsymbol{\gamma}_{\neq i}} b_i(s,\boldsymbol{\gamma}_{\neq i}) V(s,\langle \gamma_i, \boldsymbol{\gamma}_{\neq i}\rangle), \tag{3.7}$$

and we refer to the maximizing $\gamma_i$ as agent $i$'s best response for $b_i$. This illustrates that a multiagent belief is a sufficient statistic: it contains sufficient information to predict the value of any sub-tree policy $\gamma_i$.

It is possible to connect action observation histories to multiagent beliefs by fixing the policies of the other agents. Given that the other agents will act according to a profile of policies $\boldsymbol{\pi}_{\neq i}$, agent $i$ has a multiagent belief at the first stage of the Dec-POMDP: $b_i(s,\boldsymbol{\pi}_{\neq i}) = \boldsymbol{b}^0(s)$. Moreover, agent $i$ can maintain such a multiagent belief during execution. As such, given $\boldsymbol{\pi}_{\neq i}$, each history $\vec{\theta}_i$ induces a multiagent belief, which we will write as $b_i(s,\boldsymbol{\gamma}_{\neq i}|\vec{\theta}_i, \boldsymbol{\pi}_{\neq i})$ to make the dependence on $\vec{\theta}_i, \boldsymbol{\pi}_{\neq i}$ explicit. The multiagent belief for a history is defined as

$$b_i(s,\boldsymbol{\gamma}_{\neq i}|\vec{\theta}_i, \boldsymbol{\pi}_{\neq i}) \triangleq \Pr(s,\boldsymbol{\gamma}_{\neq i}|\vec{\theta}_i, \boldsymbol{b}^0, \boldsymbol{\pi}_{\neq i}), \tag{3.8}$$

and induces a best response via (3.7):

$$BR(\vec{\theta}_i|\boldsymbol{\pi}_{\neq i}) \triangleq \arg\max_{\gamma_i} \sum_s \sum_{\boldsymbol{\gamma}_{\neq i}} b_i(s,\boldsymbol{\gamma}_{\neq i}|\vec{\theta}_i, \boldsymbol{\pi}_{\neq i}) V(s,\boldsymbol{\gamma}_{\neq i},\gamma_i). \tag{3.9}$$

From this we can conclude that two AOHs $\vec{\theta}_i^a$,$\vec{\theta}_i^b$ can be clustered together if they induce the same multiagent belief.

However, this notion of multiagent belief is clearly quite different from the distributions used in our notion of PE. In particular, to establish whether two AOHs induce the same multiagent belief, we need a full specification of $\boldsymbol{\pi}_{\neq i}$. Nevertheless, we show that two AOHs that are PE are also best response equivalent and that we can therefore cluster them. The crux is that we can show that, if Criterion 1 is satisfied, the AOHs will always induce the same multiagent beliefs for *any* $\boldsymbol{\pi}_{\neq i}$ (consistent with the current past joint policy $\boldsymbol{\varphi}_{\neq i}$).

#### 3.1.3 Best-Response Equivalence Allows Lossless Clustering of Histories

We can now relate probabilistic equivalence and the multiagent belief as follows.

**Lemma 2** (PE implies multiagent belief equivalence)**.** *For any $\boldsymbol{\pi}_{\neq i}$, probabilistic equivalence implies multiagent belief equivalence:*

$$PE(\vec{\theta}_i^a,\vec{\theta}_i^b) \quad \Rightarrow \quad \forall_{s,\boldsymbol{\gamma}_{\neq i}} \Big( b_i(s,\boldsymbol{\gamma}_{\neq i}|\vec{\theta}_i^a, \boldsymbol{\pi}_{\neq i}) = b_i(s,\boldsymbol{\gamma}_{\neq i}|\vec{\theta}_i^b, \boldsymbol{\pi}_{\neq i}) \Big) \tag{3.10}$$

*Proof.* See appendix. □

This lemma shows that if two AOHs are PE, they produce the same multiagent belief. Intuitively, this gives us a justification to cluster such AOHs together: since a multiagent belief is a sufficient statistic we should act the same when we have the same multiagent belief, but since Lemma 2 shows that $\vec{\theta}_i^a$,$\vec{\theta}_i^b$ induces the same multiagent beliefs for any $\boldsymbol{\pi}_{\neq i}$ when they are PE, we can conclude that we will always act the same in those histories. Formally, we prove that $\vec{\theta}_i^a$,$\vec{\theta}_i^b$ are *best-response equivalent* if they are PE.

**Theorem 2** (PE implies best-response equivalence)**.** *Probabilistic equivalence implies best-response equivalence. That is*

$$PE(\vec{\theta}_i^a,\vec{\theta}_i^b) \quad \Rightarrow \quad \forall_{\boldsymbol{\pi}_{\neq i}} \left( BR(\vec{\theta}_i^a|\boldsymbol{\pi}_{\neq i}) = BR(\vec{\theta}_i^b|\boldsymbol{\pi}_{\neq i}) \right)$$

*Proof.* Assume any arbitrary $\boldsymbol{\pi}_{\neq i}$, then

$$\begin{aligned} BR(\vec{\theta}_i^a|\boldsymbol{\pi}_{\neq i}) &= \arg\max_{\gamma_i} \sum_s \sum_{\boldsymbol{\gamma}_{\neq i}} b_i(s,\boldsymbol{\gamma}_{\neq i}|\vec{\theta}_i^a) V(s,\boldsymbol{\gamma}_{\neq i},\gamma_i) \\ &= \arg\max_{\gamma_i} \sum_s \sum_{\boldsymbol{\gamma}_{\neq i}} b_i(s,\boldsymbol{\gamma}_{\neq i}|\vec{\theta}_i^b) V(s,\boldsymbol{\gamma}_{\neq i},\gamma_i) = BR(\vec{\theta}_i^b|\boldsymbol{\pi}_{\neq i}), \end{aligned}$$

where Lemma 2 is employed to assert the equality of $b_i(\cdot|\vec{\theta}_i^a)$ and $b_i(\cdot|\vec{\theta}_i^b)$. □

This theorem is key because it demonstrates that when two AOHs $\vec{\theta}_i^a,\vec{\theta}_i^b$ of an agent are PE, then that agent need not discriminate between them *now or in the future.* Thus, when searching the space of joint policies, we can restrict our search to those that assign the same sub-tree policy $\gamma_i$ to $\vec{\theta}_i^a$ and $\vec{\theta}_i^b$. As such, it directly provides intuition as to why lossless clustering is possible. Formally, we define the clustered joint policy space as follows.

**Definition 8** (Clustered joint policy space)**.** Let $\boldsymbol{\Pi}_C \subseteq \boldsymbol{\Pi}$ be the subset of joint policies that is clustered: i.e., each $\pi_i$ that is part of a $\boldsymbol{\pi} \in \boldsymbol{\Pi}_C$ assigns the same sub-tree policy to action observation histories that are probabilistically equivalent.

**Corollary 1** (Existence of an optimal clustered joint policy)**.** *There exists an optimal joint policy in the clustered joint policy space:*

$$\max_{\boldsymbol{\pi} \in \boldsymbol{\Pi}_C} V(\boldsymbol{\pi}) = \max_{\boldsymbol{\pi} \in \boldsymbol{\Pi}} V(\boldsymbol{\pi}) \tag{3.11}$$

*Proof.* It is clear that the left hand side of (3.11) is upper bounded by the right hand side, since $\boldsymbol{\Pi}_C \subseteq \boldsymbol{\Pi}$. Now suppose that $\boldsymbol{\pi}^* = \arg\max_{\boldsymbol{\pi} \in \boldsymbol{\Pi}} V(\boldsymbol{\pi})$ has strictly higher value than the best clustered joint policy. For at least one agent $i$ and one pair of PE histories $\vec{\theta}_i^a$, $\vec{\theta}_i^b$, $\boldsymbol{\pi}^*$ must assign different sub-tree policies $\gamma_i^a \neq \gamma_i^b$ (otherwise $\boldsymbol{\pi}^*$ would be clustered). Without loss of generality we assume that there is only one such pair. It follows directly from Theorem 2 that from this policy we can construct a clustered policy $\boldsymbol{\pi}_C \in \boldsymbol{\Pi}_C$ (by assigning either $\gamma_i^a$ or $\gamma_i^b$ to both $\vec{\theta}_i^a$, $\vec{\theta}_i^b$) that is guaranteed to have value no less than $\boldsymbol{\pi}^*$, thereby contradicting the assumption that $\boldsymbol{\pi}^*$ has strictly higher value than the best clustered joint policy. □

This formally proves that we can restrict our search to $\boldsymbol{\Pi}_C$, the space of clustered joint policies, without sacrificing optimality.

### 3.1.4 Clustering with Commitment in CBGs

Though it is now clear that two AOHs that are PE can be clustered, making this result operational requires an additional step. To this end, we use the abstraction layer provided by Bayesian games. Recall that in the CBG for a stage, the AOHs correspond to types.

Therefore, we want to cluster these types in the CBG. To accomplish the clustering of two types $\theta_i^a,\theta_i^b$, we introduce a new type $\theta_i^c$ to replace them, by defining:

$$\forall_{\boldsymbol{\theta}_{\neq i}} \ \Pr(\theta_i^c,\boldsymbol{\theta}_{\neq i}) \triangleq \Pr(\theta_i^a,\boldsymbol{\theta}_{\neq i}) + \Pr(\theta_i^b,\boldsymbol{\theta}_{\neq i}) \tag{3.12}$$

$$\forall_j \forall_{\boldsymbol{a}} \quad u(\langle\theta_i^c,\boldsymbol{\theta}_{\neq i}\rangle,\boldsymbol{a}) \triangleq \frac{\Pr(\theta_i^a,\boldsymbol{\theta}_{\neq i})u(\langle\theta_i^a,\boldsymbol{\theta}_{\neq i}\rangle,\boldsymbol{a}) + \Pr(\theta_i^b,\boldsymbol{\theta}_{\neq i})u(\langle\theta_i^b,\boldsymbol{\theta}_{\neq i}\rangle,\boldsymbol{a})}{\Pr(\theta_i^a,\boldsymbol{\theta}_{\neq i}) + \Pr(\theta_i^b,\boldsymbol{\theta}_{\neq i})}. \tag{3.13}$$

**Theorem 3** (Reduction through commitment)**.** *Given that agent $i$ in collaborative Bayesian game $B$ is committed to selecting a policy that assigns the same action for two of its types $\theta_i^a,\theta_i^b$, i.e., to selecting a policy $\beta_i$ such that $\beta_i(\theta_i^a) = \beta_i(\theta_i^b)$, the CBG can be reduced without loss in value for any agents. That is, the result is a new CBG $B'$ in which agent $i$ employs a policy $\beta_i'$ that reflects the clustering and whose expected payoff is the same as in the original CBG: $V^{B'}(\beta_i',\boldsymbol{\beta}_{\neq i}) = V^B(\beta_i,\boldsymbol{\beta}_{\neq i})$.*

*Proof.* See appendix. □

This theorem shows that, given that agent $i$ is committed to taking the same action for its types $\theta_i^a,\theta_i^b$, we can reduce the collaborative Bayesian game $B$ to a smaller one $B'$ and translate the joint CBG-policy $\boldsymbol{\beta}'$ found for $B'$ back to a joint CBG-policy $\boldsymbol{\beta}$ in $B$. This does not necessarily mean that $\boldsymbol{\beta} = \left(\beta_i,\boldsymbol{\beta}_{\neq i}\right)$ is also a solution for $B$, because the best-response of agent $i$ against $\boldsymbol{\beta}_{\neq i}$ may not select the same action for $\theta_i^a,\theta_i^b$. Rather $\beta_i$ is the best-response against $\boldsymbol{\beta}_{\neq i}$ *given that the same action needs to be taken for* $\theta_i^a,\theta_i^b$.[11]

Even though Theorem 3 only gives a conditional statement that depends on an agent being committed to select the same action for two of its types, the previous subsection discussed when a rational agent can make such a commitment. Combining these results gives the following corollary.

**Corollary 2** (Lossless Clustering with PE)**.** *Probabilistically equivalent histories $\vec{\theta}_i^a,\vec{\theta}_i^b$ can be clustered without loss in heuristic value by merging them into a single type in a CBG.*

*Proof.* Theorem 3 shows that, given that an agent $i$ is committed to take the same action for two of its types, those types can be clustered without loss in value. Since $\vec{\theta}_i^a,\vec{\theta}_i^b$ are PE, they are best-response equivalent, which means that the agent is committed to use the same sub-tree policy $\gamma_i$ and hence the same action $a_i$. Therefore we can directly apply clustering without loss in expected payoff, which in a CBG for a stage of a Dec-POMDP means no loss in expected heuristic value as given by (2.7). □

Intuitively, the maximizing action is the same for $\vec{\theta}_i^a$ and $\vec{\theta}_i^b$ regardless of what (future) joint policies $\boldsymbol{\pi}_{\neq i}$ the other agents will use and hence we can cluster them without loss in heuristic value. Note that this does not depend on which heuristic is used and hence also holds for an optimal heuristic (i.e., when using an optimal Q-value function that gives the true value). This directly relates probabilistic equivalence with equivalence in optimal value.[12]

11. Although we focus on CBGs, these results generalize to BGs with individual payoff functions. Thus, they could potentially be exploited by algorithms for general-payoff BGs. Developing methods that do so is an interesting avenue for future work.
12. The proof originally provided by Oliehoek et al. (2009) is based on showing that histories that are PE will induce identical Q-values.

**Algorithm 6** `ClusterCBG`($B$)

**Input:** CBG $B$
**Output:** Losslessly clustered CBG $B$

```
for each agent $i$ do
    for each individual type $\theta_i \in B.\Theta_i$ do
        if $\Pr(\theta_i) = 0$ then
            $B.\Theta_i \leftarrow B.\Theta_i \backslash \theta_i$                              {Prune $\theta_i$ from $B$:}
            continue
        end if
        for each individual type $\theta'_i \in B.\Theta_i$ do
            isProbabilisticallyEquivalent ← true
            for all $\langle s, \boldsymbol{\theta}_{\neq i} \rangle$ do
                if $\Pr(s, \boldsymbol{\theta}_{\neq i} | \theta_i) \neq \Pr(s, \boldsymbol{\theta}_{\neq i} | \theta'_i)$ then
                    isProbabilisticallyEquivalent ← false
                    break
                end if
            end for
            if isProbabilisticallyEquivalent then
                $B.\Theta_i \leftarrow B.\Theta_i \backslash \theta'_i$                              {Prune $\theta'_i$ from $B$:}
                for each $\boldsymbol{a} \in \boldsymbol{\mathcal{A}}$ do
                    for all $\boldsymbol{\theta}_{\neq i}$ do
                        $u(\theta_i, \boldsymbol{\theta}_{\neq i}, \boldsymbol{a}) \leftarrow \min(u(\theta_i, \boldsymbol{\theta}_{\neq i}, \boldsymbol{a}), u(\theta'_i, \boldsymbol{\theta}_{\neq i}, \boldsymbol{a}))$     { take the lowest upper bound }
                        $\Pr(\theta_i, \boldsymbol{\theta}_{\neq i}) \leftarrow \Pr(\theta_i, \boldsymbol{\theta}_{\neq i}) + \Pr(\theta'_i, \boldsymbol{\theta}_{\neq i})$
                        $\Pr(\theta'_i, \boldsymbol{\theta}_{\neq i}) \leftarrow 0$
                    end for
                end for
            end if
        end for
    end for
end for
return $B$
```

Note that this result establishes a sufficient, but not necessary condition for lossless clustering. In particular, given policies for the other agents, many types are best-response equivalent and can be clustered. However, as far as we know, the criterion must hold in order to *guarantee* that two histories have the same best-response against *any* policy of the other agents.

### 3.2 GMAA* with Incremental Clustering

Knowing which individual histories can be clustered together without loss of value has the potential to speed up many Dec-POMDP methods. In this article, we focus on its application within the GMAA* framework.

Emery-Montemerlo et al. (2005) showed how clustering can be incorporated at every stage in their algorithm: when the CBG for a stage $t$ is constructed, a clustering of the individual histories (types) is performed first and only afterwards is the (reduced) CBG solved. The same approach can be employed within GMAA* by modifying the `Expand` procedure (Algorithm 5) to cluster the CBG before calling `GenerateAllChildrenForCBG`.

Algorithm 6 shows the clustering algorithm. It takes as input a CBG and returns the clustered CBG. It performs clustering by performing pairwise comparison of all types of each

**Algorithm 7** `ConstructExtendedBG`$(B,\boldsymbol{\beta}^{t-1},\widehat{Q})$

**Input:** A CBG $B$ for stage $t-1$, and the joint BG policy followed $\boldsymbol{\beta}^{t-1}$.
**Input:** An admissible heuristic of the form $\widehat{Q}(\vec{\boldsymbol{\theta}},\boldsymbol{a})$.
**Output:** CBG $B'$ for stage $t$.

$B' \leftarrow B$ {make a copy of $B$ that we subsequently alter}
**for** each agent $i$ **do**
  $B'.\Theta_i = \text{ConstructExtendedTypeSet}(i)$ {overwrite the individual type sets}
**end for**
$B'.\boldsymbol{\Theta} \leftarrow \times_{i\in\mathcal{D}}\Theta_i$ {the new joint type set (does not have to be explicitly stored)}
**for** each joint type $\boldsymbol{\theta} = (\boldsymbol{\theta}^{t-1},\boldsymbol{a}^{t-1},\boldsymbol{o}^t) \in B'.\boldsymbol{\Theta}$ **do**
  **for** each state $s^t \in \mathcal{S}$ **do**
    Compute $\Pr(s^t|\boldsymbol{\theta})$ {from $\Pr(s^{t-1}|\boldsymbol{\theta}^{t-1})$ via Bayes' rule }
  **end for**
  $\Pr(\boldsymbol{\theta}) \leftarrow \Pr(\boldsymbol{o}^t|\boldsymbol{\theta}^{t-1},\boldsymbol{a}^{t-1})\Pr(\boldsymbol{\theta}^{t-1})$
  **for** each $\boldsymbol{a} \in \boldsymbol{\mathcal{A}}$ **do**
    $q \leftarrow \infty$
    **for** each history $\vec{\boldsymbol{\theta}}^t$ represented by $\boldsymbol{\theta}$ **do**
      $q \leftarrow \min(q,\widehat{Q}(\vec{\boldsymbol{\theta}}^t,\boldsymbol{a}))$ { if $Q^* \leq \widehat{Q}$ we can take the lowest upper bound }
    **end for**
    $B'.u(\boldsymbol{\theta},\boldsymbol{a}) \leftarrow q$
  **end for**
**end for**
**return** $B'$

agent to see if they satisfy the criterion, yielding $O(|\Theta_i|^2)$ comparisons for each agent $i$. Each comparison involves looping over all $\langle s,\boldsymbol{\theta}_{\neq i}\rangle$ (line 9). If there are many states, some efficiency could be gained by first checking (3.2) and then checking (3.3). Rather than taking the average as in (3.13), on line 19 we take the lowest payoff, which can be done if we are using upper bound heuristic values.

The following theorem demonstrates that, when incorporating clustering into GMAA*, the resulting algorithm is still guaranteed to find an optimal solution.

**Theorem 4.** *When using a heuristic of the form* (2.9) *and clustering the CBGs in GMAA* using the PE criterion, the resulting search method is complete.*

*Proof.* Applying clustering does not alter the computation of lower bound values. Also, heuristic values computed for the expanded nodes are admissible and in fact unaltered as guaranteed by Corollary 2. Therefore, the only difference with regular GMAA* is that the class of considered joint policies is restricted to $\boldsymbol{\Pi}_C$, the class of clustered joint policies: not all possible child nodes are expanded, because clustering effectively prunes away policies that would specify different actions for AOHs that are PE and thus clustered. However, Corollary 1 guarantees that there exists an optimal joint policy in this restricted class. □

The modification of the `Expand` proposed above is rather naive. To construct $B(\boldsymbol{b}^0,\boldsymbol{\varphi}^t)$ it must first construct all $|\mathcal{O}_i|^t$ possible AOHs for agent $i$ (given the past policy $\varphi_i^t$). The subsequent clustering involves pairwise comparison of all these exponentially many types. Clearly, this is not tractable for later stages.

However, because PE of AOHs propagates forwards (i.e., identical extensions of PE histories are also PE), a more efficient approach is possible. Instead of clustering this exponentially

**Algorithm 8** $\texttt{Expand-IC}(q,H)$. The expand operator for GMAA*-IC.

**Input:** $q = \langle \boldsymbol{\varphi}^t, \hat{v} \rangle$ the search node to expand.
**Input:** $H$ the admissible heuristic that is of the form $\widehat{Q}(\vec{\boldsymbol{\theta}},\boldsymbol{a})$.
**Output:** $\mathcal{Q}_{\texttt{Expand}}$ the set containing expanded child nodes.

```
B(φ^{t-1}) ← φ^{t-1}.CBG        {retrieve previous CBG, note φ^t = (φ^{t-1}, β^{t-1})}
B(φ^t) ← ConstructExtendedBG(B(φ^{t-1}), β^{t-1}, Q̂)
B(φ^t) ← ClusterBG(B(φ^t))
φ^t.CBG ← B(φ^t)        {store pointer to this CBG}
Q_Expand ← GenerateAllChildrenForCBG(B(φ^t))
return Q_Expand
```

growing set of types, we can simply extend the already clustered types of the previous stage's CBG, as shown in Algorithm 7. That is, given $\Theta_i$, the set of types of agent $i$ at the previous stage $t-1$, and $\beta_i^{t-1}$ the policy agent $i$ took at that stage, the set of types at stage $t$, $\Theta_i'$, can be constructed as

$$\Theta_i' = \left\{ \theta_i' = (\theta_i, \beta_i^{t-1}(\theta_i), o_i^t) \mid \theta_i \in \Theta_i, o_i^t \in \mathcal{O}_i \right\}. \tag{3.14}$$

This means that the size of this newly constructed set is $|\Theta_i'| = |\Theta_i| \cdot |\mathcal{O}_i|$. If the type set $\Theta_i$ at the previous stage $t-1$ was much smaller than the set of all histories $|\Theta_i| \ll |\mathcal{O}_i|^{t-1}$, then the new type set $\Theta_i'$ is also much smaller: $|\Theta_i'| \ll |\mathcal{O}_i|^t$. In this way, we bootstrap the clustering at each stage and spend significantly less time clustering. We refer to the algorithm that implements this type of clustering as GMAA* *with Incremental Clustering* (GMAA*-IC). This approach is possible only because we perform an exact, value-preserving clustering for which Lemma 1 guarantees that identical extensions will also be clustered without loss in value. When performing the same procedure in a lossy clustering scheme (e.g., as in Emery-Montemerlo et al., 2005), errors might accumulate, and a better option might be to re-cluster from scratch at every stage.

Expansion of a GMAA*-IC node takes exponential time with respect to both the number of agents and types, as there are $O(|\mathcal{A}_*|^{n|\Theta_*|})$ joint CBG-policies and thus child nodes in the GMAA*-IC search tree ($\mathcal{A}_*$ is the largest action set and $\Theta_*$ is the largest type set). Clustering involves a pairwise comparison of all types of each agent and each of these comparisons needs to check $O(|\Theta_*|^{n-1}|\mathcal{S}|)$ numbers for equality to verify (3.1). The total cost of clustering can therefore be written as

$$O(n\,|\Theta_*|^2\,|\Theta_*|^{n-1}\,|\mathcal{S}|),$$

which is only polynomial in the number of types. When clustering decreases the number of types $|\Theta_*|$, it can therefore significantly reduce the number of child nodes and thereby the overall time needed. However, when no clustering is possible, some overhead will be incurred.

### 3.3 Improved Heuristic Representation

Since clustering can reduce the number of types, GMAA*-IC has the potential to scale to larger horizons. However, doing so has important consequences for the computation of the heuristics. Previous research has shown that the upper bound provided by $\mathrm{Q_{MDP}}$ is often too loose for effective heuristic search (Oliehoek, Spaan, & Vlassis, 2008). However, the space needed to store tighter heuristics such as $\mathrm{Q_{POMDP}}$ or $\mathrm{Q_{BG}}$ grows exponentially with the horizon. Recall from Section 2.2.2 (see Fig. 5) that there are two approaches to computing

**Algorithm 9** Compute Hybrid $\widehat{Q}$ with minimum size.

$Q^{h-1} \leftarrow \{R^1, \ldots, R^{|\mathcal{A}|}\}$ {vector representation of last stage}
$z \leftarrow |\mathcal{A}| \times |\mathcal{S}|$ {the size of the $|\mathcal{A}|$ vectors}
**for** $t = h-2$ to $0$ **do**
  $y \leftarrow |\vec{\Theta}^t| \times |\mathcal{A}|$ {size of AOH representation}
  **if** $z < y$ **then**
    $\mathcal{V} \leftarrow \text{VectorBackup}(Q^{t+1})$
    $\mathcal{V}' \leftarrow \text{Prune}(\mathcal{V})$
    $Q^t \leftarrow \mathcal{V}'$
    $z \leftarrow |\mathcal{V}'| \times |\mathcal{S}|$
  **end if**
  **if** $z \geq y$ **then**
    $Q^t \leftarrow \text{TreeBackup}(Q^{t+1})$ {From now on $z \geq y$}
  **end if**
**end for**

$Q_{\text{POMDP}}$ or $Q_{\text{BG}}$. The first constructs a tree of all joint AOHs and their heuristic values, which is simple to implement but requires storing a value for each $(\vec{\boldsymbol{\theta}}^t, \boldsymbol{a})$-pair, the number of which grows exponentially with $t$. The second approach maintains a vector-based representation, as is common for POMDPs. Though pruning can provide leverage, in the worst case, no pruning is possible and the number of maintained vectors grows doubly exponentially with $h - t$, the number of stages-to-go. Similarly, the initial belief and subsequently reachable beliefs can be used to reduce the number of vectors retained at each stage, but as the number of reachable beliefs is exponential in the horizon the exponential complexity remains.

Oliehoek, Spaan, and Vlassis (2008) used a tree-based representation for the $Q_{\text{POMDP}}$ and $Q_{\text{BG}}$ heuristics. Since the computational cost of solving the Dec-POMDP was the bottleneck, the inefficiencies in the representation could be overlooked. However, this approach is no longer feasible for the longer horizons made possible by GMAA*-IC.

To mitigate this problem, we propose a hybrid representation for the heuristics, as illustrated in Fig. 6. The main insight is that the exponential growth of the two existing representations occurs in opposite directions. Therefore, we can use the low space-complexity side of both representations: the later stages, which have fewer vectors, use a vector-based representation, while the earlier stages, which have fewer histories, use a history-based representation. This is similar to the idea of utilizing reachable beliefs to reduce the size of the vector representation described above but, rather than storing vectors for the appropriate AOHs at each step, only the values are needed when using the tree-based representation.

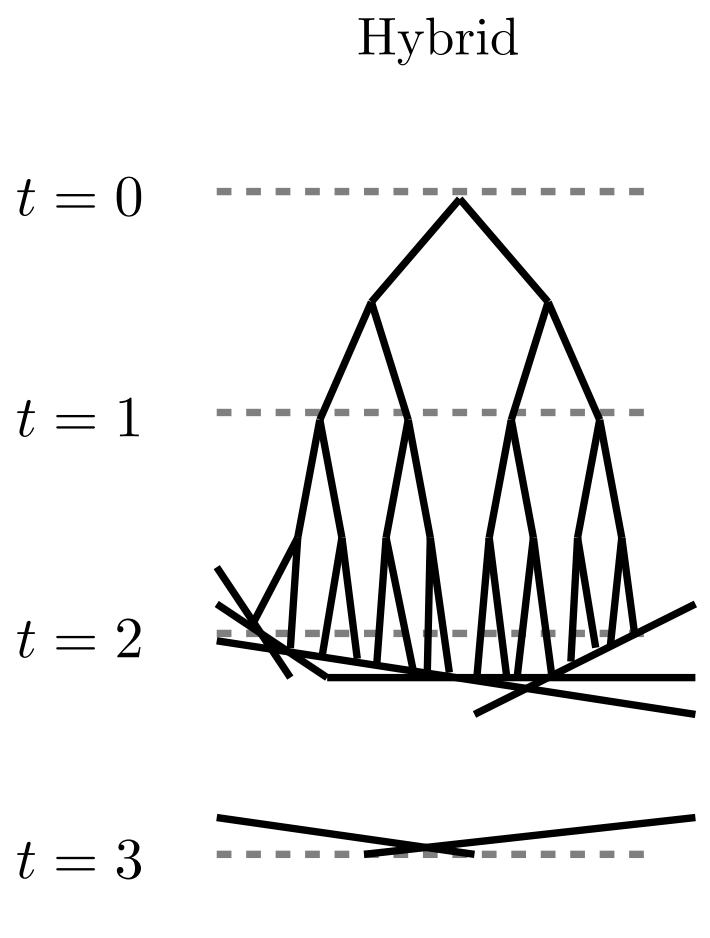

Figure 6: An illustration of the hybrid representation.

Algorithm 9 shows how, under mild assumptions, a minimally-sized representation can be computed. Starting from the last stage, the algorithm performs vector backups, switching to tree backups when they become the smaller option. For the last time step $h-1$, we represent $Q^t$ by the set of immediate reward vectors[13], and variable $z$ (initialized on line 2) keeps track of the number of parameters needed to represent $Q^t$ as vectors for the time step at hand. Note that $z$ depends on how effective the vector pruning is, i.e., how large the parsimonious representation of the piecewise linear and convex value function is. Since this is problem dependent, $z$ can be updated only after pruning has actually been performed (line 9). By contrast $y$, the number of parameters in a tree representation, can be computed directly from the Dec-POMDP (line 4). When $z > y$, the algorithm switches to tree backups.[14]

## 4. Incremental Expansion

The clustering technique presented in the previous section has the potential to significantly speed up planning if much clustering is possible. However, if little clustering is possible, the number of children in the GMAA* search tree will still grow super-exponentially. This section presents *incremental expansion*, a complementary technique to deal with this problem.

Incremental expansion exploits recent improvements in effectively solving CBGs. First note that during the expansion of the last stage $t = h - 1$ for a particular $\boldsymbol{\varphi}^{h-1}$, we are only interested in the best child $(\boldsymbol{\varphi}^{h-1}, \boldsymbol{\delta}^{h-1,*})$, which corresponds to the optimal solution of the Bayesian game $\boldsymbol{\delta}^{h-1,*} \leftrightarrow \boldsymbol{\beta}^*$. As such, for this last stage, we can use new methods for solving CBGs (Kumar & Zilberstein, 2010b; Oliehoek, Spaan, Dibangoye, & Amato, 2010) that can provide speedups of multiple orders of magnitude over brute force search (enumeration).[15] Unfortunately, the improvements to GMAA* afforded by this approach are limited: in order to guarantee optimality, it still relies on expansion of *all* (child nodes corresponding to all) joint CBG-policies $\boldsymbol{\beta}$ for the intermediate stages, thus necessitating a brute-force approach. However, many of the expanded child nodes may have low heuristic values $\widehat{V}$ and may therefore never be selected for further expansion.

Incremental expansion overcomes this problem because it exploits the following key observation: if we can generate the children in decreasing heuristic order using an admissible heuristic, we do not have to expand all the children. As before, an A* search is performed over partially specified policies and each new CBG is constructed by extending the CBG for the parent node. However, rather than fully expanding (i.e., enumerating all the CBG policies of and thereby constructing all children for) each search node, we instantiate an *incremental CBG solver* for the corresponding CBG. This incremental solver returns only one joint CBG policy at a time, which is then used to construct a single child $\boldsymbol{\varphi}^{t+1} = (\boldsymbol{\varphi}^t, \boldsymbol{\beta})$. By revisiting the nodes, only the promising child nodes are expanded incrementally.

Below, we describe GMAA*-ICE, an algorithm that combines GMAA*-IC with incremental expansion. We establish theoretical guarantees and describe the modifications to BaGaBaB, the CBG solver that GMAA*-ICE employs, that are necessary to deliver the child nodes in decreasing order.

13. Only in exceptional cases where a short horizon is combined with large state and action spaces will representing the last time step as vectors not be minimal. In such cases, the algorithm can be trivially adapted.
14. This assumes that the vector representation will not shrink again for earlier stages. Although unlikely in practice, such cases would prevent the algorithm from computing a minimal representation.
15. Kumar and Zilberstein (2010b) tackle a slightly different problem; they introduce a weighted constraint satisfaction approach to solving the point-based backup in dynamic programming for Dec-POMDPs. However, this point-based backup can be interpreted as a collection of CBGs (Oliehoek et al., 2010).

### 4.1 GMAA* with Incremental Clustering and Expansion

We begin by formalizing incremental expansion and incorporating it into GMAA*-IC, yielding GMAA* *with incremental clustering and expansion* (GMAA*-ICE). At the core of incremental expansion lies the following lemma:

**Lemma 3.** *Given two joint CBG policies $\boldsymbol{\beta}, \boldsymbol{\beta}'$ for a CBG $B(\boldsymbol{b}^0,\boldsymbol{\varphi}^t)$, if $\widehat{V}(\boldsymbol{\beta}) \geq \widehat{V}(\boldsymbol{\beta}')$, then for the corresponding child nodes $\widehat{V}(\boldsymbol{\varphi}^{t+1}) \geq \widehat{V}(\boldsymbol{\varphi}^{t+1\prime})$.*

*Proof.* This holds directly by the definition of $\widehat{V}(\boldsymbol{\varphi}^t)$ as given by (2.8):

$$\begin{aligned}\widehat{V}(\boldsymbol{\varphi}^{t+1}) &= V^{0\ldots(t-1)}(\boldsymbol{\varphi}^t) + \widehat{V}(\boldsymbol{\beta}) \\ &\geq V^{0\ldots(t-1)}(\boldsymbol{\varphi}^t) + \widehat{V}(\boldsymbol{\beta}') = \widehat{V}(\boldsymbol{\varphi}^{t+1\prime}).\end{aligned}$$ □

It follows directly that, if for $B(\boldsymbol{b}^0,\boldsymbol{\varphi}^t)$ we use a CBG solver that can generate a sequence of policies $\boldsymbol{\beta}, \boldsymbol{\beta}', \ldots$ such that

$$\widehat{V}(\boldsymbol{\beta}) \geq \widehat{V}(\boldsymbol{\beta}') \geq \ldots$$

then, for the sequence of corresponding children

$$\widehat{V}(\boldsymbol{\varphi}^{t+1}) \geq \widehat{V}(\boldsymbol{\varphi}^{t+1\prime}) \geq \ldots.$$

Exploiting this knowledge, we can expand only the first child $\boldsymbol{\varphi}^{t+1}$ and compute its heuristic value $\widehat{V}(\boldsymbol{\varphi}^{t+1})$ using (2.8). Since all the unexpanded siblings will have heuristic values less than or equal to that, we can modify GMAA*-IC to reinsert the node $q$ into the open list L to act as a *placeholder* for all its non-expanded children.

**Definition 9.** A *placeholder* is a node for which at least one child has been expanded. A placeholder has a heuristic value equal to its last expanded child.

Thus, after expansion of a search node $q$'s child, we update $q.\hat{v}$, the heuristic value of the node, to $\widehat{V}(\boldsymbol{\varphi}^{t+1})$, the value of the expanded child, i.e., we set $q.\hat{v} \leftarrow \widehat{V}(\boldsymbol{\varphi}^{t+1})$. As such, we can reinsert $q$ into L as a placeholder. As mentioned above, this is correct because all the unexpanded siblings (for which the parent node $q$ now is a placeholder) have heuristic values lower than or equal to $\widehat{V}(\boldsymbol{\varphi}^{t+1})$. Therefore the next sibling $q'$ represented by the placeholder is always expanded in time: $q'$ is always created before nodes with lower heuristic value are selected for further expansion. We keep track of whether a node is a previously expanded placeholder or not.

As before, GMAA*-ICE performs an A* search over partially specified policies. As in GMAA*-IC, each new CBG is constructed by extending the CBG for the parent node and then applying lossless clustering. However, rather than expanding all children, GMAA*-ICE requests only the next solution $\boldsymbol{\beta}$ of an *incremental CBG solver*, from which a single child $\boldsymbol{\varphi}^{t+1} = (\boldsymbol{\varphi}^t, \boldsymbol{\beta})$ is constructed. In principle GMAA*-ICE can use any CBG solver that is able to incrementally deliver all $\boldsymbol{\beta}$ in descending order of $\widehat{V}(\boldsymbol{\beta})$. We propose a modification of the BaGaBaB algorithm (Oliehoek et al., 2010), briefly discussed in Section 4.3.

Fig. 7 illustrates the process of incremental expansion in GMAA*-ICE, with $\boldsymbol{\varphi}^t$ indexed by letters. First, a CBG solver for the root node $\langle a, 7\rangle$ is created, and the optimal solution $\boldsymbol{\beta}^*$ is computed, with value 6. This results in a child $\langle b, 6\rangle$, and the root is replaced by a placeholder node $\langle a, 6\rangle$. As per Definition 5 (the node comparison operator), $b$ appears before $a$ in the open list and hence is selected for expansion. Its best child $\langle c, 4 \rangle$ is added and $\langle b, 6 \rangle$ is replaced by placeholder $\langle b, 4 \rangle$. Now the search returns to the root node, and the second best solution $\boldsymbol{\beta}'$ is obtained from the CBG solver, leading to child $\langle d, 5.5 \rangle$. Placeholder nodes are retained as long as they have unexpanded children; only their values are updated.

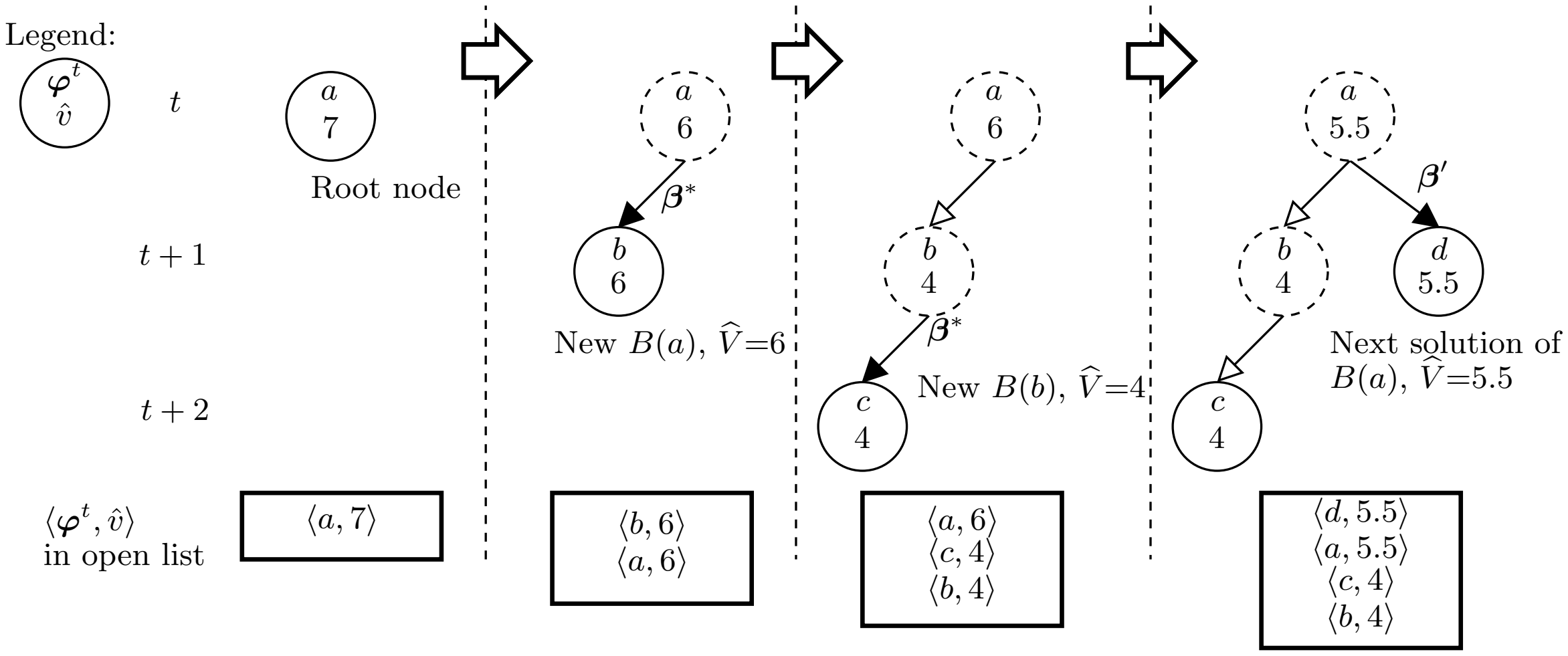

Figure 7: Illustration of incremental expansion, with the nodes in the open list at the bottom. Past joint policies $\boldsymbol{\varphi}^t$ are indexed by letters. Placeholder nodes are indicated by dashes.

When using GMAA*-ICE, we can derive lower and upper bounds for the CBG solution, which can be exploited by the incremental CBG solver. The incremental CBG solver for $B(\boldsymbol{\varphi}^t)$ can be initialized with lower bound

$$\underline{\mathrm{v}}^{CBG} = \underline{\mathrm{v}}^{GMAA} - V^{0\dots(t-1)}(\boldsymbol{\varphi}^t), \tag{4.1}$$

where $\underline{\mathrm{v}}^{GMAA}$ is the value of the current best solution, and $V^{0\dots(t-1)}(\boldsymbol{\varphi}^t)$ is the true expected value of $\boldsymbol{\varphi}^t$ over the first $t$ stages. Therefore, $\underline{\mathrm{v}}^{CBG}$ is the minimum value that a candidate must generate over the remaining $h-t$ stages in order to beat the current best solution. Note that each time the incremental CBG solver is queried for a solution, $\underline{\mathrm{v}}^{CBG}$ is re-evaluated (using (4.1)), because $\underline{\mathrm{v}}^{GMAA}$ may have changed.

When the used heuristic faitfully represents the immediate reward (i.e., is of the form (2.9)), then, for the last stage $t = h-1$, we can also specify an upper bound for the solution of the CBG

$$\bar{\mathrm{v}}^{CBG} = \widehat{V}(\boldsymbol{\varphi}^{h-1}) - V^{0\dots(h-2)}(\boldsymbol{\varphi}^{h-1}). \tag{4.2}$$

If this upper bound is attained, no further solutions will be required from the CBG solver. The upper bound holds since by (2.8)

$$\begin{aligned} \widehat{V}(\boldsymbol{\beta}) &\triangleq \widehat{V}(\boldsymbol{\varphi}^h) - V^{0\dots(h-2)}(\boldsymbol{\varphi}^{h-1}) \\ &= V(\boldsymbol{\varphi}^h) - V^{0\dots(h-2)}(\boldsymbol{\varphi}^{h-1}) \\ &\leq \widehat{V}(\boldsymbol{\varphi}^{h-1}) - V^{0\dots(h-2)}(\boldsymbol{\varphi}^{h-1}). \end{aligned}$$

In the first step, $\widehat{V}(\boldsymbol{\varphi}^h) = V(\boldsymbol{\varphi}^h)$, because $\boldsymbol{\varphi}^h$ is a fully specified policy and the heuristic value given by (2.8) equals the actual value when a heuristic that faithfully represents the expected

**Algorithm 10** Expand-ICE$(q, H)$. The expand operator for GMAA*-ICE.

**Input:** $q = \langle \boldsymbol{\varphi}^t, \hat{v} \rangle$ the search node to expand.
**Input:** $H$ the admissible heuristic that is of the form $\widehat{Q}(\vec{\boldsymbol{\theta}}, \boldsymbol{a})$.
**Output:** $\mathcal{Q}_{\texttt{Expand}}$ the set containing 0 or 1 expanded child nodes.

**if** IsPlaceholder$(q)$ **then**
  $B(\boldsymbol{\varphi}^t) \leftarrow \boldsymbol{\varphi}^t$.CBG {reuse stored CBG}
**else**
  $B(\boldsymbol{\varphi}^{t-1}) \leftarrow \boldsymbol{\varphi}^{t-1}$.CBG {retrieve previous CBG, note $\boldsymbol{\varphi}^t = (\boldsymbol{\varphi}^{t-1}, \boldsymbol{\beta}^{t-1})$}
  $B(\boldsymbol{\varphi}^t) \leftarrow$ ConstructExtendedBG$(B(\boldsymbol{\varphi}^{t-1}), \boldsymbol{\beta}^{t-1}, \widehat{Q})$
  $B(\boldsymbol{\varphi}^t) \leftarrow$ ClusterBG$(B(\boldsymbol{\varphi}^t))$
  $B(\boldsymbol{\varphi}^t)$.Solver $\leftarrow$ CreateSolver$(B(\boldsymbol{\varphi}^t))$
  $\boldsymbol{\varphi}^t$.CBG $\leftarrow B(\boldsymbol{\varphi}^t)$ {store pointer to this CBG}
**end if**
$\underline{\mathrm{v}}^{CBG} = \underline{\mathrm{v}}^{GMAA} - V^{0\ldots(t-1)}(\boldsymbol{\varphi}^t)$ {set lower bound for CBG solution}
**if** $t = h-1$ **then**
  $\bar{\mathrm{v}}^{CBG} = \widehat{V}(\boldsymbol{\varphi}^{h-1}) - V^{0\ldots(h-2)}(\boldsymbol{\varphi}^{h-1})$ {upper bound only used for last stage CBG}
**else**
  $\bar{\mathrm{v}}^{CBG} = +\infty$
**end if**
$\langle \boldsymbol{\beta}^t, \widehat{V}(\boldsymbol{\beta}^t) \rangle \leftarrow B(\boldsymbol{\varphi}^t)$.Solver.NextSolution$(\underline{\mathrm{v}}^{CBG}, \bar{\mathrm{v}}^{CBG})$ {compute next CBG solution}
**if** $\boldsymbol{\beta}^t$ **then**
  $\boldsymbol{\varphi}^{t+1} \leftarrow (\boldsymbol{\varphi}^t, \boldsymbol{\beta}^t)$ {create partial joint policy}
  $\widehat{V}(\boldsymbol{\varphi}^{t+1}) \leftarrow V^{0\ldots t-1}(\boldsymbol{\varphi}^t) + \widehat{V}(\boldsymbol{\beta}^t)$ {compute heuristic value}
  $q' \leftarrow \langle \boldsymbol{\varphi}^{t+1}, \widehat{V}(\boldsymbol{\varphi}^{t+1}) \rangle$ {create child node}
  $\mathcal{Q}_{\texttt{Expand}} \leftarrow \{q'\}$
**else**
  $\mathcal{Q}_{\texttt{Expand}} \leftarrow \emptyset$ {fully expanded: exists no solution s.t. $V(\boldsymbol{\beta}^{h-1}) \geq \underline{\mathrm{v}}^{CBG}$}
**end if**
**return** $\mathcal{Q}_{\texttt{Expand}}$

**Algorithm 11** PostProcessNode-ICE$(q, \mathrm{L})$: Post processing of a node in GMAA*-ICE.

**Input:** $q$ the last expanded node, L the open list.
**Output:** $q$ is either removed or updated.

L.Pop$(q)$
**if** $q$ is fully expanded **or** depth$(q) = h-1$ **then**
  Cleanup $q$ {delete the node and the associated CBG and Solver}
  **return**
**else**
  $c \leftarrow$ last expanded child of $q$
  $q.\hat{v} \leftarrow c.\hat{v}$ {update heuristic value of parent node}
  IsPlaceholder$(q) \leftarrow$ true {remember that $q$ is a placeholder}
  L.Insert$(q)$ {reinsert at appropriate position}
**end if**

immediate reward is used. This implies that $\widehat{V}(\boldsymbol{\beta})$ itself is a lower bound. In the second step $V(\boldsymbol{\varphi}^h) \leq \widehat{V}(\boldsymbol{\varphi}^{h-1})$, because $\widehat{V}(\boldsymbol{\varphi}^{h-1})$ is admissible. Therefore, we can stop expanding when we find a $\boldsymbol{\beta}$ with (lower bound) heuristic value equal to the upper bound $\bar{\mathrm{v}}^{CBG}$. This applies only to the last stage because only then the first step is valid.

GMAA*-ICE can be implemented by replacing the `Expand` and the `PostProcessNode` procedures of Algorithms 8 and 4 by Algorithms 10 and 11, respectively. `Expand`-ICE first determines if a placeholder is being used and either reuses the previously constructed incremental CBG solver or constructs a new one. Then, new bounds are calculated and the next CBG solution is obtained. Subsequently, only a single child node is generated (rather than expanding all children as in Algorithm 13). `PostProcessNode`-ICE removes the last node that was returned by `Select` only when all its children have been expanded. Otherwise, it updates that node's heuristic value and reinserts it in the open list. See Appendix A.2 for GMAA*-ICE shown as a single algorithm.

### 4.2 Theoretical Guarantees

In this section, we prove that GMAA*-IC and GMAA*-ICE are *search-equivalent*. As a direct result we establish that GMAA*-ICE is complete, which means that integrating incremental expansion preserves the optimality guarantees of GMAA*-IC.

**Definition 10.** We call two GMAA* variants *search-equivalent* if they select exactly the same sequence of non-placeholder nodes corresponding to past joint policies to expand in the search tree using the `Select` operator.

For GMAA*-IC and GMAA*-ICE we show that the set of *selected* nodes are the same. However, the set of *expanded* nodes can be different; in fact, it is precisely these differences that incremental expansion exploits.

**Theorem 5.** GMAA*-ICE *and* GMAA*-IC *are search-equivalent.*

*Proof.* Proof is listed in Section A.4 of the appendix. □

Note that Theorem 5 does not imply that the computational and space requirements of GMAA*-ICE and GMAA*-IC are identical. On the contrary, for each expansion, GMAA*-ICE generates only one child node to be stored on the open list. In contrast, GMAA*-IC generates a number of child nodes that is, in the worst case, doubly exponential in the depth of the selected node.[16] However, GMAA*-ICE is not guaranteed to be more efficient than GMAA*-IC. For example, in the case where all child nodes still have to be generated, GMAA*-ICE will be slower due to the overhead it incurs.

**Corollary 3.** *When using a heuristic of the form* (2.9) GMAA*-ICE *is complete.*

*Proof.* Under the stated conditions, GMAA*-IC is complete (see Theorem 4). Since GMAA*-ICE is search equivalent to GMAA*-IC, it is also complete. □

16. When a problem allows clustering, the number of child nodes grows less dramatically (see Section 3).

### 4.3 Incremental CBG Solvers

Implementing GMAA*-ICE requires a CBG solver that can incrementally deliver all $\boldsymbol{\beta}$ in descending order of $\widehat{V}(\boldsymbol{\beta})$. To this end, we propose to modify the Bayesian game Branch and Bound (BaGaBaB) algorithm (Oliehoek et al., 2010). BaGaBaB performs an A*-search over (partially specified) CBG policies. Thus, when applied within GMAA*-ICE, it performs a *second, nested* A* search. To expand each node in the GMAA* search tree, a nested A* search computes the next CBG solution.[17] This section briefly summarizes the main ideas behind BaGaBaB (for more information, see Oliehoek et al., 2010) and our modifications.

BaGaBaB works by creating a search tree in which the nodes correspond to partially specified joint CBG policies. In particular, it represents a $\boldsymbol{\beta}$ as a *joint action vector*, a vector $\langle \boldsymbol{\beta}(\boldsymbol{\theta}^1), \ldots, \boldsymbol{\beta}(\boldsymbol{\theta}^{|\boldsymbol{\Theta}|}) \rangle$ of the joint actions that $\boldsymbol{\beta}$ specifies for each joint type. Each node $g$ in the BaGaBaB search tree represents a partially specified vector and thus a partially specified joint CBG policy. For example, a completely unspecified vector $\langle \cdot, \cdot, \ldots, \cdot \rangle$ corresponds to the root node, while an internal node $g$ at depth $d$ (root being at depth 0) specifies joint actions for the first $d$ joint types $g = \langle \boldsymbol{\beta}(\boldsymbol{\theta}^1), \ldots, \boldsymbol{\beta}(\boldsymbol{\theta}^d), \cdot, \cdot, \ldots, \cdot \rangle$. The value of a node $V(g)$ is the value of the best joint CBG-policy consistent with it. Since this value is not known in advance, BaGaBaB performs an A* search guided by an optimistic heuristic.

In particular, we can compute an upper bound on the value achievable for any such partially specified vector by computing the maximum value of the *complete information joint policy* that is consistent with it (i.e., a non-admissible joint policy that selects the maximizing joint actions for the remaining joint types). Since this value is a guaranteed upper bound on the maximum value achievable by a consistent joint CBG policy, it is an admissible heuristic.

We propose a modification to BaGaBaB to allow solutions to be incrementally delivered. The main idea is to retain the search tree after a first call of BaGaBaB on a particular CBG $B(\boldsymbol{\varphi}^t)$ and update it during subsequent calls, thereby saving computational effort.

Standard A* search terminates when a single optimal solution has been found. This behavior is the same when incremental BaGaBaB is called for the first time on a $B(\boldsymbol{\varphi}^t)$. However, during standard A*, nodes whose upper bound is lower than the best known lower bound can be safely deleted, as they will never lead to an optimal solution. In contrast, in an incremental setting such nodes cannot be pruned, as they could possibly result in the $k$-th best solution and therefore might need to be expanded during subsequent calls to BaGaBaB. Only nodes returned as solutions are pruned in order to avoid returning the same solution twice. This modification requires more memory but does not affect the A* search process otherwise.

When asked it for the $k$-th solution, BaGaBaB resets its internal lower bound to the value of the next-best solution that was previously found but not returned (or to $\underline{v}^{CBG}$ as defined in (4.1) if no such solution was found). Then it starts an A* search initialized using the search tree resulting from the $(k-1)$-th solution. In essence, this method is similar to searching for the best $k$ solutions, where $k$ can be incremented on demand. Recently it was shown that, for fixed $k$, such a modification preserves all the theoretical guarantees (soundness, completeness,

17. While GMAA*-ICE could also use any other incremental CGB solver, there are few that avoid enumerating all $\boldsymbol{\beta}$ before providing the first result and thus have the potential to work incrementally. An exception may be the method of Kumar and Zilberstein (2010b), which employs AND/OR branch and bound search with the EDAC heuristic (and is thus limited to the two-agent case). As a heuristic search method, it may be amenable to an incremental implementation though to our knowledge this has not been attempted.

optimal efficiency) of the A* algorithm (Dechter, Flerova, & Marinescu, 2012), but the results trivially transfer to the setting where $k$ is allowed to increase.

## 5. Experiments

In this section, we empirically test and validate all the proposed techniques: lossless clustering of joint histories, incremental expansion of search nodes, and hybrid heuristic representations. After introducing the experimental setup, we compare the performance of GMAA*-IC and GMAA*-ICE to that of GMAA* on a suite of benchmark problems from the literature. Next, we compare the performance of the proposed methods with state-of-the-art optimal and approximate Dec-POMDP methods, followed by a case study of the scaling behavior with respect to the number of agents. Finally, we compare memory requirements of the hybrid heuristic representation to those of the tree and vector representations.

### 5.1 Experimental Setup

The most well-known Dec-POMDP benchmarks are the Dec-Tiger (Nair et al., 2003) and BroadcastChannel (Hansen et al., 2004) problems. Dec-Tiger was discussed extensively in Section 2. In BroadcastChannel, two agents have to transmit messages over a communication channel, but when both agents transmit at the same time a collision occurs that is noisily observed by the agents. The FireFighting problem models a team of $n$ firefighters that have to extinguish fires in a row of $n_h$ houses (Oliehoek, Spaan, & Vlassis, 2008). Each agent can choose to move to any of the houses to fight fires at that location; if two agents are in the same house, they will completely extinguish any fire there. The (negative) reward of the team of firefighters depends on the intensity of the fire at each house; when all fires have been extinguished, reward of zero is received. In the Hotel 1 problem (Spaan & Melo, 2008), travel agents need to assign customers to hotels with limited capacity. They can also send a customer to a resort but this yields lower reward. In addition, we also use the following problems: Recycling Robots (Amato, Bernstein, & Zilberstein, 2007), a scaled-down version of the problem described in Section 2; GridSmall with two observations (Amato, Bernstein, & Zilberstein, 2006) and Cooperative Box Pushing (Seuken & Zilberstein, 2007a), a larger two-robot benchmark. Table 1 summarizes these problems numerically, listing the number of joint policies for different planning horizons.

Experiments were run on an Intel Core i5 CPU running Linux, and GMAA*, GMAA*-IC, and GMAA*-ICE were implemented in the same code-base using the MADP Toolbox (C++) (Spaan & Oliehoek, 2008). The vector-based $Q_{\text{BG}}$ representation is computed using a variation of Incremental Pruning (adapted for computing $Q$-functions instead of regular value functions), corresponding to the NaiveIP method as described by Oliehoek and Spaan (2012). To implement the pruning, we employ Cassandra's POMDP-solve software (A. R. Cassandra, 1998).

For the results in Sections 5.2 and 5.3, we limited each process to 2Gb RAM and a maximum CPU time of 3,600s. Reported CPU times are averaged over 10 independent runs and have a resolution of 0.01s. Timings are given only for the MAA* search processes, since computation of the heuristic is the same for both methods and can be amortized over multiple runs.[18] All problem definitions are available via `http://masplan.org`.

| | problem primitives | | | | num. $\boldsymbol{\pi}$ for $h$ | | |
|---|---|---|---|---|---|---|---|
| | $n$ | $\lvert\mathcal{S}\rvert$ | $\lvert\mathcal{A}_i\rvert$ | $\lvert\mathcal{O}_i\rvert$ | 2 | 4 | 6 |
| Dec-Tiger | 2 | 2 | 3 | 2 | 7.29$e$2 | 2.06$e$14 | 1.31$e$60 |
| BroadcastChannel | 2 | 4 | 2 | 2 | 6.40$e$1 | 1.07$e$9 | 8.51$e$37 |
| GridSmall | 2 | 16 | 5 | 2 | 1.563$e$4 | 9.313$e$20 | 1.175$e$88 |
| Cooperative Box Pushing | 2 | 100 | 4 | 5 | 1.68$e$7 | 6.96$e$187 | 1.96$e$4703 |
| Recycling Robots | 2 | 4 | 3 | 2 | 7.29$e$2 | 2.06$e$14 | 1.31$e$60 |
| Hotel 1 | 2 | 16 | 3 | 4 | 5.90$e$4 | 1.29$e$81 | 3.48$e$1302 |
| FireFighting | 2 | 432 | 3 | 2 | 7.29$e$2 | 2.06$e$14 | 1.31$e$60 |

Table 1: Benchmark problem sizes and number of joint policies for different horizons.

### 5.2 Comparing GMAA*, GMAA*-IC, and GMAA*-ICE

We compared GMAA*, GMAA*-IC, and GMAA*-ICE using the hybrid $Q_{BG}$ representation. While all methods compute an optimal policy, we expect GMAA*-IC to be more efficient than GMAA* when lossless clustering is possible. Furthermore, we expect GMAA*-ICE to provide further improvements in terms of speedup and scaling to longer planning horizons.

The results are shown in Table 2. For all entries where we report results, the $Q_{BG}$ heuristics could be computed, thanks to the hybrid representation. Consequently, the performance of GMAA*-IC is much better than all previously reported results, including those of Oliehoek et al. (2009), who were often required to resort to $Q_{MDP}$ for larger problems and/or horizons. The entries marked by '§' show the limits when using $Q_{MDP}$ instead of $Q_{BG}$: in most of these problems we can reach longer horizons with $Q_{BG}$. Only for FireFighting can GMAA*-ICE with $Q_{MDP}$ compute solutions for higher $h$ than is possible with $Q_{BG}$ (hence the missing "§", and showing that GMAA*-ICE is more efficient using a loose heuristic than GMAA*-IC). Furthermore, the "†" entries indicate that the horizon to which we can solve a problem with a tree-based $Q_{BG}$ representation is often much shorter.

These results clearly illustrate that GMAA*-IC leads to a significant improvement in performance. In all problems, GMAA*-IC was able to produce a solution more quickly and to increase the largest solvable horizon over GMAA*. In some cases, GMAA*-IC is able to drastically increase the solvable horizon.

Furthermore, the results clearly demonstrate that incremental expansion allows for significant additional improvements. In fact, the table demonstrates that GMAA*-ICE significantly outperforms GMAA*-IC, especially in problems where little clustering is possible.

The results in Table 2 also illustrate the efficacy of a hybrid representation. For problems like GridSmall, Cooperative Box Pushing, FireFighting and Hotel 1 neither the tree nor vector representation is able to provide a compact $Q_{BG}$ heuristic for the longer hori-

18. The heuristics' computation time ranges from less than a second to hours (for high $h$ in some difficult problems). Table 4 presents some heuristic computation time results.

| $h$ | $V^*$ | $T_{\text{GMAA}^*}$(s) | $T_{IC}$(s) | $T_{ICE}$(s) |
|---|---|---|---|---|
| | | DEC-TIGER | | |
| 2 | −4.000000 | $\leq 0.01$ | $\leq 0.01$ | $\leq 0.01$ |
| 3 | 5.190812 | §$\leq 0.01$ | $\leq 0.01$ | $\leq 0.01$ |
| 4 | 4.802755 | 563.09 | §0.27 | $\leq 0.01$ |
| 5 | **7.026451** | – | †21.03 | §†0.02 |
| 6 | **10.381625** | | – | 46.43 |
| 7 | | | | $*$ |
| | | FIREFIGHTING $\langle n_h = 3, n_f = 3\rangle$ | | |
| 2 | −4.383496 | 0.09 | $\leq 0.01$ | $\leq 0.01$ |
| 3 | −5.736969 | §3.05 | §0.11 | 0.10 |
| 4 | **−6.578834** | 1001.53 | †950.51 | 1.00 |
| 5 | **−7.069874** | – | – | †4.40 |
| 6 | **−7.175591** | | 0.08 | 0.07 |
| 7 | | | # | # |
| | | GRIDSMALL | | |
| 2 | 0.910000 | $\leq 0.01$ | $\leq 0.01$ | $\leq 0.01$ |
| 3 | 1.550444 | §0.90 | §0.10 | $\leq 0.01$ |
| 4 | **2.241577** | $*$ | †1.77 | §†$\leq 0.01$ |
| 5 | **2.970496** | | – | 0.02 |
| 6 | **3.717168** | | – | 0.04 |
| 7 | | | # | # |
| | | HOTEL 1 | | |
| 2 | **10.000000** | §$\leq 0.0$ | $\leq 0.01$ | $\leq 0.01$ |
| 3 | **16.875000** | $*$ | $\leq 0.01$ | $\leq 0.01$ |
| 4 | **22.187500** | | §†$\leq 0.01$ | §†$\leq 0.01$ |
| 5 | **27.187500** | | $\leq 0.01$ | $\leq 0.01$ |
| 6 | **32.187500** | | $\leq 0.01$ | $\leq 0.01$ |
| 7 | **37.187500** | | $\leq 0.01$ | $\leq 0.01$ |
| 8 | **42.187500** | | $\leq 0.01$ | $\leq 0.01$ |
| 9 | **47.187500** | | 0.02 | $\leq 0.01$ |
| 10 | | | # | # |
| | | COOPERATIVE BOX PUSHING | | |
| 2 | 17.600000 | §0.02 | $\leq 0.01$ | $\leq 0.01$ |
| 3 | 66.081000 | $*$ | §†0.11 | †$\leq 0.01$ |
| 4 | 98.593613 | | $*$ | §313.07 |
| 5 | | | # | # |

| $h$ | $V^*$ | $T_{\text{GMAA}^*}$(s) | $T_{IC}$(s) | $T_{ICE}$(s) |
|---|---|---|---|---|
| | | RECYCLING ROBOTS | | |
| 2 | 7.000000 | $\leq 0.01$ | $\leq 0.01$ | $\leq 0.01$ |
| 3 | 10.660125 | §$\leq 0.01$ | $\leq 0.01$ | $\leq 0.01$ |
| 4 | 13.380000 | 713.41 | $\leq 0.01$ | $\leq 0.01$ |
| 5 | 16.486000 | – | †$\leq 0.01$ | †$\leq 0.01$ |
| 6 | **19.554200** | | $\leq 0.01$ | $\leq 0.01$ |
| 10 | **31.863889** | | $\leq 0.01$ | $\leq 0.01$ |
| 15 | **47.248521** | | §$\leq 0.01$ | $\leq 0.01$ |
| 18 | **56.479290** | | $\leq 0.01$ | §$\leq 0.01$ |
| 20 | **62.633136** | | $\leq 0.01$ | $\leq 0.01$ |
| 30 | **93.402367** | | 0.08 | 0.05 |
| 40 | **124.171598** | | 0.42 | 0.25 |
| 50 | **154.940828** | | 2.02 | 1.27 |
| 60 | **185.710059** | | 9.70 | 6.00 |
| 70 | **216.479290** | | – | 28.66 |
| 80 | | | – | – |
| | | BROADCASTCHANNEL | | |
| 2 | 2.000000 | $\leq 0.01$ | $\leq 0.01$ | $\leq 0.01$ |
| 3 | 2.990000 | $\leq 0.01$ | $\leq 0.01$ | $\leq 0.01$ |
| 4 | 3.890000 | §$\leq 0.01$ | $\leq 0.01$ | $\leq 0.01$ |
| 5 | 4.790000 | 1.27 | $\leq 0.01$ | $\leq 0.01$ |
| 6 | **5.690000** | – | $\leq 0.01$ | $\leq 0.01$ |
| 7 | **6.590000** | | †$\leq 0.01$ | †$\leq 0.01$ |
| 10 | **9.290000** | | $\leq 0.01$ | $\leq 0.01$ |
| 20 | **18.313228** | | $\leq 0.01$ | $\leq 0.01$ |
| 25 | **22.881523** | | $\leq 0.01$ | $\leq 0.01$ |
| 30 | **27.421850** | | $\leq 0.01$ | $\leq 0.01$ |
| 40 | **36.459724** | | $\leq 0.01$ | $\leq 0.01$ |
| 50 | **45.501604** | | $\leq 0.01$ | $\leq 0.01$ |
| 53 | **48.226420** | | §$\leq 0.01$ | §$\leq 0.01$ |
| 100 | **90.760423** | | $\leq 0.01$ | $\leq 0.01$ |
| 250 | **226.500545** | | 0.06 | 0.07 |
| 500 | **452.738119** | | 0.81 | 0.94 |
| 600 | **543.228071** | | 11.63 | 13.84 |
| 700 | **633.724279** | | 0.52 | 0.63 |
| 800 | | | – | – |
| 900 | **814.709393** | | 9.57 | 11.11 |
| 1000 | | | – | – |

Table 2: Experimental results comparing regular GMAA*, GMAA*-IC, and GMAA*-ICE. Listed are the computation times of GMAA* ($T_{\text{GMAA}^*}$), GMAA*-IC ($T_{IC}$), and GMAA*-ICE ($T_{ICE}$), using the hybrid $Q_{\text{BG}}$ representation. We use the following symbols: '–' memory limit violations, '$*$' time limit overruns, '#' heuristic computation exceeded memory or time limits, '§' maximum planning horizon using $Q_{\text{MDP}}$, '†' maximum planning horizon using tree-based $Q_{\text{BG}}$. Bold entries indicate that only the methods proposed in this article have computed these results.

zons. Apart from DEC-TIGER and FIREFIGHTING, computing and storing $Q_{\text{BG}}$ (or another tight heuristic) for longer horizons is the bottleneck to further scalability.

Together, these algorithmic improvements lead to the first optimal solutions for many problem horizons. In fact, for the vast majority of problems tested, we provide results for longer horizons than any previous work (the bold entries). These improvements are quite sub-stantial, especially given that lengthening the horizon by one increases the problem difficulty exponentially (cf. Table 1).

| $h$ | $\lvert BG^{h-1}\rvert$ | $\lvert cBG^t\rvert$ |
|---|---|---|
| | Dec-Tiger | |
| 2 | 4 | $1.0, 4.0$ |
| 3 | 16 | $1.0, 4.0, 9.0$ |
| 4 | 64 | $1.0, 4.0, 9.0, 23.14$ |
| 5 | 256 | $1.0, 4.0, 9.0, 16.0, 40.43$ |
| | FireFighting $\langle n_h = 3, n_f = 3\rangle$ | |
| 2 | 4 | $1.0, 4.0$ |
| 3 | 16 | $1.0, 4.0, 16.0$ |
| 4 | 64 | $1.0, 4.0, 16.0, 64.0$ |
| 5 | 256 | $1.0, 4.0, 16.0, 64.0, 256.0$ |
| 6 | 1024 | $1.0, 1.0, 2.0, 3.0, 6.0, 10.0$ |
| | GridSmall | |
| 2 | 4 | $1.0, 4.0$ |
| 3 | 16 | $1.0, 4.0, 10.50$ |
| 4 | 64 | $1.0, 4.0, 10.50, 20.0$ |
| | Hotel 1 | |
| 2 | 16 | $1.0, 4.0$ |
| 3 | 256 | $1.0, 4.0, 16.0$ |
| 4 | 4096 | $1.0, 4.0, 8.0, 16.0$ |
| 5 | 65536 | $1.0, 4.0, 4.0, 8.0, 16.0$ |
| 6 | $1.05e6$ | $1.0, 4.0, \ldots, 4.0, 8.0, 16.0$ |
| 7 | $1.68e7$ | $1.0, 4.0, \ldots, 4.0, 8.0, 16.0$ |
| 8 | $2.68e8$ | $1.0, 4.0, \ldots, 4.0, 8.0, 16.0$ |
| 9 | $4.29e9$ | $1.0, 4.0, \ldots, 4.0, 8.0, 16.0$ |
| | Cooperative Box Pushing | |
| 2 | 25 | $1.0, 4.0$ |
| 3 | 625 | $1.0, 4.0, 25.0$ |

| $h$ | $\lvert BG^{h-1}\rvert$ | $\lvert cBG^t\rvert$ |
|---|---|---|
| | Recycling Robots | |
| 2 | 4 | 1.0, remaining stages $\leq 4.0$ |
| 3 | 16 | 1.0, remaining stages $\leq 4.0$ |
| 4 | 64 | 1.0, remaining stages $\leq 4.0$ |
| 5 | 256 | 1.0, remaining stages $\leq 4.0$ |
| 10 | 262144 | 1.0, remaining stages $\leq 4.0$ |
| 15 | $2.68e8$ | 1.0, remaining stages $\leq 4.0$ |
| 18 | $1.72e10$ | 1.0, remaining stages $\leq 4.0$ |
| 20 | $2.75e11$ | 1.0, remaining stages $\leq 4.0$ |
| 30 | $2.88e17$ | 1.0, remaining stages $\leq 4.0$ |
| 40 | | 1.0, remaining stages $\leq 4.0$ |
| 50 | | 1.0, remaining stages $\leq 4.0$ |
| 60 | | 1.0, remaining stages $\leq 4.0$ |
| | BroadcastChannel | |
| 2 | 4 | 1.0 (for all $t$) |
| 3 | 16 | 1.0 (for all $t$) |
| 4 | 64 | 1.0 (for all $t$) |
| 5 | 256 | 1.0 (for all $t$) |
| 6 | 1024 | 1.0 (for all $t$) |
| 7 | 4096 | 1.0 (for all $t$) |
| 10 | 262144 | 1.0 (for all $t$) |
| 20 | $2.75e11$ | 1.0 (for all $t$) |
| 25 | $2.81e14$ | 1.0 (for all $t$) |
| 30 | $2.88e17$ | 1.0 (for all $t$) |
| 40 | | 1.0 (for all $t$) |
| 50 | | 1.0 (for all $t$) |
| 100 | | 1.0 (for all $t$) |
| 900 | | 1.0 (for all $t$) |

Table 3: Experimental results detailing the effectiveness of clustering. Listed are the size of the CBGs for $t = h - 1$ without clustering ($\lvert BG^{h-1}\rvert$), and the average CBG size for all stages with clustering ($\lvert cBG^t\rvert$).

### 5.2.1 Analysis of Clustering Histories

Table 3 provides additional details about the performance of GMAA*-IC, by listing the number of joint types in the GMAA*-IC search, $\lvert cBG^t\rvert$, for each stage $t$. These are averages since the algorithm forms CBGs for different past policies, leading to clusterings of different sizes.[19] To see the impact of clustering, the table also lists $\lvert BG^{h-1}\rvert$, the number of joint types in the CBGs constructed for the last stage without clustering, which is constant.

In Dec-Tiger, the time needed by GMAA*-IC is more than 3 orders of magnitude less than that of GMAA* for horizon $h = 4$. For $h = 5$, this test problem has $3.82e29$ joint policies, and no other method has been able to optimally solve it. GMAA*-IC, however, is able to do so in reasonable time. In Dec-Tiger, there are clear symmetries between the

19. Note that in some problem domains we report smaller clusterings than Oliehoek et al. (2009). Due to an implementation mistake, their clustering was overly conservative, and did not in all cases treat two histories as probabilistically equivalent, when in fact they were.

observations that allow for clustering, as demonstrated by Fig. 4. Another key property is that opening the door resets the problem, which may also facilitate clustering.

In FIREFIGHTING, for short planning horizons no lossless clustering is possible at any stage, and as such, the clustering incurs some overhead. However, GMAA*-IC is still faster than GMAA* because constructing the BGs using bootstrapping from the previous CBG takes less time than constructing a CBG from scratch. Interesting counterintuitive results occur for $h = 6$, which was solved within memory limits, in contrast to $h = 5$. In fact, using $Q_{MDP}$ we could compute optimal values $V^*$ for $h > 6$, and it turns out that these are equal to that for $h = 6$. The reason is that the optimal joint policy is guaranteed to extinguish all fires in 6 stages. For subsequent stages, all the rewards will be 0. While this itself does not influence clustering, the further analysis of Table 3 reveals that the CBG instances encountered during the $h = 6$ search happen to cluster much better than those in $h = 5$, which is possible because the heuristics vary with the horizon. In fact, $\boldsymbol{\pi}^*$ for $h = 6$ sends both agents to the middle house at $t = 0$, while for $h = 5$, agents are dispatched to different houses. When both agents fight fires at the same house, the fire is extinguished completely, and resulting joint observations do not provide any new information. As a result, different joint types lead to the same joint belief, which means they can be clustered. If agents visit different houses, their observations do convey information, leading to different possible joint beliefs (which cannot be clustered).

HOTEL 1 allows for a large amount of clustering, and GMAA*-IC outperforms GMAA* by a large margin, with the former reaching $h = 9$ and the latter $h = 2$. This problem is transition and observation independent (Becker, Zilberstein, Lesser, & Goldman, 2003; Nair, Varakantham, Tambe, & Yokoo, 2005; Varakantham, Marecki, Yabu, Tambe, & Yokoo, 2007), which facilitates clustering, as we further discuss in Section 5.5. Unlike methods specifically designed to exploit transition and observation independence, GMAA*-IC exploits this structure without requiring a predefined explicit representation of it. Further scalability is limited by the computation of the heuristic.

For BROADCASTCHANNEL, GMAA*-IC achieves an even more dramatic increase in performance, allowing the solution of up to horizon $h = 900$. Analysis reveals that the CBGs constructed for all stages are fully clustered: they contain only one type for each agent. The reason is as follows. When constructing a CBG for $t = 1$, there is only one joint type for the previous CBG so, given $\boldsymbol{\beta}^0$, the solution for the previous CBG, there is no uncertainty with respect to the previous joint action $\boldsymbol{a}^0$. The crucial property of BROADCASTCHANNEL is that the (joint) observation reveals nothing about the new state, but only about what joint action was taken (e.g., ‘collision’ if both agents chose to ‘send’). As a result, the different individual histories can be clustered. In a CBG constructed for stage $t = 2$, there is again only one joint type in the previous game. Therefore, given the past policy, the actions of the other agents can be perfectly predicted. Again the observation conveys no information so this process repeats. Thus, the problem has a special property which could be described as *non-observable given the past joint policy.* GMAA*-IC automatically exploits this property. Consequently, the time needed to solve each CBG does not grow with the horizon. The solution time, however, still increases super-linearly because of the increased amount of backtracking. As in FIREFIGHTING, performance is not monotonic in the planning horizon. In this case however, clustering is clearly not responsible for the difference. Rather, the only explanation is that for certain horizons, there are many near-optimal joint policies, leading to more backtracking and a higher search cost.

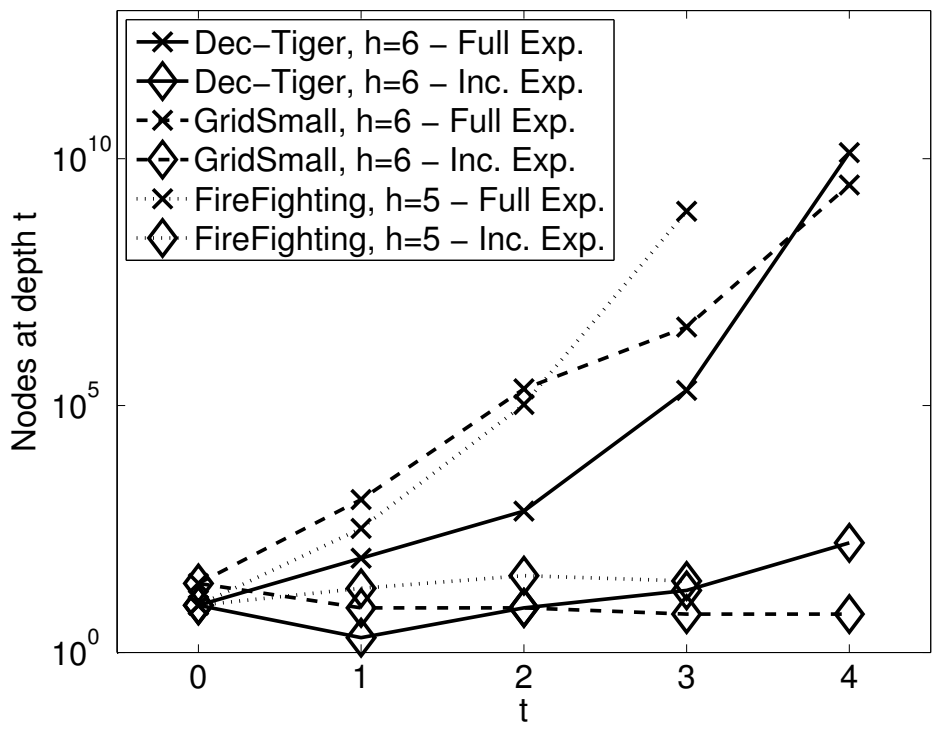

Figure 8: Number of expanded partial joint policies $\varphi^t$ for intermediate stages $t = 0, \ldots, h-2$ (in log scale).

### 5.2.2 Analysis of Incremental Expansion

In Dec-Tiger for $h = 5$, GMAA*-ICE achieves a speedup of three orders of magnitude and can compute a solution for $h = 6$, unlike GMAA*-IC. For GridSmall, it achieves a large speedup for $h = 4$ and fast solutions for $h = 5$ and 6, where GMAA*-IC runs out of memory. Similar positive results are obtained for FireFighting, Cooperative Box Pushing and Recycling Robots. In fact, when using $Q_{MDP}$, GMAA*-ICE is able to compute solutions well beyond $h = 1000$ for the FireFighting problem, which stands in stark contrast to GMAA*-IC that only computes solutions to $h = 3$ with this heuristic. Note that BroadcastChannel is the only problem for which GMAA*-IC is (slightly) faster than GMAA*-ICE. Because this problem exhibits clustering to a single joint type, the overhead of incremental expansion does not pay off.

To further analyze incremental expansion, we examined its impact on the number of nodes expanded for intermediate stages $t = 0, \ldots, h-2$. Fig. 8 shows the number of nodes expanded in GMAA*-ICE and the number that would be expanded for GMAA*-IC (which can be easily computed since they are search-tree equivalent). There is a clear relationship between the results from Fig. 8 and Table 2, illustrating, e.g., why GMAA*-IC runs out of memory on GridSmall $h = 6$. The plots confirm our hypothesis that, in practice, only a small number of child nodes are queried.

### 5.2.3 Analysis of Hybrid Heuristic Representation

Fig. 9 illustrates the memory requirements in terms of number of parameters (i.e., real numbers) for the tree, vector, and hybrid representations for $Q_{BG}$, where the latter is computed following Algorithm 9. Results for the vector representation are omitted when those representations grew beyond limits. The effectiveness of the vector pruning depends on the problem and the complexity of the value function, which can increase suddenly, as for instance happens in Fig. 9c. These results show that, for several benchmark Dec-POMDPs, the hybrid representation allows for significant savings in memory, allowing the computation of tight heuristics for longer horizons.

| $h$ | MILP | DP-LPC | DP-IPG | GMAA — $Q_{BG}$ | | |
|---|---|---|---|---|---|---|
| | | | | IC | ICE | heur |
| BroadcastChannel, ICE solvable to $h = 900$ | | | | | | |
| 2 | 0.38 | $\leq 0.01$ | 0.09 | $\leq 0.01$ | $\leq 0.01$ | $\leq 0.01$ |
| 3 | 1.83 | 0.50 | 56.66 | $\leq 0.01$ | $\leq 0.01$ | $\leq 0.01$ |
| 4 | 34.06 | $*$ | * | $\leq 0.01$ | $\leq 0.01$ | $\leq 0.01$ |
| 5 | 48.94 | | | $\leq 0.01$ | $\leq 0.01$ | $\leq 0.01$ |
| Dec-Tiger, ICE solvable to $h = 6$ | | | | | | |
| 2 | 0.69 | 0.05 | 0.32 | $\leq 0.01$ | $\leq 0.01$ | $\leq 0.01$ |
| 3 | 23.99 | 60.73 | 55.46 | $\leq 0.01$ | $\leq 0.01$ | $\leq 0.01$ |
| 4 | $*$ | $-$ | 2286.38 | 0.27 | $\leq 0.01$ | 0.03 |
| 5 | | | $-$ | 21.03 | 0.02 | 0.09 |
| FireFighting (2 agents, 3 houses, 3 firelevels), ICE solvable to $h \gg 1000$ | | | | | | |
| 2 | 4.45 | 8.13 | 10.34 | $\leq 0.01$ | $\leq 0.01$ | $\leq 0.01$ |
| 3 | $-$ | $-$ | 569.27 | 0.11 | 0.10 | 0.07 |
| 4 | | | $-$ | 950.51 | 1.00 | 0.65 |
| GridSmall, ICE solvable to $h = 6$ | | | | | | |
| 2 | 6.64 | 11.58 | 0.18 | 0.01 | $\leq 0.01$ | $\leq 0.01$ |
| 3 | $*$ | $-$ | 4.09 | 0.10 | $\leq 0.01$ | 0.42 |
| 4 | | | 77.44 | 1.77 | $\leq 0.01$ | 67.39 |
| Recycling Robots, ICE solvable to $h = 70$ | | | | | | |
| 2 | 1.18 | 0.05 | 0.30 | $\leq 0.01$ | $\leq 0.01$ | $\leq 0.01$ |
| 3 | * | 2.79 | 1.07 | $\leq 0.01$ | $\leq 0.01$ | $\leq 0.01$ |
| 4 | | 2136.16 | 42.02 | $\leq 0.01$ | $\leq 0.01$ | 0.02 |
| 5 | | $-$ | 1812.15 | $\leq 0.01$ | $\leq 0.01$ | 0.02 |
| Hotel 1, ICE solvable to $h = 9$ | | | | | | |
| 2 | 1.92 | 6.14 | 0.22 | $\leq 0.01$ | $\leq 0.01$ | 0.03 |
| 3 | 315.16 | 2913.42 | 0.54 | $\leq 0.01$ | $\leq 0.01$ | 1.51 |
| 4 | $-$ | $-$ | 0.73 | $\leq 0.01$ | $\leq 0.01$ | 3.74 |
| 5 | | | 1.11 | $\leq 0.01$ | $\leq 0.01$ | 4.54 |
| 9 | | | 8.43 | 0.02 | $\leq 0.01$ | 20.26 |
| 10 | | | 17.40 | # | # | |
| 15 | | | 283.76 | | | |
| Cooperative Box Pushing ($Q_{POMDP}$), ICE solvable to $h = 4$ | | | | | | |
| 2 | 3.56 | 15.51 | 1.07 | $\leq 0.01$ | $\leq 0.01$ | $\leq 0.01$ |
| 3 | 2534.08 | $-$ | 6.43 | 0.91 | 0.02 | 0.15 |
| 4 | $-$ | | 1138.61 | * | 328.97 | 0.63 |

Table 4: Comparison of runtimes with other methods. Total time of the GMAA* methods is given by taking the time from the method column ('IC' or 'ICE') and adding the heuristic computation time ('heur'). We use the following symbols: '$-$' memory limit violations, '*' time limit overruns, '#' heuristic computation exceeded memory or time limits.

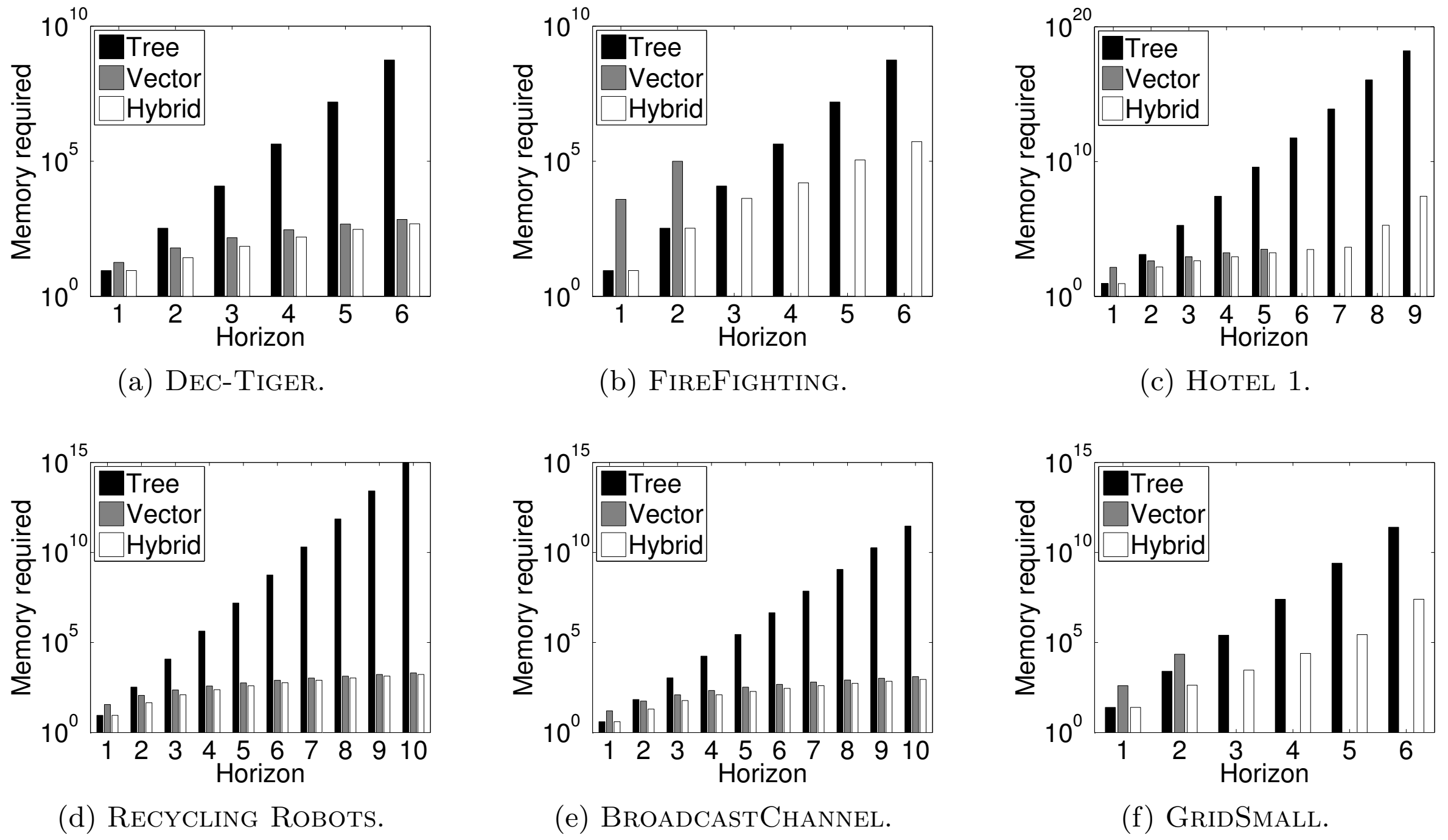

(a) Dec-Tiger. (b) FireFighting. (c) Hotel 1.

(d) Recycling Robots. (e) BroadcastChannel. (f) GridSmall.

Figure 9: Hybrid heuristic representation. The $y$-axis shows number of real numbers stored for different representations of $Q_{BG}$ for several benchmark problems (in log scale).

## 5.3 Comparing to Other Methods

In this section, we compare GMAA*-IC and GMAA*-ICE to other methods from the literature. We begin by comparing the runtimes of our methods against the following state-of-the-art optimal Dec-POMDP methods: **MILP**[20] (Aras & Dutech, 2010) converts the Dec-POMDP to a mixed integer linear program, for which numerous solvers are available. We have used MOSEK version 6.0. **DP-LPC**[21] (Boularias & Chaib-draa, 2008) performs dynamic programming with lossless policy compression, with CPLEX 12.4 as the LP solver. **DP-IPG** (Amato et al., 2009) performs exact dynamic programing with incremental policy

20. The results reported here deviate from those reported by Aras and Dutech (2010). For a number of problems, Aras et al. employed a solution method that solves the MILP as a series (a tree) of smaller MILPs by branching on the continuous realization weight variables for earlier stages. That is, for each past joint policy $\boldsymbol{\varphi}^t$ for some stage $t$, they solve a different MILP involving the subset of consistent sequences. Additionally, for FireFighting and GridSmall, we use the benchmark versions standard to the literature (Oliehoek, Spaan, & Vlassis, 2008; Amato et al., 2006), whereas Aras and Dutech (2010) use non-standard versions. This explains the difference between our results and the ones reported in their article (personal communication, Raghav Aras).
21. The goal of Boularias and Chaib-draa (2008) was to find non-dominated joint policies for all initial beliefs. The previously reported results concerned run-time to compute the non-dominated joint policies, without performing pruning on the full-length joint policies. In contrast, we report the time needed to compute the actual optimal Dec-POMDP policy (given $\boldsymbol{b}^0$). This additionally requires the final round of pruning and subsequently computing the value for each of the remaining joint policies for the initial belief. This additional overhead explains the differences in run time between what we report here and what was previously reported (personal communication, Abdeslam Boularias).

| Problem | $h$ | $m$ | $V_{\text{MBDP}}$ | $V^*$ |
|---|---|---|---|---|
| Dec-Tiger | 6 | 7 | 9.91 | 10.38 |
| Cooperative Box Pushing | 3 | 3 | 53.04 | 66.08 |
| GridSmall | 5 | 3 | 2.32 | 2.97 |

Table 5: Comparison of optimal ($V^*$) and approximate ($V_{\text{MBDP}}$) values.

generation that exploits known start state and knowledge about what states are reachable in doing the DP backup.

Table 4, which shows the results of the comparison, demonstrates that, in almost all cases, the total time of GMAA*-ICE (given by the sum of heuristic computation time and the time for the GMAA*-phase) is significantly less than that of any other state-of-the-art methods. Moreover, as demonstrated in Table 2, GMAA*-ICE can compute solutions for longer horizons for all these problems, except for Cooperative Box Pushing and Hotel 1.[22] For these problems, it is not possible to compute $Q_{\text{BG}}$ for longer horizons. Overcoming this problem could enable GMAA*-ICE to scale to further horizons as well.

The DP-LPC algorithm proposed by Boularias and Chaib-draa (2008) also improves the efficiency of optimal solutions by a form of compression. The performance of their algorithm, however, is weaker than that of GMAA*-IC. There are two main explanations for the performance difference. First, DP-LPC uses compression to more compactly represent the *values* for sets of useful sub-tree policies, by using sequence form representation. The policies themselves, however, are not compressed: they still specify actions for every possible observation history (for each policy it needs to select an exponential amount of sequences that make up that policy). Hence, it cannot compute solutions for long horizons. Second, GMAA*-IC can exploit knowledge of the initial state distribution $\boldsymbol{b}^0$.

Overall, GMAA*-ICE substantially improves the state-of-the-art in optimally solving Dec-POMDPs. Previous methods typically improved the feasible solution horizon by just one (or only provided speed-ups for horizons that could already be solved). By contrast, GMAA*-ICE dramatically extends the feasible solution horizon for many problems.

We also consider MBDP-based approaches, the leading family of approximate algorithms. Table 5, which reports the $V_{\text{MBDP}}$ values produced by PBIP-IPG (Amato et al., 2009) (with typical 'maxTrees' parameter setting $m$), demonstrates that the optimal solutions produced by GMAA*-IC or GMAA*-ICE are of higher quality. PBIP-IPG was chosen because all other MBDP algorithms with the same parameters achieve at most the same value. While not exhaustive, this comparison illustrates that even the best approximate Dec-POMDP methods in practice provide inferior joint policies on some problems. Conducting such analysis is possible only if optimal solutions can be computed. Clearly, the more data that becomes available, the more thorough the comparisons that can be made. Therefore, scalable optimal solution methods such as GMAA*-ICE are critical for improving these analyses.

| problem primitives | | | | num. $\boldsymbol{\pi}$ for $h$ | | |
|---|---|---|---|---|---|---|
| $n$ | $\lvert\mathcal{S}\rvert$ | $\lvert\boldsymbol{\mathcal{A}}\rvert$ | $\lvert\boldsymbol{\mathcal{O}}\rvert$ | 2 | 4 | 6 |
| 2 | 27 | 4 | 4 | 64 | 1.07$e$9 | 8.50$e$37 |
| 3 | 81 | 8 | 8 | 512 | 3.51$e$13 | 7.84$e$56 |
| 4 | 243 | 16 | 16 | 4.09$e$3 | 1.15$e$18 | 7.23$e$75 |
| 5 | 729 | 32 | 32 | 3.27$e$4 | 3.77$e$22 | 6.67$e$94 |
| 6 | 2187 | 64 | 64 | 2.62$e$5 | 9.80$e$55 | 6.15$e$113 |

Table 6: FIREFIGHTINGGRAPH: the number of joint policies for different numbers of agents and horizons, with 3 possible fire levels.

### 5.4 Scaling to More Agents

All of the benchmark problems in our results presented so far were limited to two agents. Here, we present a case study on FIREFIGHTINGGRAPH (Oliehoek, Spaan, Whiteson, & Vlassis, 2008), a variation of FIREFIGHTING allowing for more agents, in which each agent can only fight fires at two houses, instead of at all of them. Table 6 highlights the size of these problems, including the total number of joint policies for different horizons. We compared GMAA*, GMAA*-IC, GMAA*-ICE (all using a $\text{Q}_{\text{MDP}}$ heuristic), BRUTEFORCESEARCH, and DP-IPG, with a maximum run-time of 12 hours and running on an Intel Core i7 CPU, averaged over 10 runs. BRUTEFORCESEARCH is a simple optimal algorithm that enumerates and evaluates all joint policies, and was implemented in the same codebase as the GMAA* variations. DP-IPG results use the original implementation and were run on an Intel Xeon computer. Hence, while the timing results are not directly comparable, the overall trends are apparent. Also, since the DP-IPG implementation is limited to 2 agents, no results are shown for more agents.

Fig. 10 shows the computation times for FIREFIGHTINGGRAPH across different numbers of of agents and planning horizons, while Table 7 lists the optimal values obtained. As expected, the baseline BRUTEFORCESEARCH performs very poorly, only scaling beyond $h = 2$ for 2 agents, while DP-IPG can only reach $h = 4$. On the other hand, regular GMAA* performs relatively well, scaling to a maximum of 5 agents. However, GMAA*-IC and GMAA*-ICE improve the efficiency of GMAA* by 1–2 orders of magnitude. As such, they substantially outperform the other three methods, and scale up to 6 agents. The benefit of incremental expansion is clear for $n = 3,4$, where GMAA*-ICE can reach a higher horizon than GMAA*-IC. Hence, although this article focuses on scalability in the horizon, these results show that the methods we propose can also improve scalability in the number of agents.

### 5.5 Discussion

Overall, the empirical results demonstrate that incremental clustering and expansion offers dramatic performance gains on a diverse set of problems. In addition, the results on BROAD-

22. In HOTEL 1, DP-IPG performs particularly well because the problem structure has limited reachability. That is, each agent can fully observe its local state (but not that of the other agent) and in all local states except one there is one action that dominates all others. As a result, DP-IPG can generate a small number of possibly optimal policies.

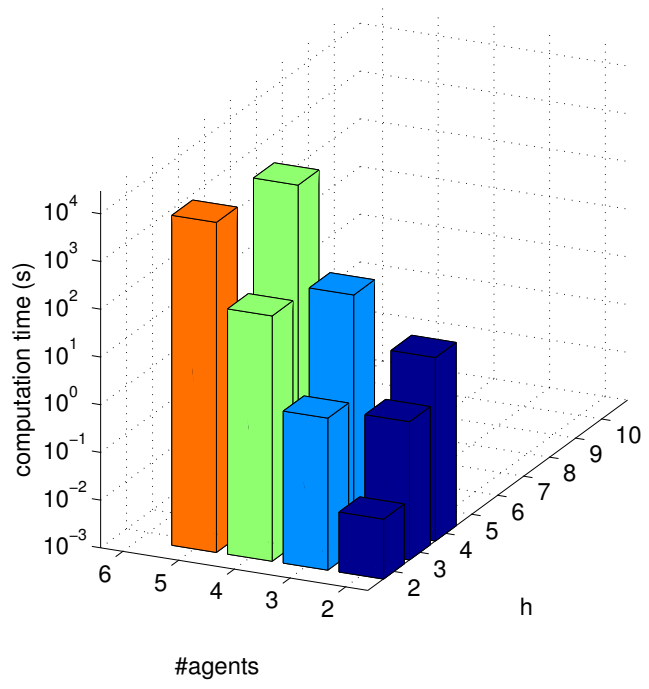

(a) GMAA* results.

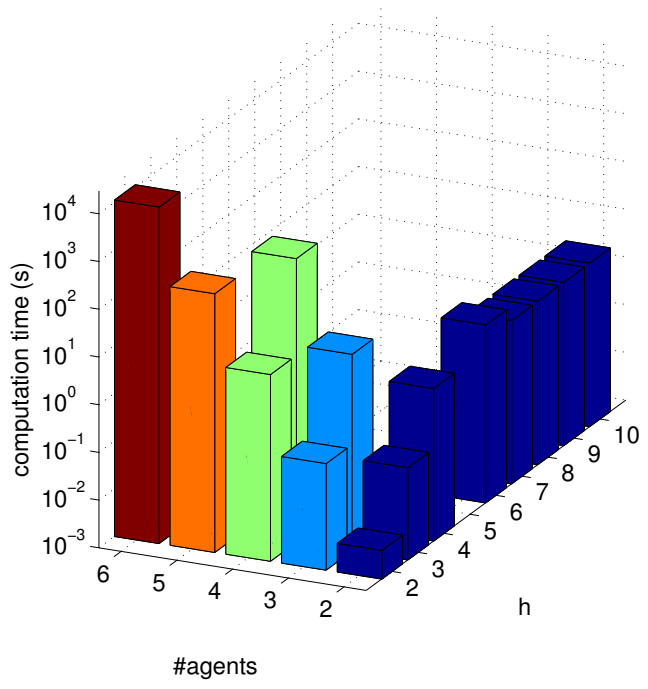

(b) GMAA*-IC results.

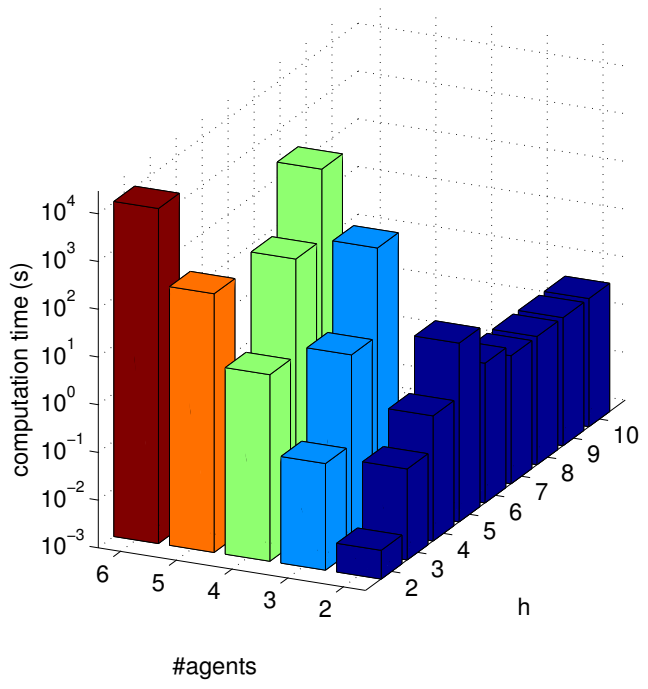

(c) GMAA*-ICE results.

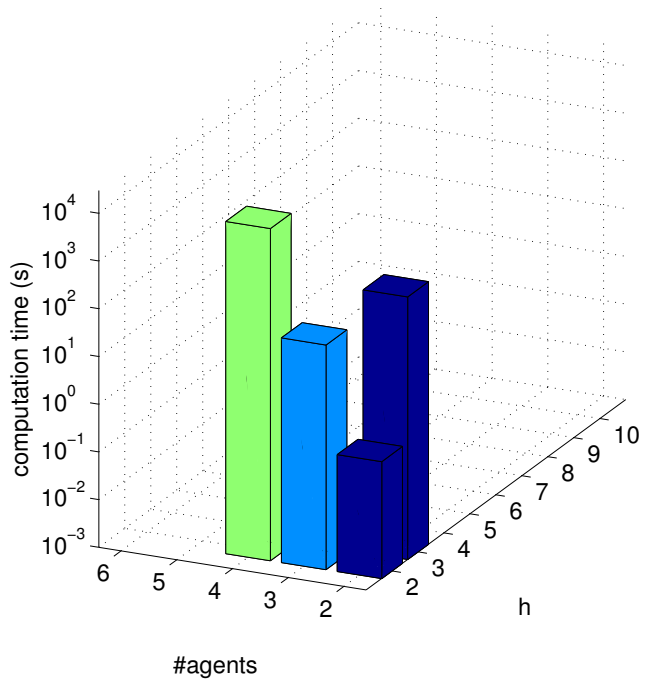

(d) BruteForceSearch results.

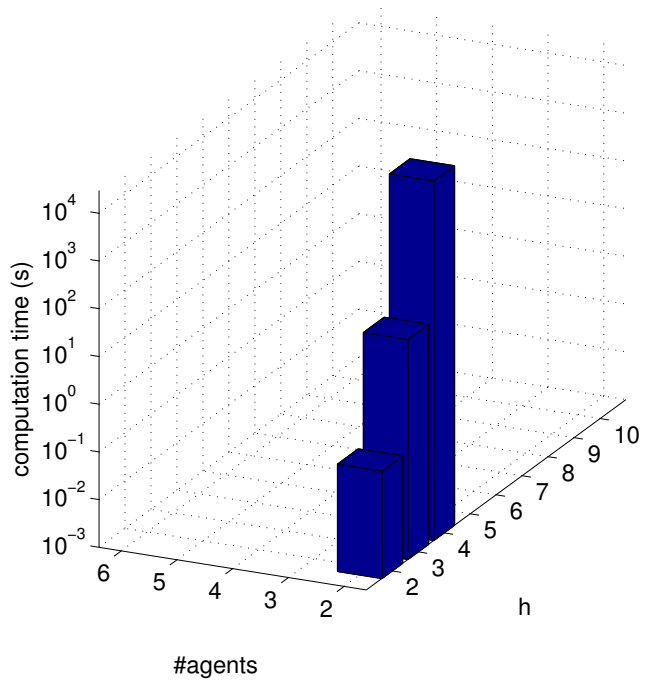

(e) DP-IPG results.

Figure 10: Comparison of GMAA*, GMAA*-IC, GMAA*-ICE, BruteForceSearch, and DP-IPG on the FireFightingGraph problem. Shown are computation time (in log scale) for various number of agents and horizons. Missing bars indicate that the method exceeded time or memory limits. However, the DP-IPG implementation only supports 2 agents.

| $h$ | $n = 2$ | $n = 3$ | $n = 4$ | $n = 5$ | $n = 6$ |
|---|---|---|---|---|---|
| 2 | −4.394252 | −5.213685 | −6.027319 | −6.846752 | −7.666185 |
| 3 | −5.806354 | −6.654551 | −7.391423 | | |
| 4 | −6.626555 | −7.472568 | −8.000277 | | |
| 5 | −7.093975 | | | | |
| $\geq 6$ | −7.196444 | | | | |

Table 7: Value $V^*$ of optimal solutions to the FireFightingGraph problem, for different horizons and numbers of agents.

castChannel illustrate a key advantage of our approach: when a problem possesses a property that makes a large amount of clustering possible, our clustering method exploits this property automatically, without requiring a predefined explicit representation of it.

Of course, not all problems admit great reductions via clustering. One domain property that allows for clustering is when the past joint policy encountered during GMAA* makes the observations superfluous, as with BroadcastChannel and FireFighting. In Dec-Tiger, we see that certain symmetries can lead to clustering. However clustering can occur even without these properties. In fact, for all problems and nearly all horizons that we tested, the size of the CBGs can be reduced. Moreover, in accordance with the analysis of Section 3.2, the improvements in planning efficiency are huge, even for modest reductions in CBG size.

One class of problems where we can say something a priori about the amount of clustering that is possible is the class of Dec-POMDPs with transition and observation independence (Becker et al., 2003). In such problems, the agents have local states and the transitions are independent, which for two agents can be expressed as

$$\Pr(s_1', s_2'|s_1, s_2, a_1, a_2) = \Pr(s_1'|s_1, a_1) \Pr(s_2'|s_2, a_2). \tag{5.1}$$

Similarly, the observations are assumed to be independent, which means that for each agent the observation probability depends only on its own action and local state: $\Pr(o_i|a_i, s_i')$. For such problems, the probabilistic equivalence criterion (3.1) factors too. In particular, due to transition and observation independence[23], (3.2) holds true for any $\vec{\theta}_i^a$,$\vec{\theta}_i^b$. Moreover, (3.3) factors as the product $\Pr(s_1, s_2|\vec{\theta}_1, \vec{\theta}_2) = \Pr(s_1|\vec{\theta}_1) \Pr(s_2|\vec{\theta}_2)$ and thus holds if $\Pr(s_1|\vec{\theta}_1^a) = \Pr(s_1|\vec{\theta}_1^b)$. That is, two histories can be clustered if they induce the same 'local belief'. As such, the size of the CBGs directly corresponds to the product of the number of reachable local beliefs. Since the transition and observation independent Hotel 1 problem is also locally fully observable, and the local state spaces consist of four states, there are only four possible local beliefs (which is consistent with the CBG size of 16 from Table 3). Moreover, we see that this maximum size is typically only reached at the end of search. This is because good policies defer sending customers to the hotel and thus do not visit local states where the hotel is filled in the earlier stages.

In more general classes of problems, even other weakly coupled models (e.g., Becker, Zilberstein, & Lesser, 2004; Witwicki & Durfee, 2010), the criterion (3.1) does not factor, and hence there is no direct correspondence to the number of local beliefs. As such, only by applying our clustering algorithm can we determine how well such a problem clusters. This is analogous to, e.g., state aggregation in MDPs (e.g., discussed in Givan, Dean, & Greig, 2003) where it is not known how to predict a priori how large a minimized model will be. Fortunately, our empirical results demonstrate that, in domains that admit little or no clustering, the overhead is small.

As expected, incremental expansion is most helpful for problems which do not allow for much clustering. However, the results for, e.g., Dec-Tiger illustrate that there is a limit to the amount of scaling that the method can currently provide. The bottleneck is the solution of the large CBGs for the later stages: the CBG solver has to solve these large CBGs when returning the first solution in order to guarantee optimality, but this takes takes a long time. We expect that further improvements to CBG solvers can directly add to the efficacy of incremental expansion.

Our experiments also clearly demonstrate that the Dec-POMDP complexity results, while important, are only worst-case results. In fact, the scalability demonstrated in our experiments clearly show that in many problems we successfully scale dramatically beyond what would be

23. This assumes no 'external' state variable $s_0$.

expected for a doubly-exponential dependence on the horizon. Even for the smallest problems, a doubly-exponential scaling in the horizon implies that it is impossible to compute solutions beyond $h = 4$ at all, as indicated by the following simple calculation: let $n = 2$, $|\mathcal{A}_i| = 2$ actions, $|\mathcal{O}_i| = 2|$ observations, then

$$|\mathcal{A}_i|^{(n*(|\mathcal{O}_i|^5))}/|\mathcal{A}_i|^{(n*(|\mathcal{O}_i|^4))} = 4.2950e9.$$

Thus, even in the simplest possible case, we see an increase of a factor $4.2950e09$ from $h = 4$ to $h = 5$. Similarly, the next increment, from $h = 5$ to $h = 6$, increases the size of the search space by a factor $1.8447e19$. However, our experiments clearly indicate that in almost all cases, things are not so dire. That is, even though matters look bleak in the light of the complexity results, we are in many cases able to perform substantially better than this worst case.

## 6. Related Work

In this section, we discuss a number of methods that are related to those proposed in this article. Some of these methods have already been discussed in earlier sections. In Section 3, we indicated that our clustering method is closely related to the approach of Emery-Montemerlo et al. (2005) but is also fundamentally different because our method is lossless. In Section 5.3, we discussed connections to the approach of Boularias and Chaib-draa (2008) which clusters policy values. This contrasts with our approach which clusters the histories and thus the policies themselves, leading to greater scalability.

In Section 3.1.2, we discussed the relationship between our notion of probabilistic equivalence (PE) and the multiagent belief. However, there is yet another notion of belief, employed in the JESP solution method (Nair et al., 2003), that is superficially more similar to the PE distribution. A 'JESP belief' for an AOH $\vec{\theta}_i$ is a probability distribution $\Pr(s,\vec{\mathbf{o}}_{\neq i}|\vec{\theta}_i, \boldsymbol{b}^0, \boldsymbol{\pi}_{\neq i})$ over states and observation histories of other agents given a (deterministic) *full* policy of all the other agents. It is a sufficient statistic, since it induces a multiagent belief, thus it also allows for the clustering of histories. The crucial difference with, and the utility of, PE lies in the fact that the PE criterion is specified over states and AOHs given only a *past* joint policy. That is, (3.1) does *not* induce a multiagent belief.

Our clustering approach also resembles a number of methods that employ other equivalence notions. First, several approaches exploit the notion of *behavioral equivalence* (Pynadath & Marsella, 2007; Zeng et al., 2011; Zeng & Doshi, 2012). They consider, from the perspective of a protagonist agent $i$, the possible models of another agent $j$. Since $j$ affects $i$ only through its actions, i.e., its *behavior*, agent $i$ can cluster together all the models of agent $j$ that lead to the same policy $\pi_j$ for that agent. That is, it can cluster all models of agent $j$ that are behaviorally equivalent. In contrast, we do not cluster models of other agents $j$, but histories of *this* agent $i$ if all the other agents, as well as the environment, are guaranteed to behave the same in expectation, thus leading to the same best response of agent $i$. That is, our method could be seen as clustering histories that are 'expected environmental behavior equivalent'.

The notion of *utility equivalence* (Pynadath & Marsella, 2007; Zeng et al., 2011) is closer to PE because it also takes into account the (value of the) best-response of agent $i$ (in particular, it clusters two models $m_j$ and $m'_j$ if using $BR(m_j)$—the best response against $m_j$— achieves the same value against $m'_j$). However, it remains a form of behavior equivalence in that it clusters models of other agents, not histories of the protagonist agent.

There are also connections between PE and work on influence-based abstraction (Becker et al., 2003; Witwicki & Durfee, 2010; Witwicki, 2011; Oliehoek et al., 2012), since the *influence* (or *point in parameter space*, Becker et al., 2003) is a compact representation of the other agents' policies. Models of the other agents can be clustered if they lead to the same influence on agent $i$. However, though more fine-grained, this is ultimately still a form of behavioral equivalence.

A final relation to our equivalence notion is the work by Dekel, Fudenberg, and Morris (2006), which constructs a distance measure and topology on the space of types with the goal of approximating the infinite universal type space (the space of all possible beliefs about beliefs about beliefs, etc.) for one-shot Bayesian games. Our setting, however, considers a simple finite type space where the types directly correspond to the private histories (in the form of AOHs) in a sequential problem. Thus, we do not need to approximate the universal type space; instead we want to know which histories lead to the same future dynamics from the perspective of an agent. Dekel et al.'s topology does not address this question.

Our incremental expansion technique is related to approaches extending $A^*$ to deal with large branching factors in the context of multiple sequence alignment (Ikeda & Imai, 1999; Yoshizumi, Miura, & Ishida, 2000). However, our approach is different because we do not discard unpromising nodes but rather provide a mechanism to generate only the necessary ones. Also, when proposing MAA*, Szer et al. (2005) developed a superficially similar approach that could be applied only to the last stage. In particular, they proposed generating the child nodes one by one, each time checking if a child is found with value equal to its parent's heuristic value. Since the value of such a child specifies a full policy, its value is a lower bound and therefore expansion of any remaining child nodes can be skipped. Unfortunately, a number of issues prevent this approach from providing substantial leverage in practice. First, it cannot be applied to intermediate stages $0 \leq t < h-1$ since no lower bound values for the expanded children are available. Second, in many problems it is unlikely that such a child node exists. Third, even if it does, Szer et al. did not specify an efficient way of finding it. Incremental expansion overcomes all of these issues, yielding an approach that, as our experiments demonstrate, significantly increases the size of the Dec-POMDPs that can be solved optimally.

This article focuses on optimal solutions for Dec-POMDPs over a finite horizon. As part of our evaluation, we compare against the MILP approach (Aras & Dutech, 2010), DP-ILP (Boularias & Chaib-draa, 2008) and DP-IPG (Amato et al., 2009), an extension of the exact dynamic programming algorithm (Hansen et al., 2004). Research on finite-horizon Dec-POMDPs has considered many other approaches such as bounded approximations (Amato, Carlin, & Zilberstein, 2007), locally optimal solutions (Nair et al., 2003; Varakantham, Nair, Tambe, & Yokoo, 2006) and approximate methods without guarantees (Seuken & Zilberstein, 2007b, 2007a; Carlin & Zilberstein, 2008; Eker & Akın, 2010; Oliehoek, Kooi, & Vlassis, 2008; Dibangoye et al., 2009; Kumar & Zilberstein, 2010b; Wu et al., 2010a; Wu, Zilberstein, & Chen, 2010b).

In particular, much research has considered the optimal and/or approximate solution of subclasses of Dec-POMDPs. One such subclass contains only Dec-POMDPs in which the agents have local states that other agents cannot influence. The resulting models, such as the TOI-Dec-MDP (Becker et al., 2003; Dibangoye, Amato, Doniec, & Charpillet, 2013) and ND-POMDP (Nair et al., 2005; Varakantham et al., 2007; Marecki, Gupta, Varakantham, Tambe, & Yokoo, 2008; Kumar & Zilberstein, 2009), can be interpreted as independent (PO)MDPs for each agent that are coupled through the reward function (and possibly an unaffectable state feature). On the other hand, event-driven interaction models (Becker et al., 2004) consider agents that have individual rewards but can influence each other's transitions.

More recently, models that allow for limited transition and reward dependence have been introduced. Examples are interaction-driven Markov games (Spaan & Melo, 2008), Dec-MDPs with sparse interactions (Melo & Veloso, 2011), distributed POMDPs with coordination locales (Varakantham et al., 2009; Velagapudi et al., 2011), event-driven interactions with complex rewards (EDI-CR) (Mostafa & Lesser, 2011), and transition decoupled Dec-POMDPs (Witwicki & Durfee, 2010; Witwicki, 2011). While the methods developed for these models often exhibit better scaling behavior than methods for standard Dec-(PO)MDPs, they typically are not suitable when agents have extended interactions, e.g., to collaborate in transporting an item. Also, there have been specialized models that consider the timing of actions whose ordering is already determined (Marecki & Tambe, 2007; Beynier & Mouaddib, 2011).

Another body of work addresses infinite-horizon problems (Amato, Bernstein, & Zilberstein, 2010; Amato, Bonet, & Zilberstein, 2010; Bernstein, Amato, Hansen, & Zilberstein, 2009; Kumar & Zilberstein, 2010a; Pajarinen & Peltonen, 2011), in which it is not possible to represent a policy as a tree. These approaches represent policies using finite-state controllers that are then optimized in various ways. Also, since the infinite-horizon case is undecidable (Bernstein et al., 2002), the approaches are approximate or optimal given a particular controller size. While there exists a boundedly optimal approach that can theoretically construct a controller within any $\epsilon$ of optimal, it is only feasible for very small problems or a large $\epsilon$ (Bernstein et al., 2009).

There has also been great interest in Dec-POMDPs that explicitly take into account communication. Some approaches try to optimize the meaning of communication actions without semantics (Xuan, Lesser, & Zilberstein, 2001; Goldman & Zilberstein, 2003; Spaan, Gordon, & Vlassis, 2006; Goldman, Allen, & Zilberstein, 2007) while others use fixed semantics (e.g., broadcasting the local observations) (Ooi & Wornell, 1996; Pynadath & Tambe, 2002; Nair et al., 2004; Roth et al., 2005; Oliehoek, Spaan, & Vlassis, 2007; Roth, Simmons, & Veloso, 2007; Spaan, Oliehoek, & Vlassis, 2008; Goldman & Zilberstein, 2008; Becker, Carlin, Lesser, & Zilberstein, 2009; Williamson, Gerding, & Jennings, 2009; Wu et al., 2011). Since models used in the first category (e.g., the Dec-POMDP-Com) can be converted to normal Dec-POMDPs (Seuken & Zilberstein, 2008), the contributions of this article are applicable to those settings.

Finally, there are numerous models closely related to Dec-POMDPs, such as POSGs (Hansen et al., 2004), interactive POMDPs (I-POMDPs) (Gmytrasiewicz & Doshi, 2005), and their graphical counterparts (Doshi, Zeng, & Chen, 2008). These models are more general in the sense that they consider self-interested settings where each agent has an individual reward function. I-POMDPs are conjectured to also require doubly exponential time (Seuken & Zilberstein, 2008). However, for the I-POMDP there have been a number of recent advances (Doshi & Gmytrasiewicz, 2009). The current paper makes a clear link between best-response equivalence of histories and the notion of best-response equivalence of beliefs in I-POMDPs. In particular, this article demonstrates that two PE action-observation histories (AOHs) induce, given only a *past* joint policy, a distribution over states and AOHs of other agents, and therefore will induce the same multiagent belief for *any* future policies of other agents. These induced multiagent beliefs, in turn, can be interpreted as special cases of I-POMDP beliefs where the model of the other agents are sub-intentional models in the form of a fixed policy tree. Rabinovich and Rosenschein (2005) introduced a method that, rather than optimizing the expected value of a joint policy, selects coordinated actions under uncertainty by tracking the dynamics of an environment. This approach, however, requires a model of the ideal system dynamics as input and in many problems, such as those considered in this article, identifying such dynamics is difficult.

## 7. Future Work

Several avenues for future work are made possible by the research presented in this article. Perhaps the most promising is the development of new approximate Dec-POMDP algorithms. While this article focused on optimal methods, GMAA*-ICE can also be seen as a framework for approximate methods. Such methods could be derived by limiting the amount of backtracking, employing approximate CBG solvers (Emery-Montemerlo, Gordon, Schneider, & Thrun, 2004; Kumar & Zilberstein, 2010b; Wu et al., 2010a), integrating GMAA* methods for factored Dec-POMDPs (Oliehoek, Spaan, Whiteson, & Vlassis, 2008; Oliehoek, 2010; Oliehoek et al., 2013), performing lossy clustering (Emery-Montemerlo, 2005; Wu et al., 2011) or using bounded approximations for the heuristics. In particular, it seems promising to combine approximate clustering with approximate factored GMAA* methods.

Lossy clustering could be achieved by generalizing the probabilistic equivalence criterion, which is currently so strict that little or no clustering may be possible in many problems. An obvious approach is to cluster histories for which the distributions over states and histories of other agents are merely similar, as measured by, e.g., Kullback-Leibler divergence. Alternately, histories could be clustered if they induce the same *individual belief* over states:

$$\Pr(s|\vec{\theta}_i) = \sum_{\vec{\boldsymbol{\theta}}_{\neq i}} \Pr(s,\vec{\boldsymbol{\theta}}_{\neq i}|\vec{\theta}_i). \tag{7.1}$$

While individual beliefs are not sufficient statistics for history, we hypothesize that they constitute effective metrics for approximate clustering. Since the individual belief simply marginalizes out the other agents' histories from the probabilities used in the probabilistic equivalence criterion, it is an intuitive heuristic metric for approximate clustering.

While this article focuses on increasing scalability with respect to the horizon, developing techniques to deal with larger number of agents is an important direction of future work. We plan to further explore performing GMAA* using factored representations (Oliehoek, Spaan, Whiteson, & Vlassis, 2008). In that previous work, we could only exploit the factorization at the last stage, since earlier stages required full expansions to guarantee optimality. However, for such larger problems, the number of joint BG policies (i.e., number of child nodes) is directly very large (earlier stages are more tightly coupled); therefore incremental expansion is crucial to improving the scalability of optimal solution methods with respect to the number of agents.

Another avenue for future work is to further generalize GMAA*-ICE. In particular, it may be possible to flatten the two nested $A^*$ searches into a single $A^*$ search. Doing so could lead to significant savings as it would obviate the need to solve an entire CBG before expanding the next one. In our work, we employed the plain $A^*$ algorithm as a basis, but a promising direction of future work is to investigate what $A^*$ enhancements from the literature (Edelkamp & Schrödl, 2012) can benefit GMAA* most. In particular, as we described in our experiments, different past joint policies can lead to CBGs of different sizes. One idea is to first expand parts of the search tree that lead to small CBGs, by biasing the selection operator (but not the pruning operator, so as to maintain optimality).

Yet another important direction for future work is the development of tighter heuristics. Though few researchers are addressing this topic, the results presented in this article underscore how important such heuristics are for solving larger problems. Currently, the heuristic is the bottleneck in four out of the seven problems we considered. Moreover, two of the problems where this is not the bottleneck can already be solved for long ($h > 50$) horizons. Therefore, we believe that computing tight heuristics for longer horizons is the single most important research direction for further improving the scalability of optimal Dec-POMDP solution methods with respect to the horizon.

A different direction is to employ our theoretical results on clustering beyond the Dec-POMDP setting to develop new solution methods for CBGs. For instance, a well-known method for computing a local optimum is *alternating maximization (AM)*: starting from an arbitrary joint policy, compute a best response for some agent given that other agents keep their policies fixed and then select another agent's policy to improve, etc. One idea is to start with a 'completely clustered' CBG, where all agents' types are clustered together and thus a random joint CBG policy has a simple form: each agent just selects a single action. Only when improving the policy of an agent do we consider all its actual possible types to compute its best response. Subsequently, we cluster together all types for which that agent selects the same action and proceed to the next agent. In addition, since our clustering results are not restricted to the collaborative setting, it may also be possible to employ them, using a similar approach, to develop new solution methods for general-payoff BGs.

Finally, two of our other contributions can have a significant impact beyond the problem of optimally solving Dec-POMDPs. First, the idea of incrementally expanding nodes introduced in GMAA*-ICE can be applied in other $A^*$ search methods. Incremental expansion is most useful when children can be generated in order of decreasing heuristic value without prohibitive computational effort, and in problems with a large branching factor such as multiple sequence alignment problems in computational biology (Carrillo & Lipman, 1988; Ikeda & Imai, 1999). Second, representing PWLC value functions as a hybrid of a tree and a set of vectors can have wider impact as well, e.g., in online search for POMDPs (Ross, Pineau, Paquet, & Chaib-draa, 2008).

## 8. Conclusions

This article presented a set of methods that advance the state-of-the-art in optimal solution methods for Dec-POMDPs. In particular, we presented several advances that aim to extend the horizon over which optimal solutions can be found. These advances build off the GMAA* heuristic search approach and include lossless incremental clustering of the CBGs solved by GMAA*, incremental expansion of nodes in the GMAA* search tree, and hybrid heuristic representations. We provided theoretical guarantees that, when a suitable heuristic is used, both incremental clustering and incremental expansion yield algorithms that are both complete and search equivalent. Finally, we presented extensive empirical results demonstrating that GMAA*-ICE can optimally solve Dec-POMDPs of unprecedented size. We significanty increase the planning horizons that can be tackled—in some cases by more than an order of magnitude. Given that an increase of the horizon by one results in an exponentially larger search space, this constitutes a very large improvement. Moreover, our techniques also im-prove scalability with respect to the number of agents, leading to the first ever solutions of general Dec-POMDPs with more than three agents. These results also demonstrated how optimal techniques can yield new insights about particular Dec-POMDPs, as incremental clustering revealed properties of BroadcastChannel that make it much easier to solve. In addition to facilitating optimal solutions, we hope these advances will inspire new principled approximation methods, as incremental clustering has already done (Wu et al., 2011), and enable them to be meaningfully benchmarked.

## Acknowledgments

We thank Raghav Aras and Abdeslam Boularias for making their code available to us. Research supported in part by AFOSR MURI project #FA9550-09-1-0538 and in part by NWO CATCH project #640.005.003. M.S. is funded by the FP7 Marie Curie Actions Individual Fellowship #275217 (FP7-PEOPLE-2010-IEF).

## Appendix A. Appendix

### A.1 Auxiliary algorithms

Algorithm 12 implements the `BestJointPolicyAndValue` function, which prunes all child nodes that are not fully specified. Algorithm 13 generates all children of a particular CBG.

**Algorithm 12** `BestJointPolicyAndValue`($\mathcal{Q}_{\texttt{Expand}}$): Prune fully expanded nodes from a set of nodes $\mathcal{Q}_{\texttt{Expand}}$ returning only the best one and its value.

**Input:** $\mathcal{Q}_{\texttt{Expand}}$ a set of nodes for fully specified joint policies.
**Output:** the best full joint policy in the input set and its value.

$v^* = -\infty$
**for** $q \in \mathcal{Q}_{\texttt{Expand}}$ **do**
  $\mathcal{Q}_{\texttt{Expand}}$.Remove($q$)
  $\langle \boldsymbol{\pi}, \hat{v} \rangle \leftarrow q$
  **if** $v > v^*$ **then**
    $v^* \leftarrow v$
    $\boldsymbol{\pi}^* \leftarrow \boldsymbol{\pi}$
  **end if**
**end for**
**return** $\langle \boldsymbol{\pi}^*, v^* \rangle$

### A.2 Detailed GMAA*-ICE algorithm

The complete GMAA*-ICE algorithm is shown in Algorithm 14.

### A.3 Computation of $V^{0...t-1}(\boldsymbol{\varphi}^t)$

The quantity $V^{0...t-1}(\boldsymbol{\varphi}^t)$ is defined recursively via:

$$V^{0...t-1}(\boldsymbol{\varphi}^t) = V^{0...t-2}(\boldsymbol{\varphi}^{t-1}) + \mathbf{E}_{s^{t-1},\vec{\boldsymbol{\theta}}^{t-1}}[R(s^{t-1},\boldsymbol{\delta}^{t-1}(\vec{\boldsymbol{\theta}}^{t-1})) \mid \boldsymbol{b}^0, \boldsymbol{\varphi}^t]. \tag{A.1}$$

**Algorithm 13** `GenerateAllChildrenForCBG`$(B(\boldsymbol{\varphi}^t))$.

**Input:** CBG $B(\boldsymbol{\varphi}^t)$.
**Output:** $\mathcal{Q}_{\texttt{Expand}}$ the set containing all expanded child nodes for this CBG.

$\mathcal{Q}_{\texttt{Expand}} \leftarrow \{\}$
**for** all joint CBG policies $\boldsymbol{\beta}$ for $B$ **do**
  $\widehat{V}(\boldsymbol{\beta}) \leftarrow \sum_{\boldsymbol{\theta}} \Pr(\boldsymbol{\theta}) u(\boldsymbol{\theta},\boldsymbol{\beta}(\boldsymbol{\theta}))$
  $\boldsymbol{\varphi}^{t+1} \leftarrow (\boldsymbol{\varphi}^t, \boldsymbol{\beta}^t)$ {create partial joint policy}
  $\widehat{V}(\boldsymbol{\varphi}^{t+1}) \leftarrow V^{0\dots t-1}(\boldsymbol{\varphi}^t) + \widehat{V}(\boldsymbol{\beta}^t)$ {compute heuristic value}
  $q' \leftarrow \langle \boldsymbol{\varphi}^{t+1}, \widehat{V}(\boldsymbol{\varphi}^{t+1}) \rangle$ {create child node}
  $\mathcal{Q}_{\texttt{Expand}}$.Insert$(q')$
**end for**
**return** $\mathcal{Q}_{\texttt{Expand}}$

The expectation is taken with respect to the joint probability distribution over states and joint AOHs that is induced by $\boldsymbol{\varphi}^t$:

$$\Pr(s^t,\vec{\boldsymbol{\theta}}^t|\boldsymbol{b}^0,\boldsymbol{\varphi}^t) = \sum_{s^{t-1}\in\mathcal{S}} \Pr(\boldsymbol{o}^t|\boldsymbol{a}^{t-1},s^t) \Pr(s^t|s^{t-1},\boldsymbol{a}^{t-1}) \Pr(\boldsymbol{a}^{t-1}|\boldsymbol{\varphi}^t,\vec{\boldsymbol{\theta}}^{t-1}) \Pr(s^{t-1},\vec{\boldsymbol{\theta}}^{t-1}|\boldsymbol{b}^0,\boldsymbol{\varphi}^t). \tag{A.2}$$

Here, $\vec{\boldsymbol{\theta}}^t = (\vec{\boldsymbol{\theta}}^{t-1},\boldsymbol{a}^{t-1},\boldsymbol{o}^t)$ and $\Pr(\boldsymbol{a}^{t-1}|\boldsymbol{\varphi}^t,\vec{\boldsymbol{\theta}}^{t-1})$ is the probability that $\boldsymbol{\varphi}^t$ specifies $\boldsymbol{a}^{t-1}$ for AOH $\vec{\boldsymbol{\theta}}^{t-1}$ (which is 0 or 1 in case of deterministic past joint policy $\boldsymbol{\varphi}^t$).

### A.4 Proofs

Proof of Theorem 1

Substituting (2.9) in (2.7) yields

$$\begin{aligned}
\widehat{V}(\boldsymbol{\beta}) = \widehat{V}(\boldsymbol{\delta}^t) &= \sum_{\vec{\boldsymbol{\theta}}^t} \Pr(\vec{\boldsymbol{\theta}}^t|\boldsymbol{b}^0,\boldsymbol{\varphi}^t)\widehat{Q}(\vec{\boldsymbol{\theta}}^t,\boldsymbol{\delta}^t(\vec{\boldsymbol{\theta}}^t)) \\
&= \sum_{\vec{\boldsymbol{\theta}}^t} \Pr(\vec{\boldsymbol{\theta}}^t|\boldsymbol{b}^0,\boldsymbol{\varphi}^t)\Big(\mathbf{E}_{s^t}[R(s^t,\boldsymbol{\delta}^t(\vec{\boldsymbol{\theta}}^t)) \mid \vec{\boldsymbol{\theta}}^t] + \mathbf{E}_{\vec{\boldsymbol{\theta}}^{t+1}}[\widehat{V}(\vec{\boldsymbol{\theta}}^{t+1}) \mid \vec{\boldsymbol{\theta}}^t, \boldsymbol{\delta}^t(\vec{\boldsymbol{\theta}}^t)]\Big) \\
&= \mathbf{E}_{s^t,\vec{\boldsymbol{\theta}}^t}[R(s^t,\boldsymbol{\delta}^t(\vec{\boldsymbol{\theta}}^t) \mid \boldsymbol{b}^0, \boldsymbol{\varphi}^t] + \mathbf{E}_{\vec{\boldsymbol{\theta}}^{t+1}}[\widehat{V}(\vec{\boldsymbol{\theta}}^{t+1}) \mid \boldsymbol{b}^0, \boldsymbol{\varphi}^t, \boldsymbol{\delta}^t] \\
&\geq \mathbf{E}_{s^t,\vec{\boldsymbol{\theta}}^t}[R(s^t,\boldsymbol{\delta}^t(\vec{\boldsymbol{\theta}}^t) \mid \boldsymbol{b}^0, \boldsymbol{\varphi}^t] + \mathbf{E}_{\vec{\boldsymbol{\theta}}^{t+1}}[Q_{\boldsymbol{\pi}^*}(\vec{\boldsymbol{\theta}}^{t+1}, \boldsymbol{\pi}^*(\vec{\boldsymbol{\theta}}^{t+1})) \mid \boldsymbol{b}^0, \boldsymbol{\varphi}^{t+1} = (\boldsymbol{\varphi}^t, \boldsymbol{\delta}^t)] \\
&= \mathbf{E}_{s^t,\vec{\boldsymbol{\theta}}^t}[R(s^t,\boldsymbol{\delta}^t(\vec{\boldsymbol{\theta}}^t) \mid \boldsymbol{b}^0, \boldsymbol{\varphi}^t] + H^{*,t+1\dots h-1}(\boldsymbol{\varphi}^{t+1}),
\end{aligned}$$

where $H^*$ is an optimal admissible heuristic. Substituting this into (2.8) we obtain

$$\begin{aligned}
\widehat{V}(\boldsymbol{\varphi}^{t+1} = (\boldsymbol{\varphi}^t, \boldsymbol{\delta}^t)) &= V^{0\dots t-1}(\boldsymbol{\varphi}^t) + \mathbf{E}_{s^t,\vec{\boldsymbol{\theta}}^t}[R(s^t,\boldsymbol{\delta}^t(\vec{\boldsymbol{\theta}}^t) \mid \boldsymbol{b}^0, \boldsymbol{\varphi}^t] + \mathbf{E}_{\vec{\boldsymbol{\theta}}^{t+1}}[\widehat{V}(\vec{\boldsymbol{\theta}}^{t+1}) \mid \boldsymbol{b}^0, \boldsymbol{\varphi}^t, \boldsymbol{\delta}^t] \\
&\geq V^{0\dots t-1}(\boldsymbol{\varphi}^t) + \mathbf{E}_{s^t,\vec{\boldsymbol{\theta}}^t}[R(s^t,\boldsymbol{\delta}^t(\vec{\boldsymbol{\theta}}^t)) \mid \boldsymbol{b}^0, \boldsymbol{\varphi}^{t+1}] + H^{*,t+1\dots h-1}(\boldsymbol{\varphi}^{t+1}) \\
\{\text{via (A.1)}\} \quad &= V^{0\dots t}(\boldsymbol{\varphi}^{t+1}) + H^{*,t+1\dots h-1}(\boldsymbol{\varphi}^{t+1}),
\end{aligned}$$

which demonstrates that the heuristic value $\widehat{V}(\boldsymbol{\varphi}^t)$ used by GMAA* via CBGs using heuristic of a form (2.9) is admissible, as it is lower bounded by the actual value for the first $t$ plus an admissible heuristic. Since it performs heuristic search with this admissible heuristic, this algorithm is also complete. □

**Algorithm 14** GMAA*-ICE

1: $\underline{\mathrm{v}}^{GMAA} \leftarrow -\infty$
2: $\boldsymbol{\varphi}^0 \leftarrow ()$
3: $\hat{v} \leftarrow +\infty$
4: $q^0 \leftarrow \langle \boldsymbol{\varphi}^0, \hat{v} \rangle$
5: $\mathrm{L}^{\mathrm{IE}} \leftarrow \{q^0\}$
6: **repeat**
7:     $q \leftarrow$ `Select`$(\mathrm{L}^{\mathrm{IE}})$     $\{q = \langle \boldsymbol{\varphi}^t, \hat{v} \rangle\}$
8:     $\mathrm{L}^{\mathrm{IE}}.\mathrm{pop}(q)$
9:     **if** IsPlaceholder$(q)$ **then**
10:         $B(\boldsymbol{\varphi}^t) \leftarrow \boldsymbol{\varphi}^t.\mathrm{CBG}$     {reuse stored CBG}
11:     **else**
12:         {Construct extended BG and solver:}
13:         $B(\boldsymbol{\varphi}^{t-1}) \leftarrow \boldsymbol{\varphi}^{t-1}.\mathrm{CBG}$     {note $\boldsymbol{\varphi}^t = (\boldsymbol{\varphi}^{t-1}, \boldsymbol{\beta}^{t-1})$}
14:         $B(\boldsymbol{\varphi}^t) \leftarrow$ ConstructExtendedBG$(B(\boldsymbol{\varphi}^{t-1}),\boldsymbol{\beta}^{t-1})$
15:         $B(\boldsymbol{\varphi}^t) \leftarrow$ ClusterBG$(B(\boldsymbol{\varphi}^t))$
16:         $B(\boldsymbol{\varphi}^t)$.Solver $\leftarrow$ CreateSolver$(B(\boldsymbol{\varphi}^t))$
17:         $\boldsymbol{\varphi}^t.\mathrm{CBG} \leftarrow B(\boldsymbol{\varphi}^t)$
18:     **end if**
19:     {Expand a single child:}
20:     $\underline{\mathrm{v}}^{CBG} = \underline{\mathrm{v}}^{GMAA} - V^{0\ldots(t-1)}(\boldsymbol{\varphi}^t)$
21:     $\bar{\mathrm{v}}^{CBG} = +\infty$
22:     **if** last stage $t = h-1$ **then**
23:         $\bar{\mathrm{v}}^{CBG} = \widehat{V}(\boldsymbol{\varphi}^{h-1}) - V^{0\ldots(h-2)}(\boldsymbol{\varphi}^{h-1})$
24:     **end if**
25:     $\langle \boldsymbol{\beta}^t, \widehat{V}(\boldsymbol{\beta}^t) \rangle \leftarrow B(\boldsymbol{\varphi}^t)$.Solver.NextSolution$(\underline{\mathrm{v}}^{CBG},\bar{\mathrm{v}}^{CBG})$
26:     **if not** $\boldsymbol{\beta}^t$ **then**
27:         {fully expanded: no solution s.t. $V(\boldsymbol{\beta}^{h-1}) \geq \underline{\mathrm{v}}^{CBG}$}
28:         delete $q$ (and its CBG + solver)
29:         continue     {(i.e. goto line 8)}
30:     **end if**
31:     $\boldsymbol{\varphi}^{t+1} \leftarrow (\boldsymbol{\varphi}^t, \boldsymbol{\beta}^t)$
32:     $\widehat{V}(\boldsymbol{\varphi}^{t+1}) \leftarrow V^{0\ldots t-1}(\boldsymbol{\varphi}^t) + \widehat{V}(\boldsymbol{\beta}^t)$
33:     **if** last stage $t = h-1$ **then**
34:         {Note that $\boldsymbol{\pi} = \boldsymbol{\varphi}^{t+1}, V(\boldsymbol{\pi}) = \widehat{V}(\boldsymbol{\varphi}^{t+1})$ }
35:         **if** $V(\boldsymbol{\pi}) > \underline{\mathrm{v}}^{GMAA}$ **then**
36:             $\underline{\mathrm{v}}^{GMAA} \leftarrow V(\boldsymbol{\pi})$     {found new lower bound}
37:             $\boldsymbol{\pi}^\star \leftarrow \boldsymbol{\pi}$
38:             $\mathrm{L}^{\mathrm{IE}}.\mathrm{prune}(\underline{\mathrm{v}}^{GMAA})$
39:         **end if**
40:         delete $q$ (and its CBG + solver)
41:     **else**
42:         $q' \leftarrow \langle \boldsymbol{\varphi}^{t+1}, \widehat{V}(\boldsymbol{\varphi}^{t+1}) \rangle$
43:         $\mathrm{L}^{\mathrm{IE}}.\mathrm{insert}(q')$
44:         $q \leftarrow \langle \boldsymbol{\varphi}^t, \widehat{V}(\boldsymbol{\varphi}^{t+1}) \rangle$     { Update parent node $q$, which now is a placeholder }
45:         $\mathrm{L}^{\mathrm{IE}}.\mathrm{insert}(q)$
46:     **end if**
47: **until** $\mathrm{L}^{\mathrm{IE}}$ is empty

Proof of Lemma 1

*Proof.* Assume an arbitrary $a_i^t$,$o_i^{t+1}$, $\boldsymbol{\delta}_{\neq i}^t$,$s^{t+1}$ and $\vec{\boldsymbol{\theta}}_{\neq i}^{t+1} = (\vec{\boldsymbol{\theta}}_{\neq i}^t,\boldsymbol{a}_{\neq i}^t,\boldsymbol{o}_{\neq i}^{t+1}))$. We have that

$$\begin{aligned}
&\Pr(s^{t+1},\vec{\boldsymbol{\theta}}_{\neq i}^{t+1},o_i^{t+1}|\vec{\theta}_i^{a,t},a_i^t,\boldsymbol{\delta}_{\neq i}^t)\\
&= \sum_{s^t} \Pr(o_i^{t+1},\boldsymbol{o}_{\neq i}^{t+1}|a_i^t,\boldsymbol{a}_{\neq i}^t,s^{t+1}) \Pr(s^{t+1}|s^t,a_i^t,\boldsymbol{a}_{\neq i}^t) \Pr(\boldsymbol{a}_{\neq i}^t|\vec{\boldsymbol{\theta}}_{\neq i}^t,\boldsymbol{\delta}_{\neq i}^t) \Pr(s^t,\vec{\boldsymbol{\theta}}_{\neq i}^t|\vec{\theta}_i^{a,t})\\
&= \sum_{s^t} \Pr(o_i^{t+1},\boldsymbol{o}_{\neq i}^{t+1}|a_i^t,\boldsymbol{a}_{\neq i}^t,s^{t+1}) \Pr(s^{t+1}|s^t,a_i^t,\boldsymbol{a}_{\neq i}^t) \Pr(\boldsymbol{a}_{\neq i}^t|\vec{\boldsymbol{\theta}}_{\neq i}^t,\boldsymbol{\delta}_{\neq i}^t) \Pr(s^t,\vec{\boldsymbol{\theta}}_{\neq i}^t|\vec{\theta}_i^{b,t})\\
&= \Pr(s^{t+1},\vec{\boldsymbol{\theta}}_{\neq i}^{t+1},o_i^{t+1}|\vec{\theta}_i^{b,t},a_i^t,\boldsymbol{\delta}_{\neq i}^t)
\end{aligned}$$

Because we assumed an arbitrary $s^{t+1}$,$\vec{\boldsymbol{\theta}}_{\neq i}^{t+1}$,$o_i^{t+1}$, we have that

$$\forall_{s^{t+1},\vec{\boldsymbol{\theta}}_{\neq i}^{t+1},o_i^{t+1}} \quad \Pr(s^{t+1},\vec{\boldsymbol{\theta}}_{\neq i}^{t+1},o_i^{t+1}|\vec{\theta}_i^{a,t},a_i^t,\boldsymbol{\delta}_{\neq i}^t) = \Pr(s^{t+1},\vec{\boldsymbol{\theta}}_{\neq i}^{t+1},o_i^{t+1}|\vec{\theta}_i^{b,t},a_i^t,\boldsymbol{\delta}_{\neq i}^t) \tag{A.3}$$

In general we have that

$$\begin{aligned}
\Pr(s^{t+1},\vec{\boldsymbol{\theta}}_{\neq i}^{t+1}|\vec{\theta}_i^t,a_i^t,o_i^{t+1},\boldsymbol{\delta}_{\neq i}^t) &= \frac{\Pr(s^{t+1},\vec{\boldsymbol{\theta}}_{\neq i}^{t+1},o_i^{t+1}|\vec{\theta}_i^t,a_i^t,\boldsymbol{\delta}_{\neq i}^t)}{\Pr(o_i^{t+1}|\vec{\theta}_i^t,a_i^t,\boldsymbol{\delta}_{\neq i}^t)}\\
&= \frac{\Pr(s^{t+1},\vec{\boldsymbol{\theta}}_{\neq i}^{t+1},o_i^{t+1}|\vec{\theta}_i^t,a_i^t,\boldsymbol{\delta}_{\neq i}^t)}{\sum_{s^{t+1},\vec{\boldsymbol{\theta}}_{\neq i}^{t+1}} \Pr(s^{t+1},\vec{\boldsymbol{\theta}}_{\neq i}^{t+1},o_i^{t+1}|\vec{\theta}_i^t,a_i^t,\boldsymbol{\delta}_{\neq i}^t)}
\end{aligned}$$

Now, because of (A.3), both the numerator and denominator are the same when substituting $\vec{\theta}_i^{a,t}$,$\vec{\theta}_i^{b,t}$ in this equation. Consequently, we can conclude that

$$\Pr(s^{t+1},\vec{\boldsymbol{\theta}}_{\neq i}^{t+1}|\vec{\theta}_i^{a,t},a_i^t,o_i^{t+1},\boldsymbol{\delta}_{\neq i}^t) = \Pr(s^{t+1},\vec{\boldsymbol{\theta}}_{\neq i}^{t+1}|\vec{\theta}_i^{b,t},a_i^t,o_i^{t+1},\boldsymbol{\delta}_{\neq i}^t)$$

Finally, because $a_i^t$, $o_i^{t+1}$, $\boldsymbol{\delta}_{\neq i}^t$,$s^{t+1}$, and $\vec{\boldsymbol{\theta}}_{\neq i}^{t+1}$ were all arbitrarily chosen, we can conclude that (3.4) holds. □

Proof of Lemma 2

*Proof.* Assume an arbitrary $\boldsymbol{\pi}_{\neq i}$,$s$ and $\boldsymbol{\gamma}_{\neq i}$, then we have

$$\begin{aligned}
b_i(s,\boldsymbol{\gamma}_{\neq i}|\vec{\theta}_i^a, \boldsymbol{\pi}_{\neq i}) &\triangleq \Pr(s,\boldsymbol{\gamma}_{\neq i}|\vec{\theta}_i^a, \boldsymbol{\pi}_{\neq i},\boldsymbol{b}^0)\\
&= \sum_{\vec{\boldsymbol{\theta}}_{\neq i}} \Pr(s,\boldsymbol{\gamma}_{\neq i},\vec{\boldsymbol{\theta}}_{\neq i}|\vec{\theta}_i^a,\boldsymbol{\pi}_{\neq i},\boldsymbol{b}^0)\\
\text{\{factoring the joint distribution\}} \quad &= \sum_{\vec{\boldsymbol{\theta}}_{\neq i}} \Pr(s,\vec{\boldsymbol{\theta}}_{\neq i}|\vec{\theta}_i^a,\boldsymbol{\pi}_{\neq i},\boldsymbol{b}^0) \Pr(\boldsymbol{\gamma}_{\neq i}|s,\vec{\boldsymbol{\theta}}_{\neq i}, \vec{\theta}_i^a,\boldsymbol{\pi}_{\neq i},\boldsymbol{b}^0)
\end{aligned}$$

$$
\begin{aligned}
\{\boldsymbol{\gamma}_{\neq i} \text{ only depends on } \vec{\boldsymbol{\theta}}_{\neq i},\boldsymbol{\pi}_{\neq i}\} &= \sum_{\vec{\boldsymbol{\theta}}_{\neq i}} \Pr(s,\vec{\boldsymbol{\theta}}_{\neq i}|\vec{\theta}_i^a,\boldsymbol{\pi}_{\neq i},\boldsymbol{b}^0)\Pr(\boldsymbol{\gamma}_{\neq i}|\vec{\boldsymbol{\theta}}_{\neq i},\boldsymbol{\pi}_{\neq i}) \\
\{\ s,\vec{\boldsymbol{\theta}}_{\neq i} \text{ only depend on } \boldsymbol{\varphi}_{\neq i}\ \} &= \sum_{\vec{\boldsymbol{\theta}}_{\neq i}} \Pr(s,\vec{\boldsymbol{\theta}}_{\neq i}|\vec{\theta}_i^a,\boldsymbol{\varphi}_{\neq i},\boldsymbol{b}^0)\Pr(\boldsymbol{\gamma}_{\neq i}|\vec{\boldsymbol{\theta}}_{\neq i},\boldsymbol{\pi}_{\neq i}) \\
\{\text{due to PE}\} &= \sum_{\vec{\boldsymbol{\theta}}_{\neq i}} \Pr(s,\vec{\boldsymbol{\theta}}_{\neq i}|\vec{\theta}_i^b,\boldsymbol{\varphi}_{\neq i},\boldsymbol{b}^0)\Pr(\boldsymbol{\gamma}_{\neq i}|\vec{\boldsymbol{\theta}}_{\neq i},\boldsymbol{\pi}_{\neq i}) \\
&= [...] = \Pr(s,\boldsymbol{\gamma}_{\neq i}|\vec{\theta}_i^b,\boldsymbol{\pi}_{\neq i},\boldsymbol{b}^0) = b_i(s,\boldsymbol{\gamma}_{\neq i}|\vec{\theta}_i^b,\boldsymbol{\pi}_{\neq i})
\end{aligned}
$$

We can conclude this holds for all $\boldsymbol{\pi}_{\neq i}$,$s$ and $\boldsymbol{\gamma}_{\neq i}$. □

Proof of Theorem 5 (Search Equivalence)

To prove search equivalence, we explicitly write a node as a tuple $q = \langle \boldsymbol{\varphi}^t, \hat{v}, \mathrm{PH}\rangle$, where $\boldsymbol{\varphi}^t$ is the past joint policy, $\hat{v}$ the node's heuristic value, and PH a boolean indicating whether it is a placeholder. We consider the *equivalence* of the maintained open lists. The open list L maintained by GMAA*-IC contains only non-expanded nodes $q$. In contrast, the open list $\mathrm{L}^{\mathrm{IE}}$ of GMAA*-ICE contains both non-expanded nodes $q$ and placeholders (previously expanded nodes), $\bar{q}$. We denote the ordered subset of $\mathrm{L}^{\mathrm{IE}}$ containing non-expanded nodes with $Q$ and that containing placeholders with $\bar{Q}$. We treat these open lists as ordered sets of heuristic values and their associated nodes.

**Definition 11.** L and $\mathrm{L}^{\mathrm{IE}}$ are *equivalent*, $\mathrm{L} \equiv \mathrm{L}^{\mathrm{IE}}$ if:

1. $Q \subseteq \mathrm{L}$.
2. The $q$'s have the same ordering: $\mathrm{L.remove}(\mathrm{L} \setminus Q) = Q$.[24]
3. Nodes $q$ in $L$ but not $Q$ have a placeholder $\bar{q}$ that is the parent of and higher ranked than $q$:

$$\forall_{q=\langle \boldsymbol{\varphi}^t,\hat{v}_q,\mathrm{false}\rangle\in(\mathrm{L}\setminus Q)} \quad \exists_{\bar{q}=\langle \boldsymbol{\varphi}^{t-1},\hat{v}_{\bar{q}},\mathrm{true}\rangle\in\bar{Q}} \quad s.t. \ \ (\boldsymbol{\varphi}^t = (\boldsymbol{\varphi}^{t-1},\boldsymbol{\beta}) \quad \wedge \quad q < \bar{q}).$$

4. There are no other placeholders.

Fig. 11 illustrates two equivalent lists in which the past joint policies are indexed with letters. Note that the placeholders in $\mathrm{L}^{\mathrm{IE}}$ are ranked higher than the nodes in L that they represent.

Let us write IT-IC(L) and IT-ICE($\mathrm{L}^{\mathrm{IE}}$) for one iteration (i.e., one loop of the main repeat in Algorithm 1) of the respective algorithms. Let IT-ICE* denote the operation that repeats IT-ICE as long as a placeholder was selected (so it ends when a $q$ is expanded).

**Lemma 4.** *If $L \equiv L^{IE}$, then executing IT-IC(L) and IT-ICE*($L^{IE}$) leads to new open lists that are again equivalent: $L' \equiv L^{IE\prime}$.*

*Proof.* When IT-ICE* selects a placeholder $\bar{q}$, it generates child $q'$ that was already present in L (due to properties 3 and 4 of Definition 11) and inserts it. Insertion occurs at the same relative location as IT-IC because both algorithms use the same comparison operator (Definition 5). Together these facts guarantee that the insertion preserves properties 1 and 2.

24. $A$.remove($B$) removes the elements of $B$ from $A$ without changing $A$'s ordering.

| L | | $L^{IE}$ | | | | | |
|---|---|---|---|---|---|---|---|
| | | $Q$ | | $\bar{Q}$ | | | |
| $\widehat{V}$ | $\boldsymbol{\varphi}^t$ | $\widehat{V}$ | $\boldsymbol{\varphi}^t$ | $\widehat{V}$ | $\boldsymbol{\varphi}^t$ | | |
| | | | | 8 | $a$ | $\leftarrow$ | placeholder for $\{c,e,j\}$ |
| 7 | $c$ | | | | | | |
| 5 | $d$ | 5 | $d$ | | | $\leftarrow$ | same nodes: same position |
| 4.5 | $e$ | | | | | | |
| | | | | 4 | $b$ | $\leftarrow$ | placeholder for $\{h,i\}$ |
| 3 | $f$ | 3 | $f$ | | | $\}$ | consistent ordering |
| 3 | $g$ | 3 | $g$ | | | | for equal values |
| 2.5 | $h$ | | | | | | |
| 1 | $i$ | | | | | | |
| 0.5 | $j$ | | | | | | |

Figure 11: Illustration of equivalent lists. Past joint policies are indexed by letters. In this example, $a$ and $b$ have been expanded earlier (but are not yet fully expanded in the ICE-case).

If there are remaining unexpanded children of $\bar{q}$, IT-ICE* reinserts $\bar{q}$ with an updated heuristic value $\bar{q}.\hat{v} \leftarrow q'.\hat{v}$ that is guaranteed to be an upper bound on the value of unexpanded siblings $q''$ since $q'.\hat{v} = \widehat{V}(q'.\boldsymbol{\varphi}) \geq \widehat{V}(q''.\boldsymbol{\varphi}) = q''.\hat{v}$ (preserving properties 3 and 4).

When IT-ICE* finally selects a non-placeholder $q$, it is guaranteed to be the same $q$ as selected by IT-IC (due to properties 1 and 2). Expansion in ICE generates one child $q'$ (again inserted at the same relative location as in IC) and inserts placeholder $\bar{q} = \langle q.\boldsymbol{\varphi}, q'.\hat{v}, \text{true}\rangle$ for the other siblings $q''$ (again preserving properties 3 and 4). $\square$

*Proof of Theorem 5.* The fact that GMAA*-ICE and GMAA*-IC are search-equivalent follows directly from Lemma 4. Search equivalence means that both algorithms select the same non-placeholders $q$ to expand. Since both algorithms begin with identical (and therefore trivially equivalent) open lists, they maintain equivalent open lists throughout search. As such, property 2 of Definition 11 ensures that every time IT-ICE* selects a non-placeholder, IT-IC selects it too. $\square$